%% file: Text2Video-Zero.tex
\documentclass[10pt,twocolumn,letterpaper]{article}

\usepackage{iccv}
\makeatletter
\@namedef{ver@everyshi.sty}{}
\makeatother
\usepackage{times}
\usepackage{epsfig}
\usepackage{graphicx}
\usepackage{xcolor}
\usepackage{amsmath}
\usepackage{amssymb}
\usepackage{booktabs}
\usepackage{float}
\usepackage{array}
\usepackage{varwidth}
\newcolumntype{M}[1]{>{\centering\arraybackslash}m{#1}}
\usepackage{makecell}
\usepackage{algorithm}
\usepackage[noEnd,indLines]{algpseudocodex}
\newcommand{\authorsep}{\hspace{8pt}}
\newcommand{\affiliationsep}{\hspace{8pt}}

\usepackage{cuted}
\usepackage{caption}
\usepackage{subcaption}
\usepackage{capt-of}
\usepackage[pagebackref=true,breaklinks=true,letterpaper=true,colorlinks,bookmarks=false]{hyperref}

\iccvfinalcopy 

\def\iccvPaperID{****} 

\ificcvfinal\fi

\begin{document}

\title{Text2Video-Zero: \\ Text-to-Image Diffusion Models are Zero-Shot Video Generators
}


\author{Levon Khachatryan$^1$$^{*}$ \authorsep
Andranik Movsisyan$^1$$^{*}$\authorsep 
Vahram Tadevosyan$^1$$^{*}$ \authorsep
Roberto Henschel$^1$$^{*}$ \authorsep \\
Zhangyang Wang$^{1,2}$ 
Shant Navasardyan$^1$ \authorsep
Humphrey Shi$^{1,3,4}$ \\
\small$^1$Picsart AI Resarch (PAIR) \affiliationsep \small$^2$UT Austin \affiliationsep \small$^3$U of Oregon \affiliationsep \small$^4$UIUC \affiliationsep \\
    {\small \textbf{\url{https://github.com/Picsart-AI-Research/Text2Video-Zero}}}
}

\date{}
\maketitle

\begin{strip}
\vspace{-15mm}
    \centering
    \includegraphics[width = \textwidth]{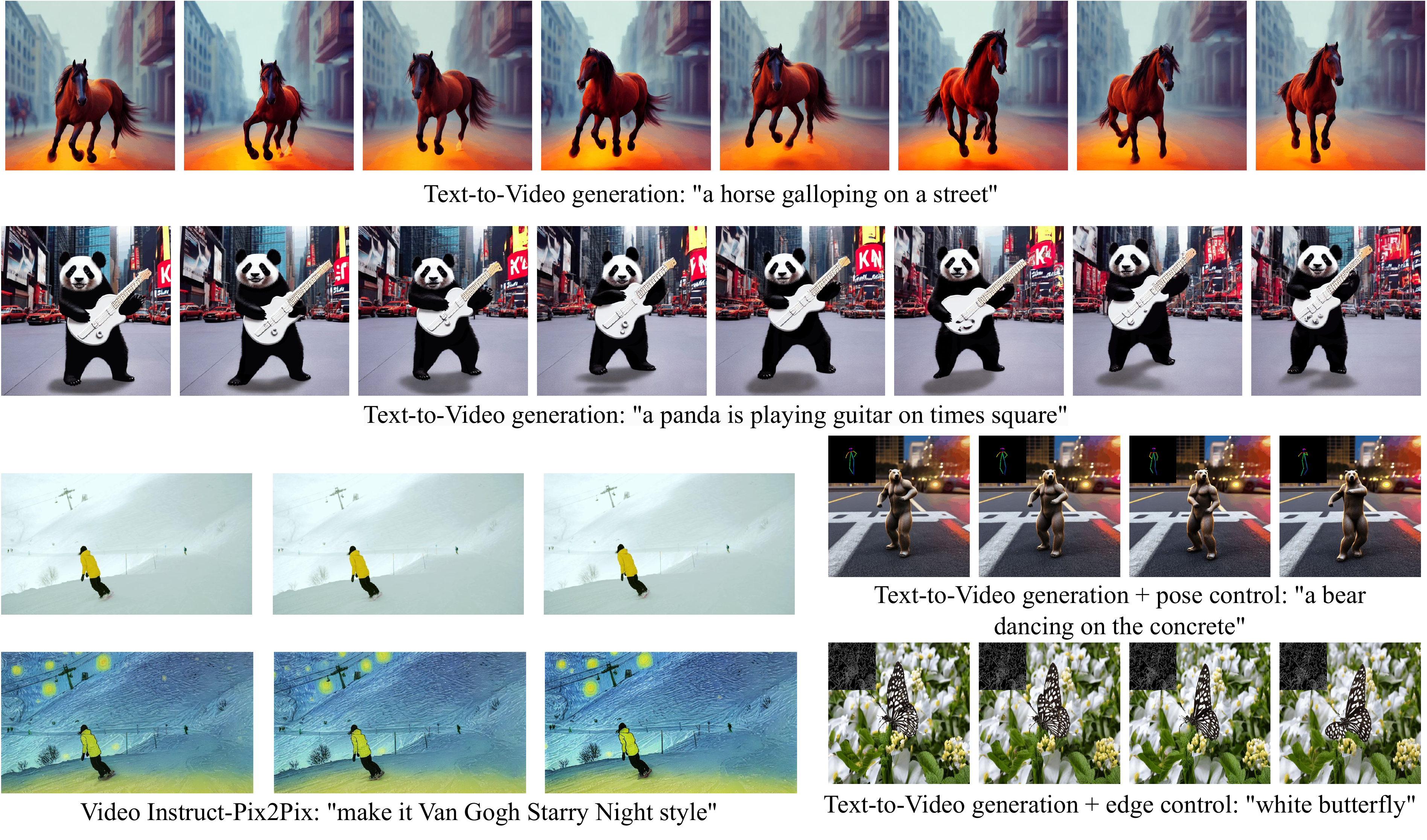}
    \captionof{figure}{Our method Text2Video-Zero enables zero-shot video generation using (i) a textual prompt (see rows 1, 2),  (ii) a prompt combined with guidance from poses or edges (see lower right), and  (iii)  Video Instruct-Pix2Pix, \ie, instruction-guided video editing (see lower left). 
    Results are temporally consistent and follow closely the guidance and textual prompts. 
    }
    \label{fig:teaser_img}
\end{strip}

\ificcvfinal\thispagestyle{empty}\fi

\def\thefootnote{*}\footnotetext{Equal contribution.}\def\thefootnote{\arabic{footnote}}

\begin{abstract}
\vspace{-4mm}
    Recent text-to-video generation approaches rely on computationally heavy training and require large-scale video datasets. In this paper, we introduce a new task of zero-shot text-to-video generation and propose a low-cost approach (without any training or optimization) by leveraging the power of existing text-to-image synthesis methods (\eg Stable Diffusion), making them suitable for the video domain.  
    Our key modifications include (i) enriching the latent codes of the generated frames with motion dynamics to keep the global scene and the background time consistent; and (ii) reprogramming frame-level self-attention using a new cross-frame attention of each frame on the first frame, to preserve the context, appearance, and identity of the foreground object. 
    Experiments show that this leads to low overhead, yet high-quality and remarkably consistent video generation. Moreover, our approach is not limited to text-to-video synthesis but is also applicable to other tasks such as conditional and content-specialized video generation, and Video Instruct-Pix2Pix, \ie, instruction-guided video editing.
    As experiments show, our method performs comparably or sometimes better than recent approaches, despite not being trained on additional video data. Our code will be open sourced at: \href{https://github.com/Picsart-AI-Research/Text2Video-Zero}{https://github.com/Picsart-AI-Research/Text2Video-Zero}.
\end{abstract}


\section{Introduction}
In recent years, generative AI has attracted enormous attention in the computer vision community. With the advent of diffusion models \cite{sohl2015deep,DDPM_paper, DDIM_paper,song2020score}, it has become tremendously popular and successful to generate high-quality images from textual prompts, also called \textit{text-to-image synthesis} \cite{Dalle2_paper,stable_diff,imagen,make-a-scene,xu2022versatile}.
Recent works \cite{video-diffusion-models, make-a-video, imagen_video, tune-a-video, gen1_paper, molad2023dreamix} attempt to extend the success to text-to-video generation and editing tasks, by reusing text-to-image diffusion models in the video domain.
While such approaches yield promising outcomes, most of them require substantial training with a massive amount of labeled data which can be costly and unaffordable for many users.
With the aim of making video generation cheaper, Tune-A-Video \cite{tune-a-video} introduces a mechanism that can adopt Stable Diffusion (SD) model \cite{stable_diff} for the video domain. The training effort is drastically reduced to tuning one video. 
While that is much more efficient than previous approaches, it still requires an optimization process.
In addition, the generation abilities of Tune-A-Video are limited to text-guided video editing applications; video synthesis from scratch, however, remains out of its reach.

In this paper, we take one step forward in studying the novel problem of \textit{zero-shot, ``training-free" text-to-video synthesis}, which is the task of generating videos from textual prompts without requiring any optimization or fine-tuning. 
%
A key concept of our approach is to modify   
a pre-trained text-to-image model (\eg, Stable Diffusion), enriching it with temporally consistent generation. 
%
%
By building upon already trained text-to-image models, our method takes advantage of their excellent image generation quality and enhances their applicability to the video domain without performing additional training.
To enforce temporal consistency, we present two innovative and lightweight modifications: (1) we first enrich the latent codes of generated frames with motion information to keep the global scene and the background time consistent; (2) we then use cross-frame attention of each frame on the first frame to preserve the context, appearance, and identity of the foreground object throughout the entire sequence.
Our experiments show that these simple modifications lead to high-quality and time-consistent video generations (see Fig.~\ref{fig:teaser_img} and further results in the appendix). Despite the fact that other works train on large-scale video data, our method achieves similar or sometimes even better performance (see Figures \ref{fig:ourVSCogVideo}, \ref{fig:videoediting} and appendix Figures \ref{fig:unconstrained-text2video-ours-vs-cog}, \ref{fig:table_of_EDITING/comparison1}, \ref{fig:table_of_EDITING/comparison2}). 
Furthermore, our method is not limited to text-to-video synthesis but is also applicable to conditional (see Figures \ref{fig:edgeControl}, \ref{fig:poseControl} and appendix Figures \ref{fig:qual_results_full_method_edge_guidance}, \ref{fig:qual_results_pose}, \ref{fig:table_of_edge_study}, \ref{fig:table_of_pose_study}) and specialized video generation (see Fig.~\ref{fig:edgeControl_with_DB}), and instruction-guided video editing, which we refer as \textit{Video Instruct-Pix2Pix} motivated by Instruct-Pix2Pix \cite{brooks2022instructpix2pix} (see Fig.~\ref{fig:videoediting} and appendix Figures \ref{fig:video_editing_results}, \ref{fig:table_of_EDITING/comparison1}, \ref{fig:table_of_EDITING/comparison2}). 

Our contributions are summarized as three-folds:
\begin{itemize}
    \item A new problem setting of zero-shot text-to-video synthesis, aiming at making text-guided video generation and editing ``freely affordable". We use only a pre-trained text-to-image diffusion model without any further fine-tuning or optimization. 
    \item Two novel post-hoc techniques to enforce temporally consistent generation, via encoding motion dynamics in the latent codes, and reprogramming each frame's self-attention using a new cross-frame attention. 
    \item A broad variety of applications that demonstrate our method's effectiveness, including conditional and specialized video generation, and \textit{Video Instruct-Pix2Pix} \ie, video editing by textual instructions.
\end{itemize}


\section{Related Work}

\subsection{Text-to-Image Generation}
Early approaches to text-to-image synthesis relied on methods such as template-based generation \cite{mansimov2015generating} and feature matching \cite{reed2016learning}. However, these methods were limited in their ability to generate realistic and diverse images. 

Following the success of GANs \cite{goodfellow2020generative}, several other deep learning-based methods were proposed for text-to-image synthesis. These include StackGAN \cite{zhang2017stackgan}, AttnGAN \cite{xu2018attngan}, and MirrorGAN \cite{qiao2019mirrorgan}, which further improve image quality and diversity by introducing novel architectures and attention mechanisms.

Later, with the advancement of transformers \cite{vaswani2017attention}, new approaches emerged for text-to-image synthesis. 
Being a 12-billion-parameter transformer model, Dall-E \cite{Dalle_paper} introduces two-stage training process: First, it generates image tokens, which later are combined with text tokens for joint training of an autoregressive model. 
Later Parti \cite{parti_paper} proposed a method to generate content-rich images with multiple objects. 
Make-a-Scene \cite{make-a-scene} enables a control mechanism by segmentation masks for text-to-image generation.

Current approaches build upon diffusion models, thereby taking text-to-image synthesis quality to the next level. 
%
GLIDE \cite{nichol2021glide} improved Dall-E by adding classifier-free guidance \cite{ho2022classifier}.
Later, Dall-E 2 \cite{Dalle2_paper} utilizes the contrastive model CLIP \cite{CLIP_paper}. By means of diffusion processes, (i) a mapping from CLIP text encodings to image encodings, and (ii) a CLIP decoder is obtained. 
LDM / SD \cite{stable_diff} applies a diffusion model on lower-resolution encoded signals of VQ-GAN \cite{VQ_GAN_paper}, showing competitive quality with a significant gain in speed and efficiency.
Imagen \cite{imagen} shows incredible performance in text-to-image synthesis by utilizing large language models for text processing. Versatile Diffusion~\cite{xu2022versatile} further unifies text-to-image, image-to-text and variations in a single multi-flow diffusion model.

Because of their great image quality, it is desired to exploit text-to-image models for video generation. However,  applying diffusion models in the video domain is not straightforward, especially due to their probabilistic generation procedure, making it difficult to ensure temporal consistency. As we show in our ablation experiments with Fig.~\ref{fig:abltion_study} (see also appendix), our modifications are crucial for temporal consistency in terms of both global scene and background motion, and for the preservation of the foreground object identity.

\subsection{Text-to-Video Generation}
Text-to-video synthesis is a relatively new research direction. Existing approaches try to leverage autoregressive transformers and diffusion processes for the generation. 
NUWA \cite{wu2022nuwa} introduces a 3D transformer encoder-decoder framework and supports both text-to-image and text-to-video generation.
Phenaki \cite{villegas2022phenaki} introduces a bidirectional masked transformer with a causal attention mechanism that allows the generation of arbitrary-long videos from text prompt sequences.
CogVideo \cite{hong2022cogvideo} extends the text-to-image model CogView 2 \cite{ding2022cogview2} by tuning it using a multi-frame-rate hierarchical training strategy to better align text and video clips.
Video Diffusion Models (VDM) \cite{video-diffusion-models} naturally extend text-to-image diffusion models and train jointly on image and video data.
Imagen Video \cite{imagen_video} constructs a cascade of video diffusion models and utilizes spatial and temporal super-resolution models to generate high-resolution time-consistent videos.
Make-A-Video \cite{make-a-video} builds upon a text-to-image synthesis model and leverages video data in an unsupervised manner.
Gen-1 \cite{gen1_paper} extends SD and proposes a structure and content-guided video editing method based on visual or textual descriptions of desired outputs.
Tune-A-Video \cite{tune-a-video} proposes a new task of one-shot video generation by extending and tuning SD on a single reference video.

Unlike the methods mentioned above, our approach is completely training-free, does not require massive computing power or dozens of GPUs, which makes the video generation process affordable for everyone. 
In this respect, Tune-a-Video \cite{tune-a-video} comes closest to our work, as it reduces the necessary computations to tuning on only one video. However, it still requires an optimization process and is heavily dependent on the reference video.



\section{Method}
We start this section with a brief introduction of diffusion models, particularly Stable Diffusion (SD) \cite{stable_diff}. 
Then we introduce the problem formulation of zero-shot text-to-video synthesis, followed by a subsection presenting our approach.
After that, to show the universality of our method, we use it in combination with ControlNet \cite{controlnet} and DreamBooth \cite{ruiz2022dreambooth} diffusion models for generating conditional and specialized videos.
Later we demonstrate the power of our approach with the application of instruction-guided video editing, namely, Video Instruct-Pix2Pix.


\subsection{Stable Diffusion}
SD is a diffusion model operating in the latent space of an autoencoder $\mathcal{D}(\mathcal{E}(\cdot))$, namely VQ-GAN \cite{VQ_GAN_paper} or VQ-VAE \cite{VQ_VAE_paper}, where $\mathcal{E}$ and $\mathcal{D}$ are the corresponding encoder and decoder, respectively. 
More precisely if $x_0\in\mathbb{R}^{h\times w\times c}$ is the latent tensor of an input image $Im$ to the autoencoder, i.e. $x_0 = \mathcal{E}(Im)$, diffusion forward process iteratively adds Gaussian noise to the signal $x_0$:
\begin{equation}
    q(x_t|x_{t-1}) = \mathcal{N}(x_t; \sqrt{1-\beta_t}x_{t-1}, \beta_t I), \;
    t = 1,..,T
\end{equation}
where $q(x_t|x_{t-1})$ is the conditional density of $x_t$ given $x_{t-1}$, and $\{\beta_t\}_{t=1}^{T}$ are hyperparameters.
$T$ is chosen to be as large that the forward process completely destroys the initial signal $x_0$ resulting in $x_T \sim \mathcal{N}(0,I)$. 
The goal of SD is then to learn a backward process
\begin{equation}
    p_\theta(x_{t-1}|x_t) = \mathcal{N}(x_{t-1};\mu_\theta(x_t,t),\Sigma_\theta(x_t,t))
\end{equation}
for $t=T,\ldots,1$, 
which allows to generate a valid signal $x_0$ from the standard Gaussian noise $x_T$. 
To get the final image generated from $x_T$ it remains to pass $x_0$ to the decoder of the initially chosen autoencoder: $Im = \mathcal{D}(x_0)$.

After learning the abovementioned backward diffusion process (see DDPM \cite{DDPM_paper}) one can apply a deterministic sampling process, called DDIM \cite{DDIM_paper}:
\begin{equation}
\begin{split}
    x_{t-1} = \sqrt{\alpha_{t-1}}\left(\frac{x_t - \sqrt{1-\alpha_t}\epsilon^t_{\theta}(x_t)}{\sqrt{\alpha_t}}\right) + \\
    \sqrt{1-\alpha_{t-1}}\epsilon^t_{\theta}(x_t), \quad t=T,\ldots,1, 
\end{split}
\end{equation}
where $\alpha_t = \prod_{i=1}^{t}(1-\beta_i)$ and 
\begin{equation}
    \epsilon^t_{\theta}(x_t) = \frac{\sqrt{1-\alpha_t}}{\beta_t}x_t + 
    \frac{(1-\beta_t)(1-\alpha_t)}{\beta_t}\mu_\theta(x_t,t).
\end{equation}

To get a text-to-image synthesis framework, SD guides the diffusion processes with a textual prompt $\tau$. Particularly for DDIM sampling, we get:
\begin{equation}
\begin{split}
    x_{t-1} = \sqrt{\alpha_{t-1}}\left(\frac{x_t - \sqrt{1-\alpha_t}\epsilon^t_{\theta}(x_t,\tau)}{\sqrt{\alpha_t}}\right) + \\
    \sqrt{1-\alpha_{t-1}}\epsilon^t_{\theta}(x_t,\tau), \quad t=T,\ldots,1.
\end{split}
\end{equation}
It is worth noting that in SD, the function $\epsilon^t_{\theta}(x_t,\tau)$ is modeled as a neural network with a UNet-like \cite{UNet_paper} architecture composed of convolutional and (self- and cross-) attentional blocks.
$x_T$ is called the latent code of the signal $x_0$ and there is a method \cite{dhariwal2021diffusion} to apply a deterministic forward process to reconstruct the latent code $x_T$ given a signal $x_0$. 
This method is known as DDIM inversion. Sometimes for simplicity, we will call $x_t, t=1,\ldots,T$ also the \textit{latent codes} of the initial signal $x_0$.

\subsection{Zero-Shot Text-to-Video Problem Formulation}
Existing text-to-video synthesis methods require either costly training on a large-scale (ranging from $1M$ to $15M$ data-points)
text-video paired data \cite{wu2022nuwa,hong2022cogvideo,villegas2022phenaki,imagen_video,video-diffusion-models,gen1_paper} or tuning on a reference video \cite{tune-a-video}.
To make video generation cheaper and easier, we propose a new problem: zero-shot text-to-video synthesis. Formally, given a text description $\tau$ and a positive integer $m\in\mathbb{N}$, the goal is to design a function $\mathcal{F}$ that outputs video frames  $\mathcal{V} \in\mathbb{R}^{m \times H\times W\times 3}$ (for predefined resolution $H\times W$) that exhibit temporal consistency. 
To determine the function $\mathcal{F}$, no training or fine-tuning must be performed on a video dataset.

Our problem formulation provides a new paradigm for text-to-video. Noticeably, a zero-shot text-to-video method naturally leverages quality improvements of text-to-image models.   

\subsection{Method}

In this paper, we approach the zero-shot text-to-video task by exploiting the text-to-image synthesis power of Stable Diffusion (SD).
As we need to generate videos instead of images, SD should operate on sequences of latent codes.
The na\"ive approach is to independently sample $m$ latent codes from standard Gaussian distribution $x^1_T,\ldots,x^m_T\sim \mathcal{N}(0,I)$ and apply DDIM sampling to obtain the corresponding tensors $x_0^{k}$ for $k = 1,\ldots, m$, followed by decoding to obtain the generated video sequence $\{\mathcal{D}(x^k_0)\}_{k=1}^{m}\in\mathbb{R}^{m\times H\times W\times 3}$.
However, as shown in Fig.~\ref{fig:abltion_study} (first row), this leads to completely random generation of images sharing only the semantics described by $\tau$ but neither object appearance nor motion coherence.

\begin{figure*}[t]
    \centering
    \includegraphics[width = \textwidth]{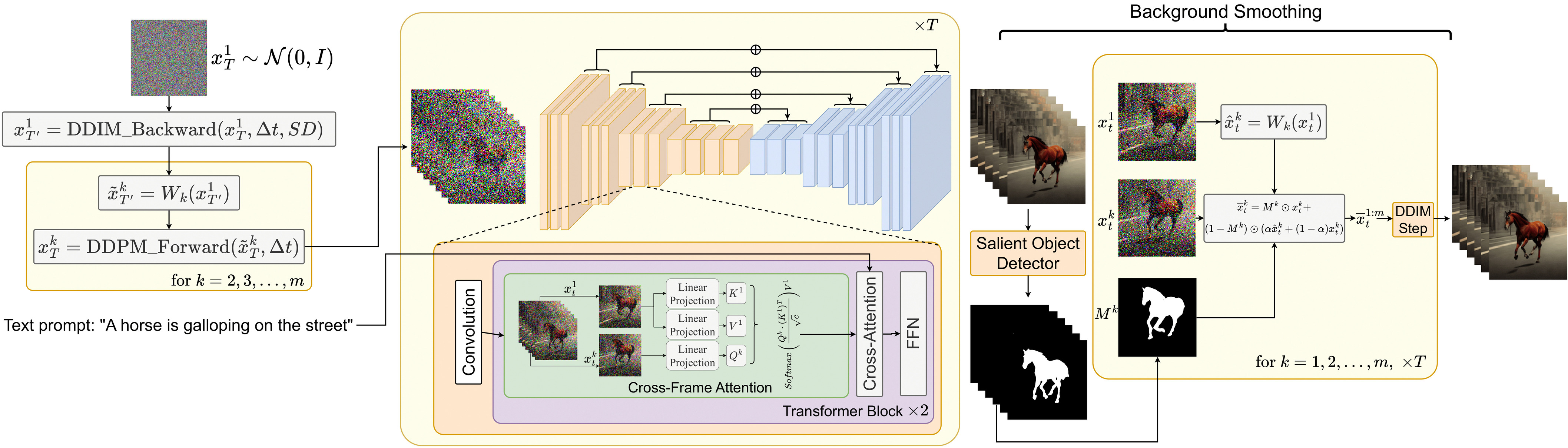}
    \caption{Method overview: Starting from a randomly sampled latent code $x_{T}^{1}$, we apply $\Delta t$ DDIM backward steps to obtain $x_{T'}^{1}$ using a pre-trained Stable Diffusion model (SD). A specified motion field results for each frame $k$ in a warping function $W_k$ that turns $x_{T'}^{1}$ to $x_{T'}^{k}$. By enhancing the latent codes with motion dynamics, we determine the global scene and camera motion and achieve temporal consistency in the background and the global scene. 
    A subsequent DDPM forward application delivers latent codes $x_{T}^{k}$ for $k=1,\ldots,m$. By using the (probabilistic) DDPM method, a greater degree of freedom is achieved with respect to the motion of objects (see appendix Sec. \ref{apendix:unconstrained-text2video_ablation}).
    Finally, the latent codes are passed to our modified SD model using the proposed cross-frame attention, which uses keys and values from the first frame to generate the image of frame $k=1,\ldots,m$. By using cross-frame attention,  the appearance and the identity of the foreground object are preserved  throughout the sequence. 
    Optionally, we apply background smoothing. To this end, we employ salient object detection to obtain for each frame $k$ a mask $M^{k}$ indicating the foreground pixels. Finally, for the background (using the mask $M^{k}$), a convex combination between the latent code $x_{t}^{1}$ of frame one warped to frame $k$ and the latent code $x_{t}^{k}$  is used to further improve the temporal consistency of the background. 
    %
    %
    %
    %
    }
    \label{fig:main_diagram}
\end{figure*}

To address this issue, we propose to (i) introduce motion dynamics between the latent codes $x^1_T,\ldots,x^m_T$ to keep the global scene time consistent and (ii) use cross-frame attention mechanism to preserve the appearance and the identity of the foreground object. Each of the components of our method are described below in detail.
The overview of our method can be found in Fig.~\ref{fig:main_diagram}.

Note, to simplify notation, we will denote the entire sequence of latent codes by $x^{1:m}_T = [x^1_T,\ldots,x^m_T]$.


\subsubsection{Motion Dynamics in Latent Codes}
Instead of sampling the latent codes $x^{1:m}_T$ randomly and independently from the standard Gaussian distribution, we \textit{construct} them by performing the following steps (see also Algorithm \ref{alg:motion_dynamics_in_latents} and Fig.~\ref{fig:main_diagram}).
\begin{algorithm}[t]
    \caption{Motion dynamics in latent codes}
    \begin{algorithmic}[1]
        \Require $\Delta t \geq 0, m\in \mathbb{N}, \lambda > 0, \delta=(\delta_x,\delta_y)\in\mathbb{R}^2, \mbox{Stable Diffusion } (SD)$
        \State 
            $x^1_T\sim\mathcal{N}(0,I)$
            \Comment{random sample the first latent code}
        \State
            $x^1_{T'} \leftarrow \mbox{DDIM\_Backward}(x^1_{T},\Delta t, SD)$ 
            \Comment{perform $\Delta t$ backward steps by SD}
        \For {all $k=2,3,\ldots,m$} 
            \State
            $\delta^k \leftarrow \lambda\cdot(k-1)\delta$ 
            \Comment{computing global translation vectors}
            \State
            $W_k \leftarrow \mbox{Warping by }\delta^k$
            \Comment{defining warping functions}
            \State
            $\tilde{x}^k_{T'} \leftarrow W_k(x^1_{T'})$
            \State
            $x^k_{T} \leftarrow \mbox{DDPM\_Forward}(\tilde{x}^k_{T'},\Delta t)$
            \Comment{DDPM forward for more motion freedom}
        \EndFor
        \Return $x^{1:m}_T$
    \end{algorithmic}
    \label{alg:motion_dynamics_in_latents}
\end{algorithm}
\begin{enumerate}
    \item Randomly sample the latent code of the first frame: $x^1_T\sim\mathcal{N}(0,I)$.
    \item Perform $\Delta t \geq 0$ DDIM backward steps on the latent code $x^1_T$ by using the SD model and get the corresponding latent $x^1_{T'}$, where $T' = T - \Delta t$.
    \item Define a direction $\delta=(\delta_x,\delta_y)\in\mathbb{R}^2$ for the global scene and camera motion. By default $\delta$ can be the main diagonal direction $\delta_x = \delta_y = 1$.
    \item For each frame $k=1,2,\ldots,m$ we want to generate, compute the global translation vector $\delta^k = \lambda\cdot (k-1)\delta$, where $\lambda$ is a hyperparameter controlling the amount of the global motion.
    \item Apply the constructed motion (translation) flow $\delta^{1:m}$ to $x^1_{T'}$, denote the resulting sequence by $\tilde{x}^{1:m}_{T'}$:
    \begin{equation}
        \tilde{x}^{k}_{T'} = W_k(x^1_{T'}) \; 
        \mbox{for } k = 1,2,\ldots,m,
    \end{equation}
    where $W_k(x^1_{T'})$ is the warping operation for translation by the vector $\delta^k$.
    \item Perform $\Delta t$ DDPM forward steps on each of the latents $\tilde{x}^{2:m}_{T'}$ and get the corresponding latent codes $x^{2:m}_T$.
\end{enumerate}

Then we take the sequence $x^{1:m}_T$ as the starting point of the backward (video) diffusion process.
As a result, the latent codes generated with our proposed motion dynamics lead to better temporal consistency of the global scene as well as the background, see Fig.~\ref{fig:abltion_study}.
Yet, the initial latent codes are not constraining enough to describe particular colors, identities or shapes, thus still leading to  temporal inconsistencies, especially for the foreground object.

\subsubsection{Reprogramming Cross-Frame Attention}
To address the issue mentioned above, we use a cross-frame attention mechanism to preserve the information about (in particular) the foreground object's appearance, shape, and identity throughout the generated video. 

To leverage the power of cross-frame attention and at the same time exploit a pretrained SD without retraining, we replace each of its self-attention layers with a cross-frame attention, with the attention for each frame being on the first frame.
More precisely in the original SD UNet architecture  $\epsilon^t_\theta(x_t,\tau)$, each self-attention layer takes a feature map $x\in\mathbb{R}^{h\times w\times c}$, linearly projects it into query, key, value features $Q,K,V\in \mathbb{R}^{h\times w\times c}$, and computes the layer output by the following formula (for simplicity  described here for only one attention head) \cite{vaswani2017attention}:
\begin{equation}
    \mbox{Self-Attn}(Q,K,V) = \mbox{Softmax}\left(\frac{QK^T}{\sqrt{c}}\right)V.
\end{equation}
In our case, each attention layer receives $m$ inputs: $x^{1:m} = [x^1,\ldots,x^m]\in\mathbb{R}^{m\times h\times w\times c}$. Hence,  the linear projection layers produce  $m$ queries, keys, and values $Q^{1:m}, K^{1:m}, V^{1:m}$, respectively.

Therefore, we replace each self-attention layer with a cross-frame attention of each frame on the first frame as follows:
\begin{equation}
\label{eq:cross-frame-attn}
\begin{split}
    \mbox{Cross-Frame-Attn}(Q^{k},K^{1:m},V^{1:m})= \\
    \mbox{Softmax}\left(\frac{Q^k(K^1)^T}{\sqrt{c}}\right)V^1
\end{split}
\end{equation}
for $k=1,\ldots,m$.
By using cross frame attention, the appearance and structure of the objects and background as well as identities are carried over from the first frame to subsequent frames, which significantly increases the temporal consistency of the generated frames (see Fig.~\ref{fig:abltion_study} and the appendix, Figures \ref{fig:ablation-unconditional-attention-motion}, \ref{fig:table_of_edge_study}, \ref{fig:table_of_pose_study}).



\begin{figure*}
    \centering
    \begin{subfigure}{\textwidth}
    \includegraphics[width=0.12\textwidth]{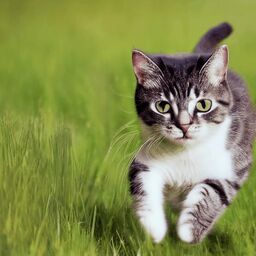} 
    \hfill
    \includegraphics[width=0.12\textwidth]{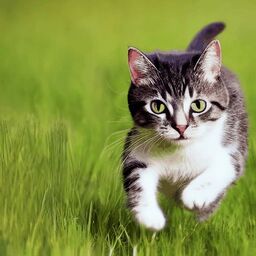} 
    \hfill
    \includegraphics[width=0.12\textwidth]{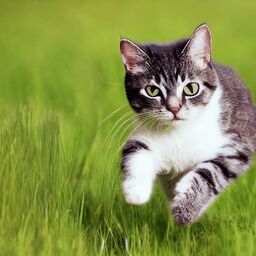} 
    \hfill
    \includegraphics[width=0.12\textwidth]{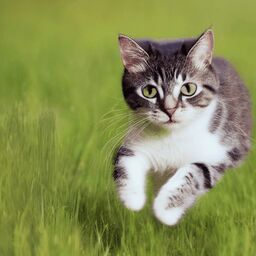} 
    \hfill
    \includegraphics[width=0.12\textwidth]{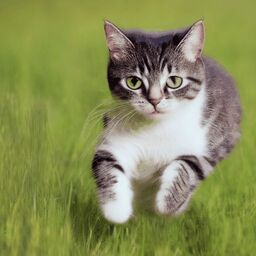} 
    \hfill
    \includegraphics[width=0.12\textwidth]{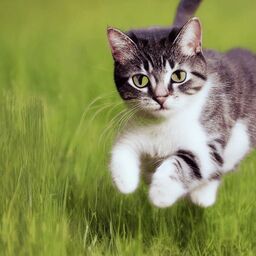} 
    \hfill
    \includegraphics[width=0.12\textwidth]{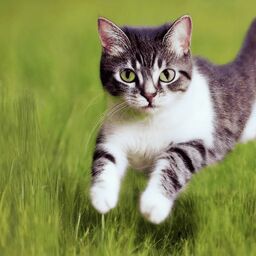} 
    \hfill
    \includegraphics[width=0.12\textwidth]{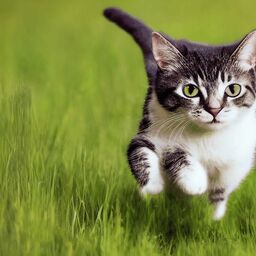} 
    \vskip -.4\baselineskip
    \caption{a high quality realistic photo of a cute cat running in a beautiful meadow}
    \end{subfigure}
    \vskip .3\baselineskip
    \begin{subfigure}{\textwidth}
    \includegraphics[width=0.12\textwidth]{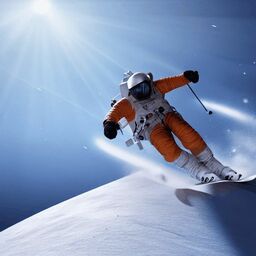} 
    \hfill
    \includegraphics[width=0.12\textwidth]{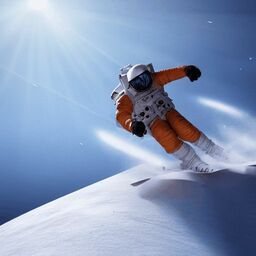} 
    \hfill
    \includegraphics[width=0.12\textwidth]{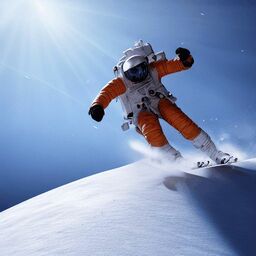} 
    \hfill
    \includegraphics[width=0.12\textwidth]{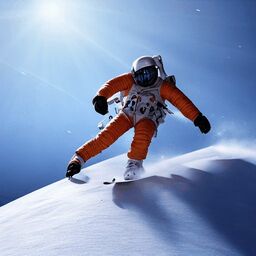} 
    \hfill
    \includegraphics[width=0.12\textwidth]{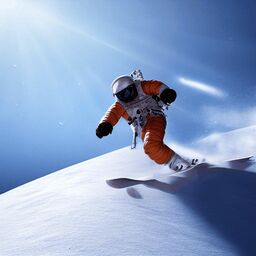} 
    \hfill
    \includegraphics[width=0.12\textwidth]{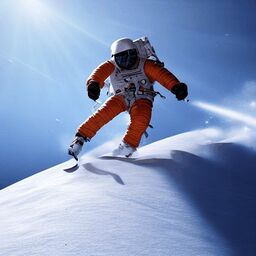} 
    \hfill
    \includegraphics[width=0.12\textwidth]{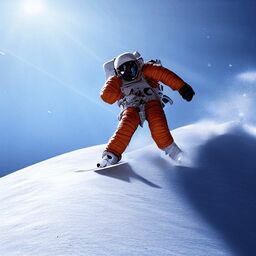} 
    \hfill
    \includegraphics[width=0.12\textwidth]{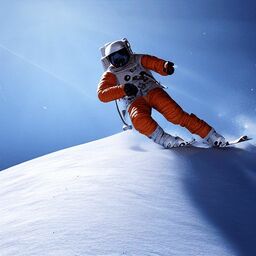}
    \vskip -.4\baselineskip
    \caption{an astronaut is skiing down a hill}
    \end{subfigure}
    \vskip .3\baselineskip
    \begin{subfigure}{\textwidth}
    \includegraphics[width=0.12\textwidth]{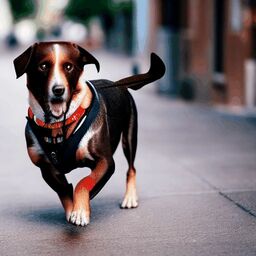} 
    \hfill
    \includegraphics[width=0.12\textwidth]{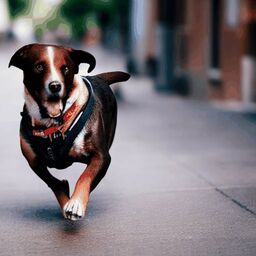} 
    \hfill
    \includegraphics[width=0.12\textwidth]{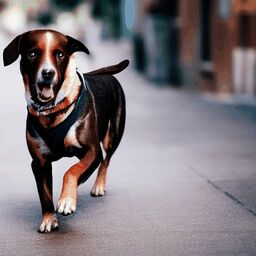} 
    \hfill
    \includegraphics[width=0.12\textwidth]{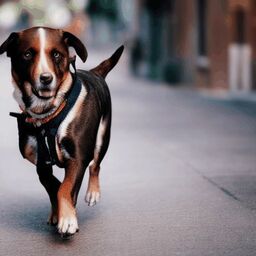} 
    \hfill
    \includegraphics[width=0.12\textwidth]{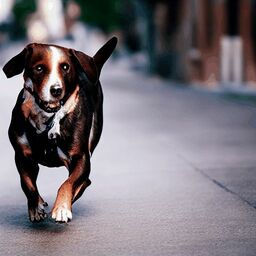} 
    \hfill
    \includegraphics[width=0.12\textwidth]{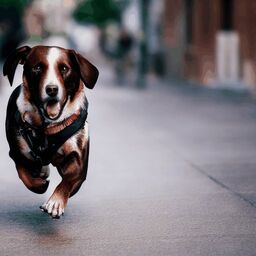} 
    \hfill
    \includegraphics[width=0.12\textwidth]{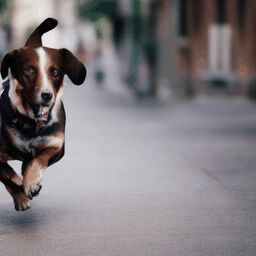} 
    \hfill
    \includegraphics[width=0.12\textwidth]{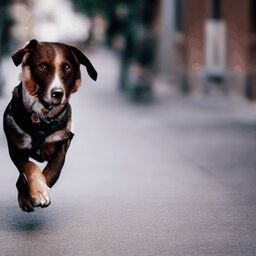}
    \vskip -.4\baselineskip
    \caption{a dog is walking down the street}
    \end{subfigure}
    \vskip .3\baselineskip

    \begin{subfigure}{\textwidth}
    \includegraphics[width=0.12\textwidth]{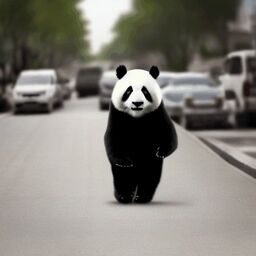} 
    \hfill
    \includegraphics[width=0.12\textwidth]{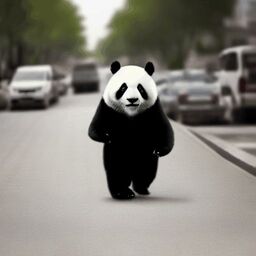} 
    \hfill
    \includegraphics[width=0.12\textwidth]{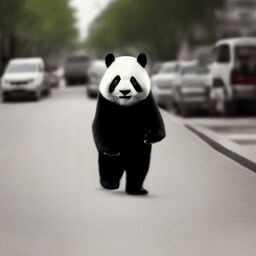} 
    \hfill
    \includegraphics[width=0.12\textwidth]{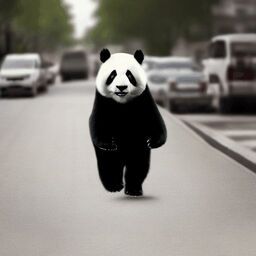} 
    \hfill
    \includegraphics[width=0.12\textwidth]{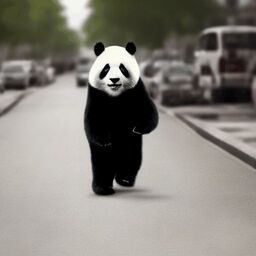} 
    \hfill
    \includegraphics[width=0.12\textwidth]{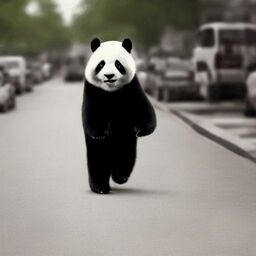} 
    \hfill
    \includegraphics[width=0.12\textwidth]{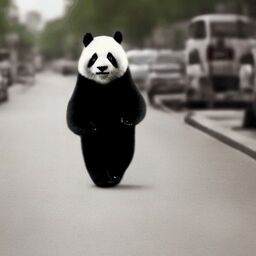} 
    \hfill
    \includegraphics[width=0.12\textwidth]{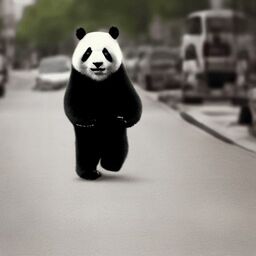}
    \vskip -.4\baselineskip
    \caption{a high quality realistic photo of a panda walking alone down the street}
    \end{subfigure}
    \vskip -.3\baselineskip
    \caption{Text-to-Video results of our method. Depicted frames show that identities and appearances are temporally consistent and fitting to the textual prompt. For more results, see Appendix Sec. \ref{apendix:unconstrained-text2video}.}
    \label{fig:text2video_results}
\end{figure*}

\begin{figure}[t]
    \centering
    \includegraphics[width = 1\linewidth]{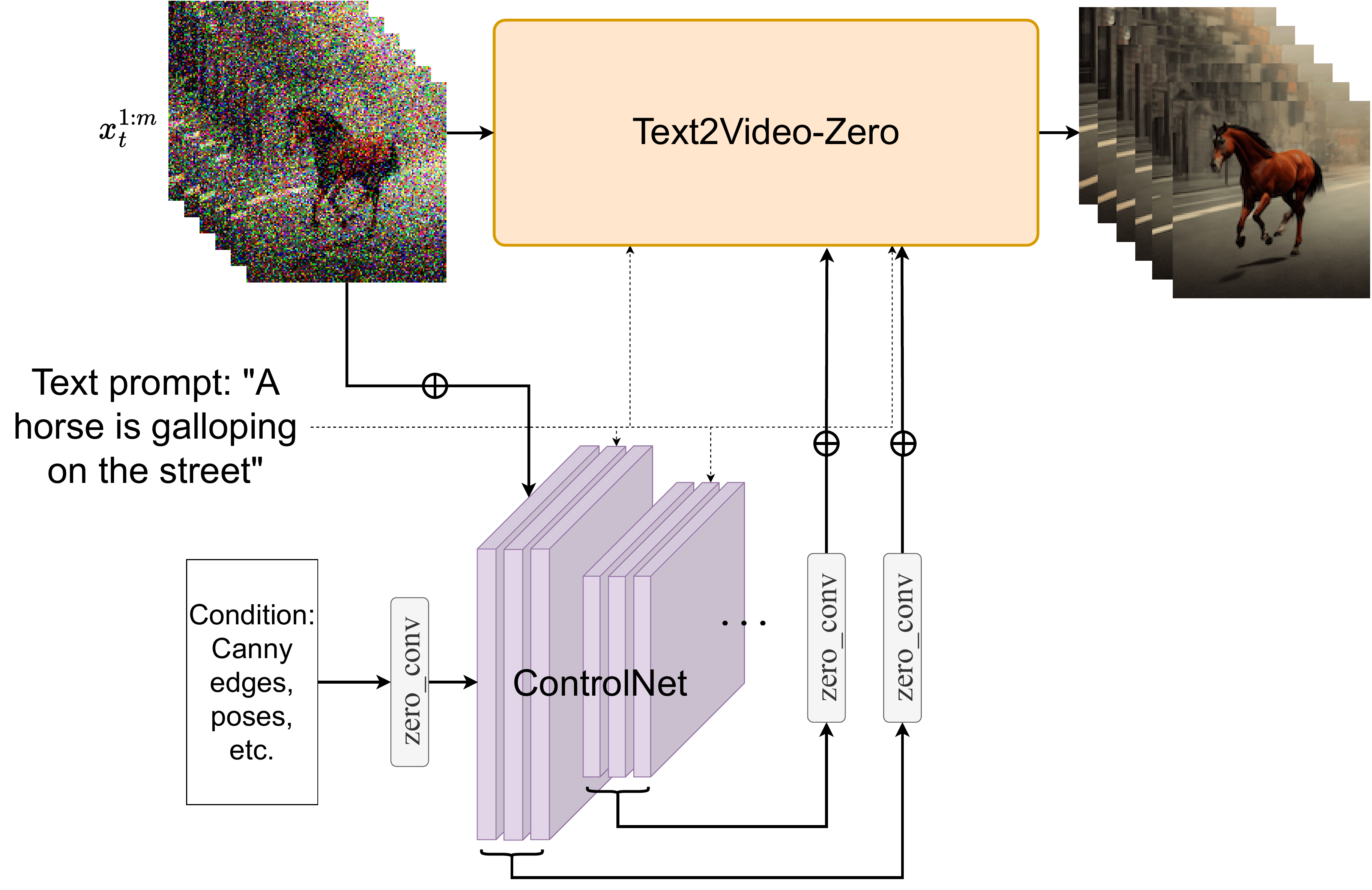}
    \caption{The overview of Text2Video-Zero + ControlNet}
    \label{fig:AddingControlNet}
\end{figure}


\subsubsection{Background smoothing (Optional)}
We further improve temporal consistency of the background using a convex combination of background-masked latent codes between the first frame and frame $k$.
This is especially helpful for video generation from textual prompts when one or no initial image and no further guidance are provided. 

In more detail, given the generated sequence of our video generator, $x_{0}^{1:m}$, we apply (an in-house solution for) salient object detection \cite{wang2021salient} to the decoded images to obtain a corresponding foreground mask $M^{k}$ for each frame $k$.
Then we warp $x^1_{t}$ according to the employed  motion dynamics defined by $W_k$ and denote the result by $\hat{x}_{t}^{k} := W_k(x_{t}^{1})$.

Background smoothing is achieved by a convex combination between the actual latent code $x_{t}^{k}$ and the warped latent code $\hat{x}_{t}^{k}$ on the background, i.e., 
\begin{equation}
    \overline{x}_{t}^{k} = M^{k} \odot x_{t}^{k}  + (1 - M^{k}) \odot (\alpha \hat{x}_{t}^{k}+ (1- \alpha) x_{t}^{k}),
\end{equation}
for $k = 1,\ldots, m$, where $\alpha$ is a hyperparameter, which we empirically choose $\alpha=0.6$. Finally, DDIM sampling is employed on $\overline{x}_{t}^{k}$, which delivers video generation with background smoothing.
We use background smoothing in our video generation from text when no guidance is provided. For an ablation study on background smoothing, see the appendix, Sec. \ref{apendix:unconstrained-text2video_ablation}.

\subsection{Conditional and Specialized Text-to-Video}

Recently powerful controlling mechanisms 
\cite{controlnet,mou2023t2i,liu2023more} emerged to guide the diffusion process for text-to-image generation. 
Particularly, ControlNet \cite{controlnet} enables to condition the generation process using edges, pose, semantic masks, image depths, etc. 
However, a direct application of ControlNet in the video domain leads to temporal inconsistencies and to severe changes of object appearance, identity, and the background (see Fig.~\ref{fig:abltion_study} and the appendix Figures \ref{fig:ablation-unconditional-attention-motion}, \ref{fig:table_of_edge_study}, \ref{fig:table_of_pose_study}). It turns out that our modifications on the basic diffusion process for videos result in more consistent videos guided by ControlNet conditions. We would like to point out again that our method does not require any fine-tuning or optimization processes.

More specifically, ControlNet creates a trainable copy of the encoder (including the middle blocks) of the UNet $\epsilon^t_\theta(x_t,\tau)$ while additionally taking the input $x_t$ and a condition $c$, and adds the outputs of each layer to the skip-connections of the original UNet. 
Here $c$ can be any type of condition, such as edge map, scribbles, pose (body landmarks), depth map, segmentation map, etc.
The trainable branch is being trained on a specific domain for each type of the condition $c$ resulting in an effective conditional text-to-image generation mechanism.

To guide our video generation process with ControlNet we apply our method to the basic diffusion process, i.e. enrich the latent codes $x^{1:m}_T$ with motion information and change the self-attentions into cross-frame attentions in the main UNet. 
While adopting the main UNet for video generation task, we apply the ControlNet pretrained copy branch per-frame on each $x^k_t$ for $k=1,\ldots,m$ in each diffusion time-step $t=T,\ldots,1$ and add the ControlNet branch outputs to the skip-connections of the main UNet.

\begin{figure}
    \centering
    \begin{subfigure}{.45\textwidth}
    \includegraphics[width=0.24\textwidth]{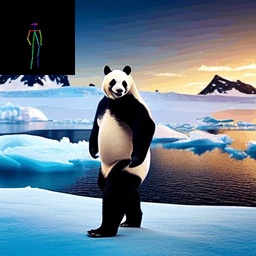} 
    \includegraphics[width=0.24\textwidth]{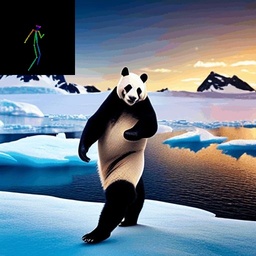} 
    \includegraphics[width=0.24\textwidth]{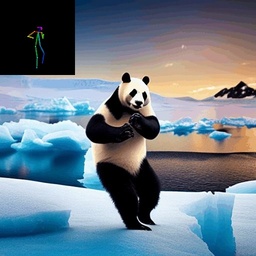} 
    \includegraphics[width=0.24\textwidth]{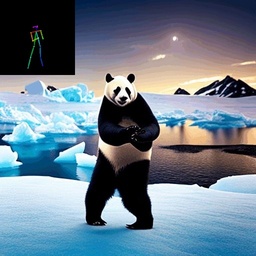}
    \vskip -.4\baselineskip
    \caption{a panda dancing in Antarctica}
    \end{subfigure}
    \vskip .3\baselineskip
    \begin{subfigure}{.45\textwidth}
    \includegraphics[width=0.24\textwidth]{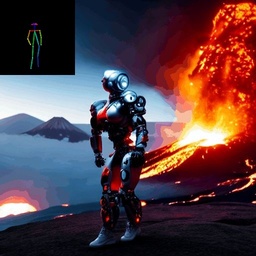} 
    \includegraphics[width=0.24\textwidth]{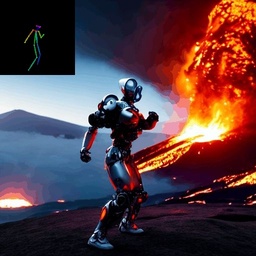} 
    \includegraphics[width=0.24\textwidth]{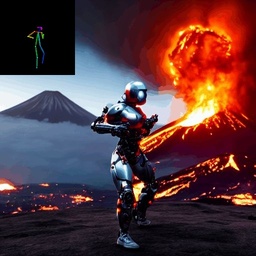} 
    \includegraphics[width=0.24\textwidth]{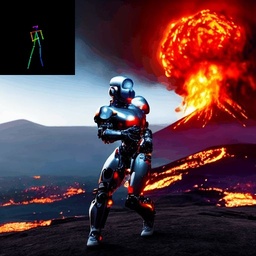}
    \vskip -.4\baselineskip
    \caption{a cyborg dancing near a volcano}
    \end{subfigure}
    \vskip .3\baselineskip
    \begin{subfigure}{.45\textwidth}
    \includegraphics[width=0.24\textwidth]{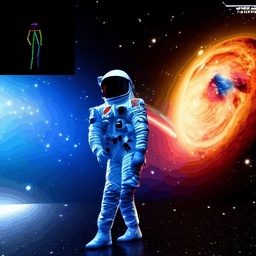} 
    \includegraphics[width=0.24\textwidth]{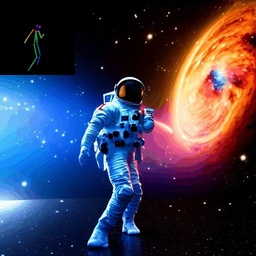} 
    \includegraphics[width=0.24\textwidth]{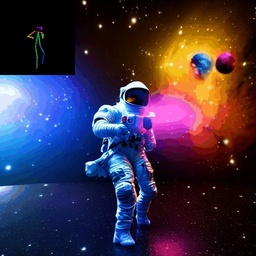} 
    \includegraphics[width=0.24\textwidth]{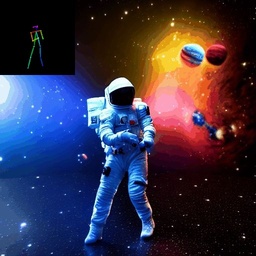} 
    \vskip -.4\baselineskip
    \caption{an astronaut dancing in the outer space}
    \end{subfigure}
    \vskip -.3\baselineskip
    \caption{Conditional generation with pose control. For more results see appendix, Sec. \ref{apendix:t2v_pose_guidance}.}
    \label{fig:poseControl}
\end{figure}


\begin{figure}
    \centering
    \begin{subfigure}{.45\textwidth}
    \includegraphics[width=0.24\textwidth]{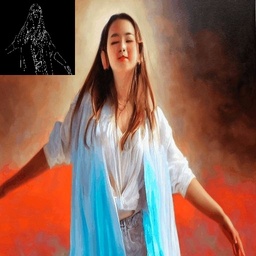} 
    \includegraphics[width=0.24\textwidth]{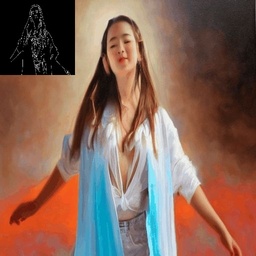} 
    \includegraphics[width=0.24\textwidth]{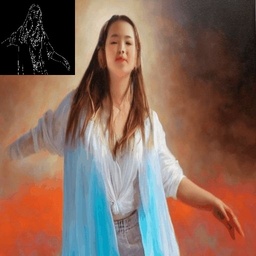} 
    \includegraphics[width=0.24\textwidth]{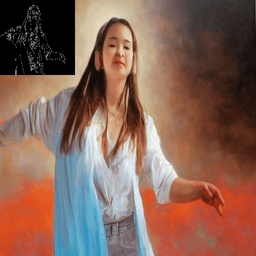}
    \vskip -.4\baselineskip
    \caption{oil painting of a girl dancing close-up}
    \end{subfigure}
    \vskip .3\baselineskip
    \begin{subfigure}{.45\textwidth}
    \includegraphics[width=0.24\textwidth]{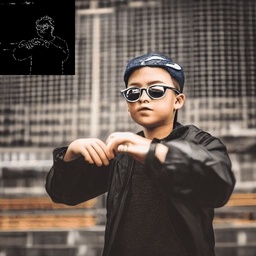} 
    \includegraphics[width=0.24\textwidth]{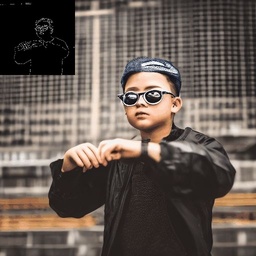} 
    \includegraphics[width=0.24\textwidth]{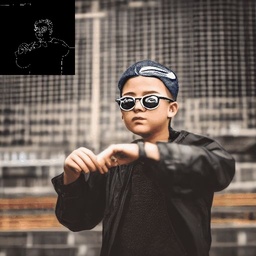} 
    \includegraphics[width=0.24\textwidth]{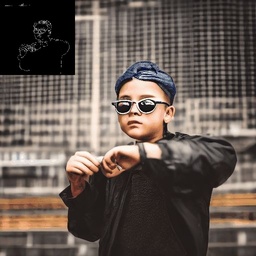}
    \vskip -.4\baselineskip
    \caption{Cyberpunk boy with a hat dancing close-up}
    \end{subfigure}
    \vskip -.3\baselineskip
    \caption{Conditional generation with edge control. For more results see appendix, Sec. \ref{apendix:t2v_edge_guidance}.}
    \label{fig:edgeControl}
\end{figure}
\vskip\baselineskip

\begin{figure}
    \centering
    \begin{subfigure}{.45\textwidth}
    \includegraphics[width=0.24\textwidth]{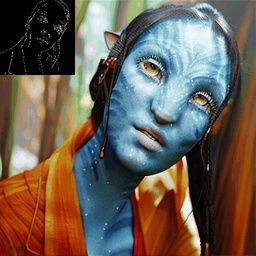} 
    \includegraphics[width=0.24\textwidth]{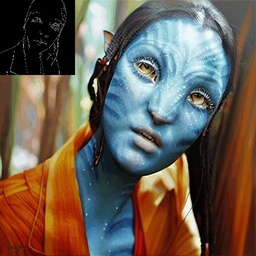} 
    \includegraphics[width=0.24\textwidth]{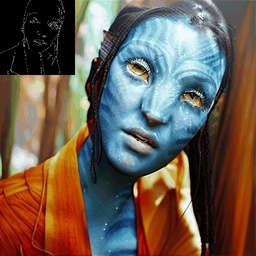} 
    \includegraphics[width=0.24\textwidth]{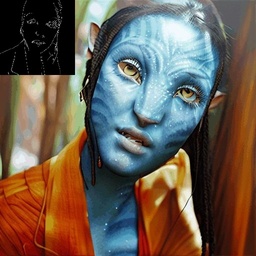}
    \vskip -.4\baselineskip
    \caption{oil painting of a beautiful girl avatar style}
    \end{subfigure}
    \vskip .3\baselineskip
    \begin{subfigure}{.45\textwidth}
    \includegraphics[width=0.24\textwidth]{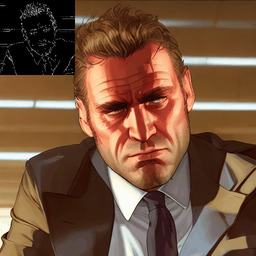} 
    \includegraphics[width=0.24\textwidth]{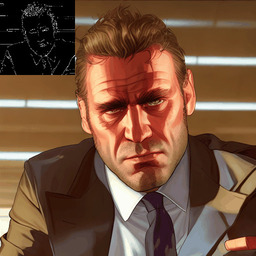} 
    \includegraphics[width=0.24\textwidth]{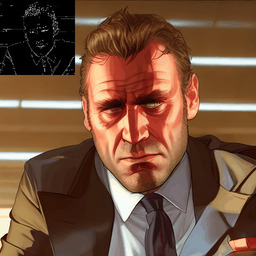} 
    \includegraphics[width=0.24\textwidth]{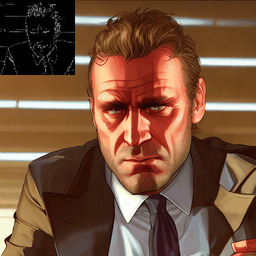}
    \vskip -.4\baselineskip
    \caption{gta-5 style}
    \end{subfigure}
    \vskip -.3\baselineskip
    \caption{Conditional generation with edge control and DB models.}
    \label{fig:edgeControl_with_DB}
\end{figure}

Furthermore, for our conditional generation task, we adopted the weights of specialized DreamBooth (DB) \cite{ruiz2022dreambooth} models\footnote{Avatar model: \url{https://civitai.com/models/9968/avatar-style}. GTA-5 model:  \url{https://civitai.com/models/1309/gta5-artwork-diffusion}.}. This gives us specialized time-consistent video generations (see Fig.~\ref{fig:edgeControl_with_DB}).



\subsection{Video Instruct-Pix2Pix}
With the rise of text-guided image editing methods such as Prompt2Prompt \cite{prompt2prompt}, Instruct-Pix2Pix \cite{brooks2022instructpix2pix}, SDEdit \cite{meng2021sdedit}, etc., text-guided video editing approaches emerged \cite{bar2022text2live,lee2023shape,tune-a-video}.
While these methods require complex optimization processes, our approach enables the adoption of any SD-based text-guided image editing algorithm to the video domain without any training or fine-tuning. 
Here we take the text-guided image editing method Instruct-Pix2Pix and combine it with our approach. 
More precisely, we change the self-attention mechanisms in Instruct-Pix2Pix to cross-frame attentions according to Eq.~\ref{eq:cross-frame-attn}. 
Our experiments show that this adaptation significantly improves the consistency of the edited videos (see Fig.~\ref{fig:videoediting}) over the na\"ive per-frame usage of Instruct-Pix2Pix.

\section{Experiments}
\subsection{Implementation Details}
We take the Stable Diffusion \cite{stable_diff} code\footnote{ \url{https://github.com/huggingface/diffusers}. We also benefit from the codebase of Tune-A-Video \url{https://github.com/showlab/Tune-A-Video}.} with its pre-trained weights from version 1.5 as basis and implement our modifications. 
%
In our experiments, we generate $m=8$ frames with $512\times 512$ resolution for each video. However, our framework allows generating any number of frames, either by increasing $m$, or by employing our method in an auto-regressive fashion where the last generated frame $m$ becomes the first frame in computing the next $m$ frames. For text-to-video generation, we take $T' = 881, T = 941$, while for conditional and specialized generation, and for Video Instruct-Pix2Pix, we take $T'=T=1000$.


For a conditional generation, we use the codebase\footnote{ \url{https://github.com/lllyasviel/ControlNet}.} of ControlNet \cite{controlnet}. For specialized models, we take DreamBooth \cite{ruiz2022dreambooth} models from publicly available sources. For Video Instruct-Pix2Pix, we use the codebase\footnote{ \url{https://github.com/timothybrooks/instruct-pix2pix}.} of Instruct Pix2Pix \cite{brooks2022instructpix2pix}.

\subsection{Qualitative Results}

All applications of Text2Video-Zero show that it successfully generates videos where the global scene and the background are time consistent and the context, appearance, and identity of the foreground object are maintained throughout the entire sequence. 

In the case of text-to-video, we observe that it generates high-quality videos that are well-aligned to the text prompt (see Fig.~\ref{fig:text2video_results} and the appendix). For instance, the depicted panda shows a naturally walking on the street. Likewise, using additional guidance from edges or poses (see Fig.~\ref{fig:poseControl}, Fig,~\ref{fig:edgeControl} and Fig.~\ref{fig:edgeControl_with_DB} and the appendix), high quality videos are generated matching the prompt and the guidance that show great temporal consistency and identity preservation. 

In the case of Video Instruct-Pix2Pix (see Fig. \ref{fig:teaser_img} and the appendix) generated videos possess high-fidelity with respect to the input video, while following closely the instruction.

\subsection{Comparison with Baselines}
We compare our method with two publicly available baselines: CogVideo \cite{hong2022cogvideo} and Tune-A-Video \cite{tune-a-video}. 
Since CogVideo is a text-to-video method we compare with it in pure text-guided video synthesis settings. 
With Tune-A-Video we compare in our Video Instruct-Pix2Pix setting.

\begin{figure*}
    \centering
    \begin{subfigure}{\textwidth}
    \centering
    \includegraphics[width=0.115\textwidth]{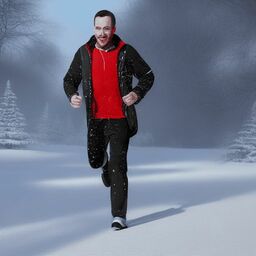} 
    \includegraphics[width=0.115\textwidth]{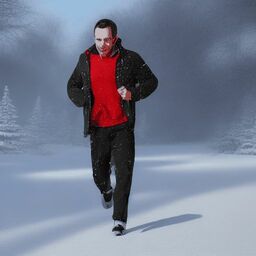} 
    \includegraphics[width=0.115\textwidth]{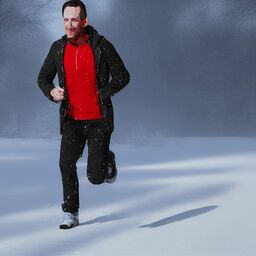} 
    \includegraphics[width=0.115\textwidth]{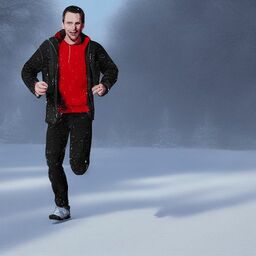} 
    \hskip\baselineskip
    \includegraphics[width=0.115\textwidth]{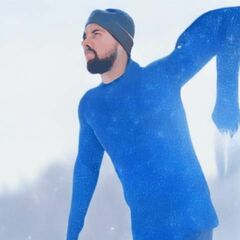} 
    \includegraphics[width=0.115\textwidth]{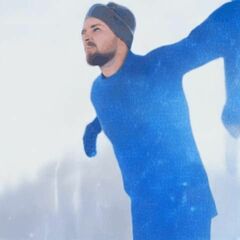} 
    \includegraphics[width=0.115\textwidth]{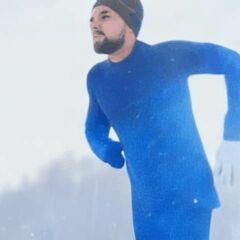} 
    \includegraphics[width=0.115\textwidth]{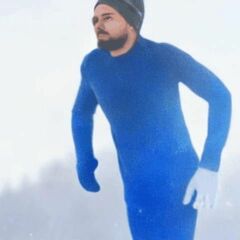} 
    \vskip -.4\baselineskip
    \caption{a man is running in the snow}
    \end{subfigure}
    \vskip .3\baselineskip
    \begin{subfigure}{\textwidth}
    \centering
    \includegraphics[width=0.115\textwidth]{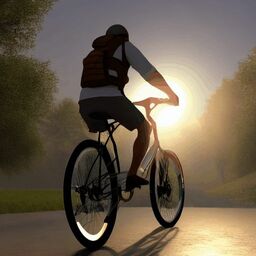} 
    \includegraphics[width=0.115\textwidth]{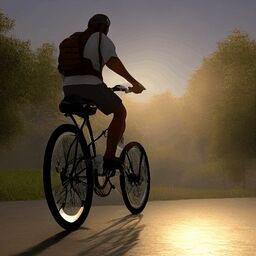} 
    \includegraphics[width=0.115\textwidth]{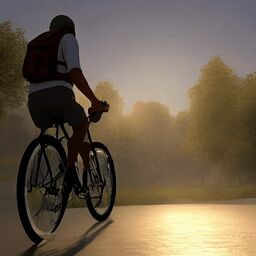} 
    \includegraphics[width=0.115\textwidth]{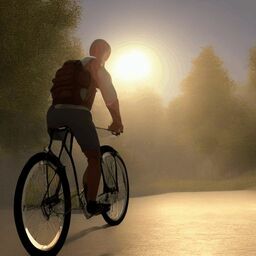} 
    \hskip\baselineskip
    \includegraphics[width=0.115\textwidth]{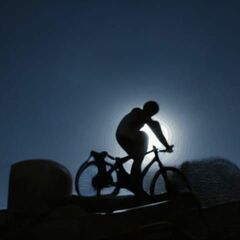} 
    \includegraphics[width=0.115\textwidth]{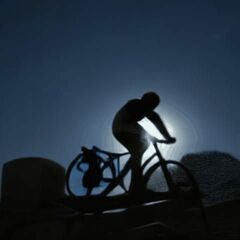} 
    \includegraphics[width=0.115\textwidth]{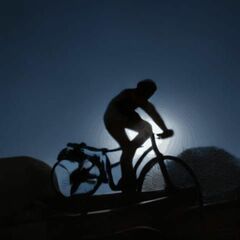} 
    \includegraphics[width=0.115\textwidth]{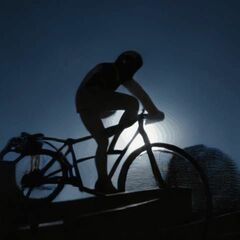} 
    \vskip -.4\baselineskip
    \caption{a man is riding a bicycle in the sunshine}
    \end{subfigure}
    \vskip .3\baselineskip

    \begin{subfigure}{\textwidth}
    \centering
    \includegraphics[width=0.115\textwidth]{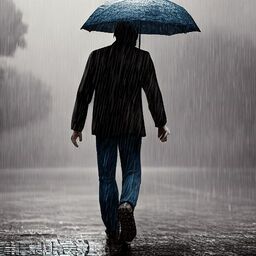} 
    \includegraphics[width=0.115\textwidth]{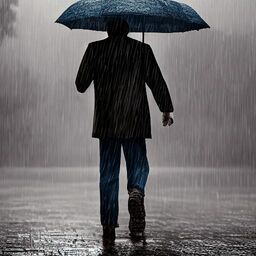} 
    \includegraphics[width=0.115\textwidth]{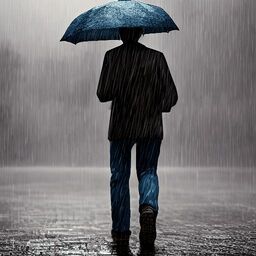} 
    \includegraphics[width=0.115\textwidth]{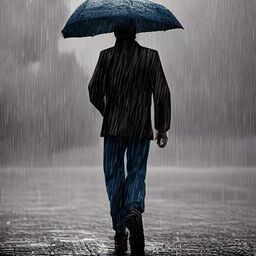} 
    \hskip\baselineskip
    \includegraphics[width=0.115\textwidth]{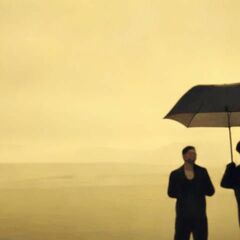} 
    \includegraphics[width=0.115\textwidth]{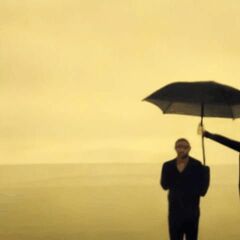} 
    \includegraphics[width=0.115\textwidth]{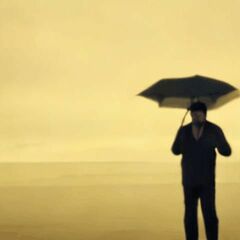} 
    \includegraphics[width=0.115\textwidth]{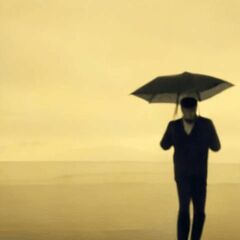} 
    \vskip -.4\baselineskip
    \caption{a man is walking in the rain}
    \end{subfigure}
    \vskip -.3\baselineskip
    \caption{Comparison of our method vs CogVideo on text-to-video generation task (left is ours, right is CogVideo \cite{hong2022cogvideo}). For more comparisons, see appendix Fig. \ref{fig:unconstrained-text2video-ours-vs-cog}.}
    \label{fig:ourVSCogVideo}
\end{figure*}


\subsubsection{Quantitative Comparison}



To show quantitative results, we evaluate the CLIP score \cite{hessel2021clipscore}, which indicates video-text alignment. We randomly take 25 videos generated by CogVideo and synthesize corresponding videos using the same prompts according to our method. The CLIP scores for our method and CogVideo are $31.19$ and $29.63$, respectively. Our method thus slightly outperforms  CogVideo, even though the latter has 9.4 billion parameters and requires large-scale training on videos.


\subsubsection{Qualitative Comparison}
We present several results of our method in Fig.~\ref{fig:ourVSCogVideo} and provide a qualitative comparison to CogVideo \cite{hong2022cogvideo}. Both methods show good temporal consistency throughout the sequence, preserving the identity of the object and background. However, our method shows better text-video alignment. For instance, while our method correctly generates a video of a man riding a bicycle in the sunshine in Fig.~\ref{fig:ourVSCogVideo}(b), CogVideo sets the background to moon light. Also in Fig.~\ref{fig:ourVSCogVideo}(a), our method correctly shows a man running in the snow, while neither the snow nor a man running are clearly visible in the video generated by CogVideo.


Qualitative results of \textit{Video Instruct-Pix2Pix} and a visual comparison with per-frame Instruct-Pix2Pix and Tune-A-Video are shown in Fig.~\ref{fig:videoediting}. While Instruct-Pix2Pix shows a good editing performance per frame, it lacks temporal consistency. This becomes evident especially in the video depicting a skiing person, where the snow and the sky are drawn using different styles and colors. Using our Video Instruct-Pix2Pix method, these issues are solved resulting in temporally consistent video edits throughout the entire sequence.


While Tune-A-Video creates temporally consistent video generations, it is less aligned to the instruction guidance than our method, struggles creating local edits and losses details of the input sequence. 
This becomes apparent when looking at the edit of the dancer video depicted in Fig.~\ref{fig:videoediting}  (left side). In contrast to Tune-A-Video, our method draws the entire dress brighter and at the same time better preserves the background, \eg the wall behind the dancer is almost kept the same. Tune-A-Video draws a severely modified wall. Moreover, our method is more faithful to the input details,  \eg, Video Instruct-Pix2Pix draws the dancer using the pose exactly as provided (Fig.~\ref{fig:videoediting}  left), and shows all skiing persons appearing in the input video (compare last frame of Fig.~\ref{fig:videoediting}(right)), in constrast to Tune-A-Video.
All the above-mentioned weaknesses of Tune-A-Video can also be observed in our additional evaluations that are provided in the appendix, Figures \ref{fig:table_of_EDITING/comparison1}, \ref{fig:table_of_EDITING/comparison2}.




\setcounter{table}{\value{figure}} 
\begin{table*}
\captionsetup{name=Figure}
        \centering
        \begin{tabular}{ M{13mm}M{15mm}M{15mm}M{15mm}M{16mm}M{15mm}M{15mm}M{15mm}M{15mm}M{15mm}}
            \multicolumn{9}{M{74mm}}{} \\
            \begin{flushleft}
                \fontsize{7}{12}\selectfont Original
            \end{flushleft} &  {\includegraphics[width=.1\textwidth]{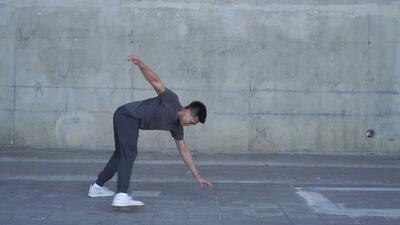}} &  {\includegraphics[width=.1\textwidth]{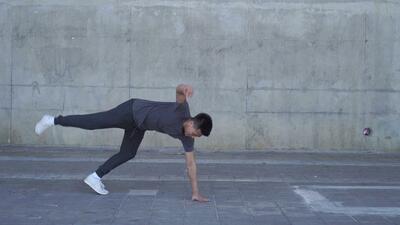}} &  {\includegraphics[width=.1\textwidth]{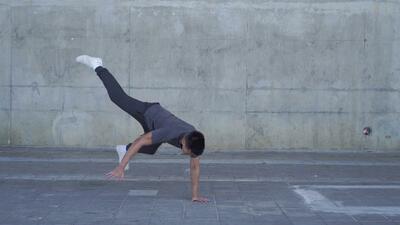}} &  {\includegraphics[width=.1\textwidth]{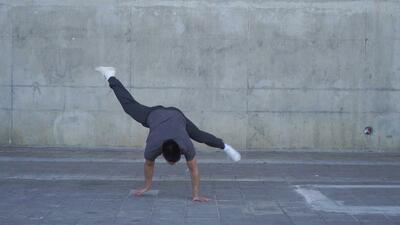}} &  
            {\includegraphics[width=.1\textwidth]{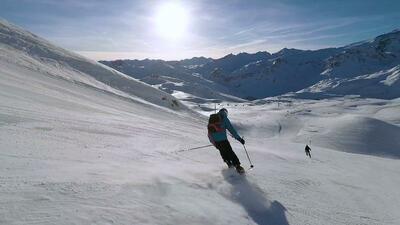}} &  {\includegraphics[width=.1\textwidth]{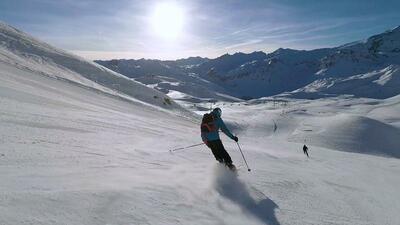}} &  {\includegraphics[width=.1\textwidth]{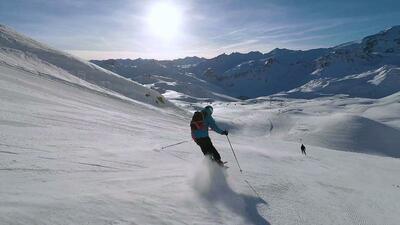}} & 
            {\includegraphics[width=.1\textwidth]{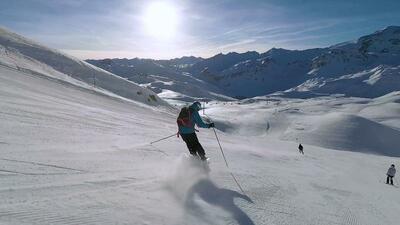}} \\ 
            \multicolumn{9}{M{74mm}}{} \\
            & \multicolumn{4}{M{65mm}}{color his dress white} & \multicolumn{4}{M{65mm}}{make it Van Gogh Starry Night style} \\
            \begin{flushleft}\vspace{0.1cm}\fontsize{7}{12}\selectfont Video Instruct-Pix2Pix (Ours) \end{flushleft} &  {\includegraphics[width=.1\textwidth]{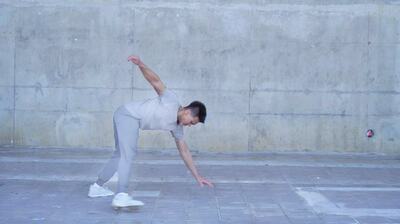}} &  {\includegraphics[width=.1\textwidth]{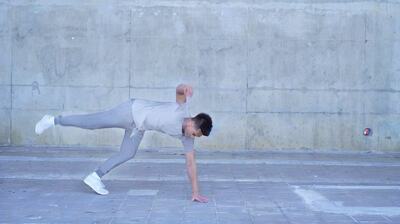}} &  {\includegraphics[width=.1\textwidth]{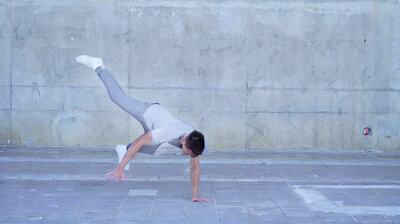}} &  {\includegraphics[width=.1\textwidth]{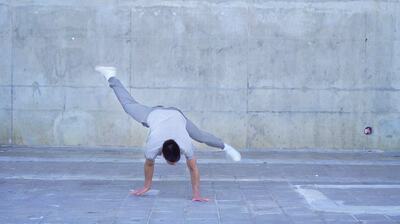}} &  
            {\includegraphics[width=.1\textwidth]{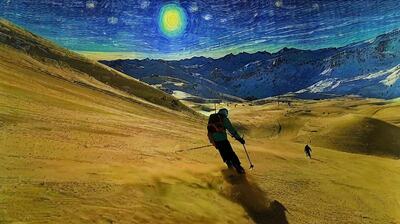}} &  {\includegraphics[width=.1\textwidth]{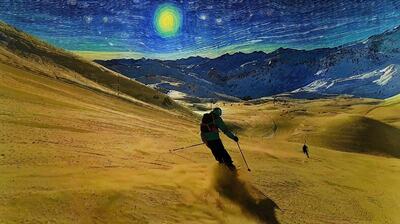}} &  {\includegraphics[width=.1\textwidth]{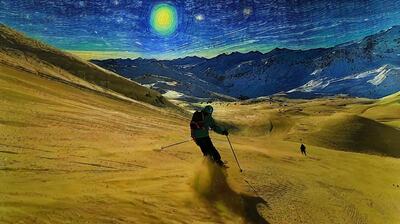}} & 
            {\includegraphics[width=.1\textwidth]{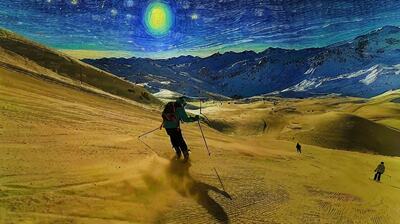}} \\  
            \begin{flushleft}
              \vspace{-0.2cm}  \fontsize{7}{12}\selectfont \mbox{Instruct-Pix2Pix}
            \end{flushleft} &  {\includegraphics[width=.1\textwidth]{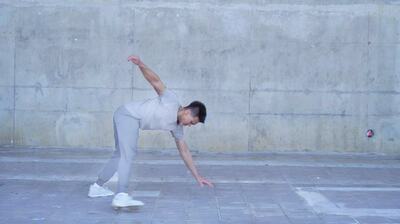}} &  {\includegraphics[width=.1\textwidth]{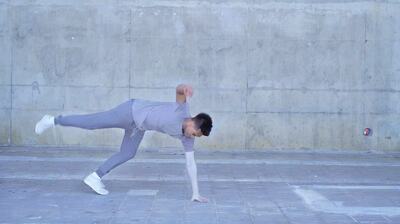}} &  {\includegraphics[width=.1\textwidth]{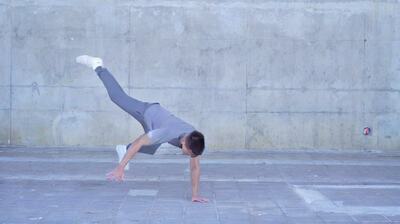}} &  {\includegraphics[width=.1\textwidth]{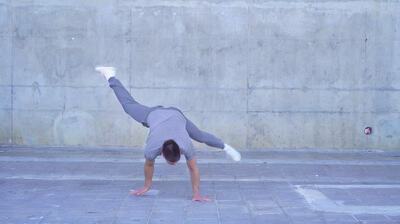}} &  
            {\includegraphics[width=.1\textwidth]{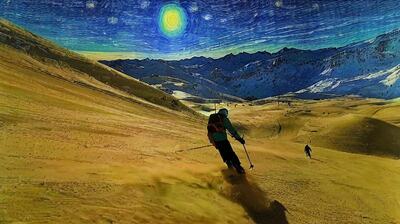}} &  {\includegraphics[width=.1\textwidth]{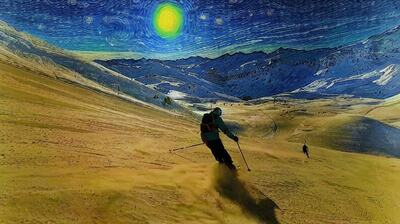}} &  {\includegraphics[width=.1\textwidth]{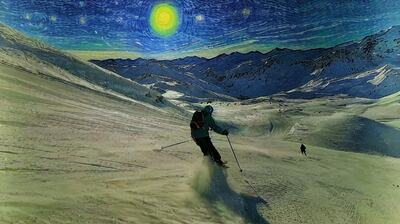}} & 
            {\includegraphics[width=.1\textwidth]{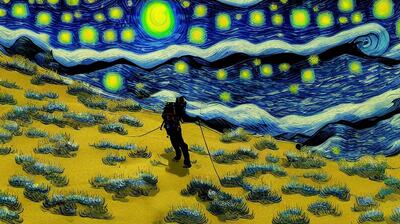}} \\
            \begin{flushleft}
               \vspace{-0.2cm} \fontsize{7}{12}\selectfont \mbox{Tune-A-Video}
            \end{flushleft} &  {\includegraphics[width=.1\textwidth]{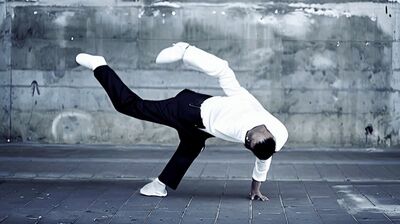}} &  {\includegraphics[width=.1\textwidth]{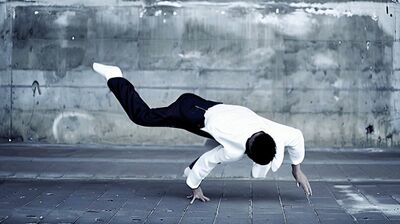}} &  {\includegraphics[width=.1\textwidth]{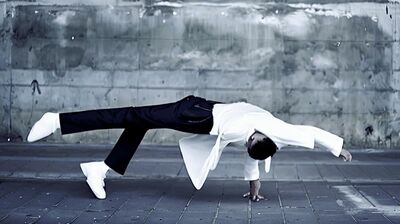}} &  {\includegraphics[width=.1\textwidth]{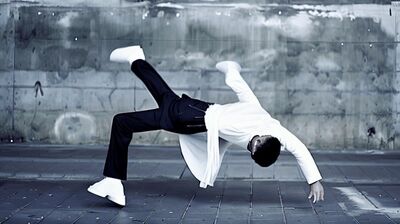}} &  
            {\includegraphics[width=.1\textwidth]{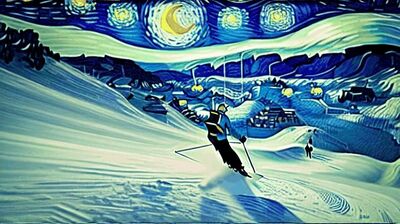}} &  {\includegraphics[width=.1\textwidth]{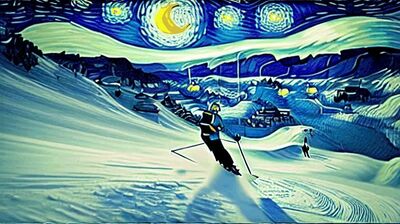}} &  {\includegraphics[width=.1\textwidth]{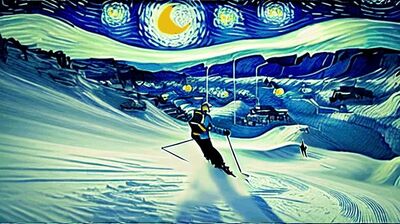}} & 
            {\includegraphics[width=.1\textwidth]{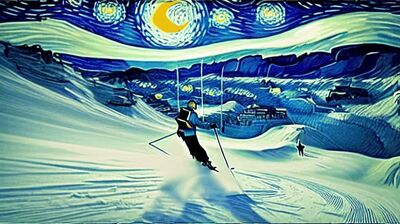}} \\
        \end{tabular}
        \caption{Comparison of Video Instruct-Pix2Pix(ours) with Tune-A-Video and per-frame Instruct-Pix2Pix. For more comparisons see our appendix.}
        \label{fig:videoediting}
    \end{table*}
\setcounter{figure}{\value{table}} 


\subsection{Ablation Study}
We perform an ablation study on two main components of our method: making the initial latent codes coherent to a motion, and using cross-frame attention on the first frame instead of self-attention (for an ablation study on background smoothing see appendix Sec. \ref{apendix:unconstrained-text2video_ablation}). 
The qualitative results are presented in Fig.~\ref{fig:abltion_study}. With the base model only, i.e. without our changes (first row), no temporal consistency is achieved.  
This is especially severe for unconstrained text-to-video generations. For example, the appearance and position of the horse changes very quickly, and the background is utterly inconsistent. Using our proposed motion dynamics (second row), the general concept of the video is preserved better throughout the sequence. For example, all frames show a close-up of a horse in motion.  Likewise, the appearance of the woman and the background in the middle four figures  (using ControlNet with edge guidance) is greatly improved. 

Using our proposed cross frame attention (third row), we see across all generations improved preservation of the object identities and their appearances. 
Finally, by combining both concepts (last row), we achieve the best temporal coherence.
%
For instance, we see the same background motifs and also about object identity preservation in the last four columns and at the same time a natural transition between the generated images.

\setcounter{table}{\value{figure}} 
\begin{table*}
\captionsetup{name=Figure}
        \centering
        \begin{tabular}{ M{23mm}M{7.4mm}M{7.4mm}M{7.4mm}M{9mm}M{7.4mm}M{7.4mm}M{7.4mm}M{9mm}M{7.4mm}M{7.4mm}M{7.4mm}M{9mm}}
            \multicolumn{13}{M{100mm}}{} \\
            & & & & & {\includegraphics[width=.06\textwidth]{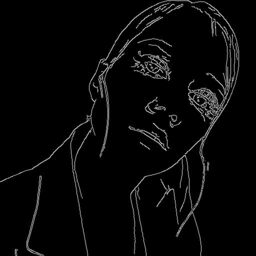}} & {\includegraphics[width=.06\textwidth]{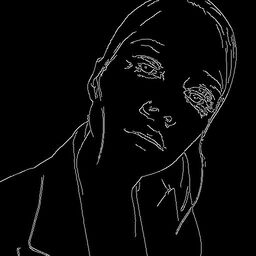}} & {\includegraphics[width=.06\textwidth]{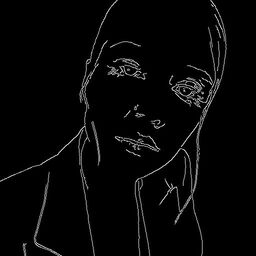}} & {\includegraphics[width=.06\textwidth]{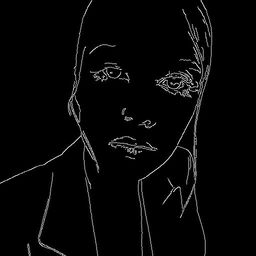}} & {\includegraphics[width=.06\textwidth]{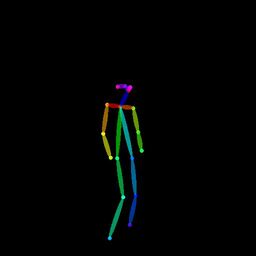}} & {\includegraphics[width=.06\textwidth]{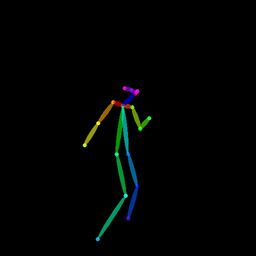}} & {\includegraphics[width=.06\textwidth]{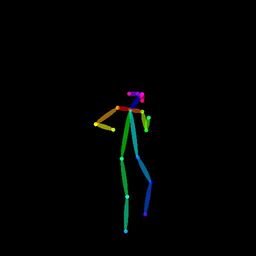}} & {\includegraphics[width=.06\textwidth]{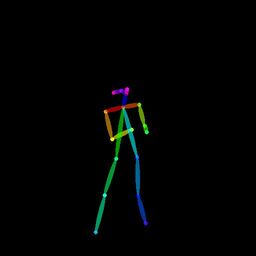}} \\

            \makecell{
              {\fontsize{7}{8}\selectfont \raisebox{0.5ex}{No Motion in Latents}} \\[-2pt]
              {\fontsize{7}{8}\selectfont \setlength{\baselineskip}{0.5ex} No Cross-Frame Attention}
            } & {\includegraphics[width=.06\textwidth]{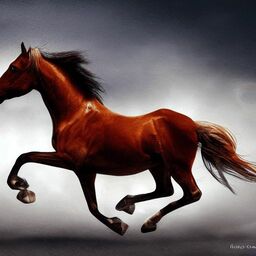}} & {\includegraphics[width=.06\textwidth]{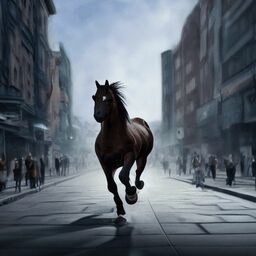}} & {\includegraphics[width=.06\textwidth]{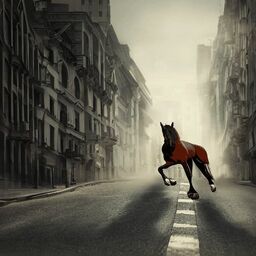}} & {\includegraphics[width=.06\textwidth]{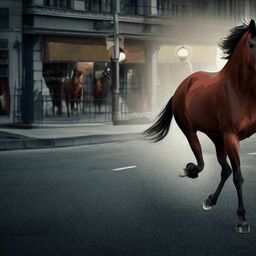}} &{\includegraphics[width=.06\textwidth]{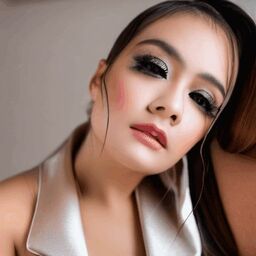}} & {\includegraphics[width=.06\textwidth]{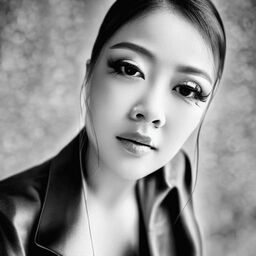}} & {\includegraphics[width=.06\textwidth]{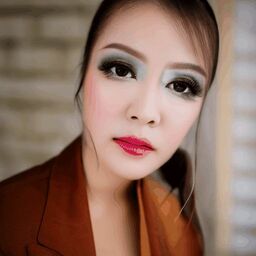}} & {\includegraphics[width=.06\textwidth]{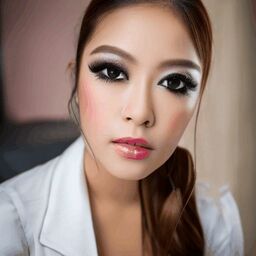}} & {\includegraphics[width=.06\textwidth]{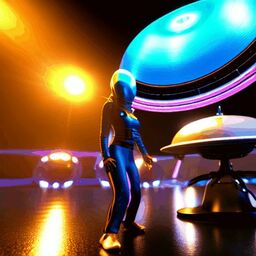}} & {\includegraphics[width=.06\textwidth]{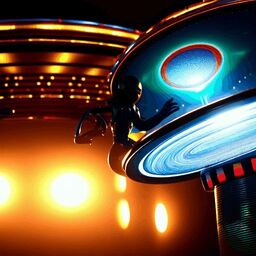}} & {\includegraphics[width=.06\textwidth]{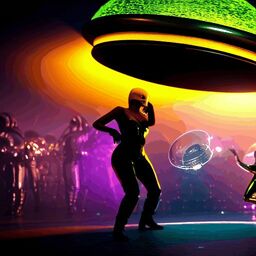}} & {\includegraphics[width=.06\textwidth]{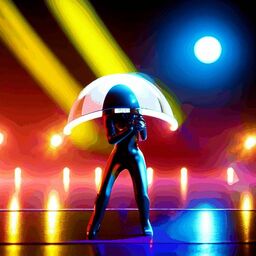}} \\

            \makecell{
              {\fontsize{7}{8}\selectfont \raisebox{0.5ex}{Motion in Latents}} \\[-2pt]
              {\fontsize{7}{8}\selectfont \setlength{\baselineskip}{0.5ex} No Cross-Frame Attention}
            } & {\includegraphics[width=.06\textwidth]{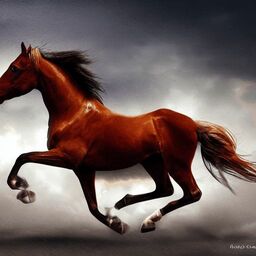}} & {\includegraphics[width=.06\textwidth]{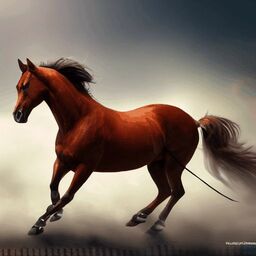}} & {\includegraphics[width=.06\textwidth]{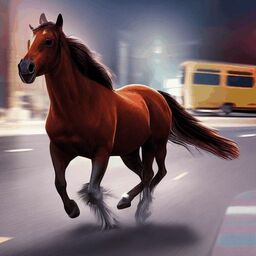}} & {\includegraphics[width=.06\textwidth]{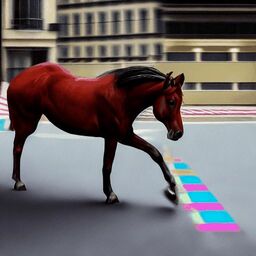}} &{\includegraphics[width=.06\textwidth]{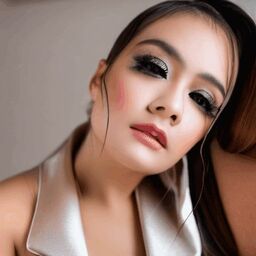}} & {\includegraphics[width=.06\textwidth]{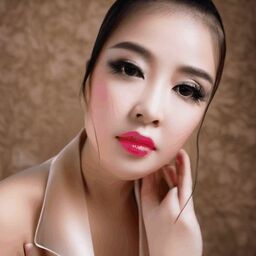}} & {\includegraphics[width=.06\textwidth]{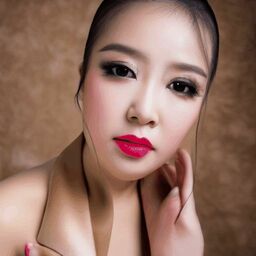}} & {\includegraphics[width=.06\textwidth]{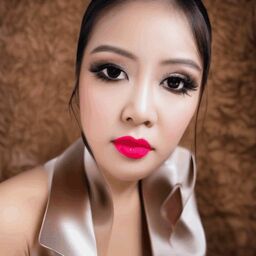}} & {\includegraphics[width=.06\textwidth]{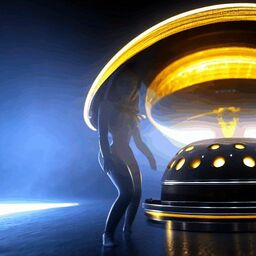}} & {\includegraphics[width=.06\textwidth]{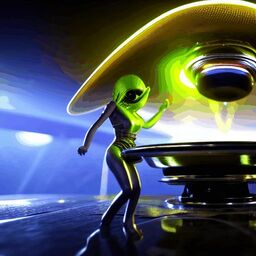}} & {\includegraphics[width=.06\textwidth]{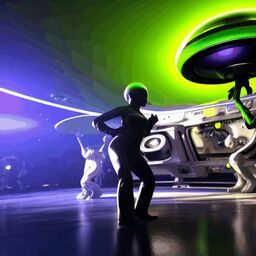}} & {\includegraphics[width=.06\textwidth]{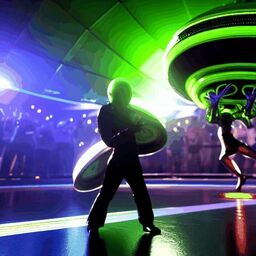}} \\            
            \makecell{
              {\fontsize{7}{8}\selectfont \raisebox{0.5ex}{No Motion in Latents}} \\[-2pt]
              {\fontsize{7}{8}\selectfont \setlength{\baselineskip}{0.5ex} Cross-Frame Attention}
            } & {\includegraphics[width=.06\textwidth]{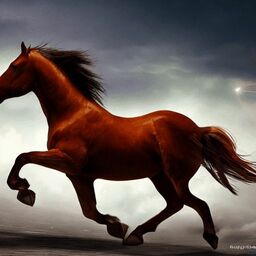}} & {\includegraphics[width=.06\textwidth]{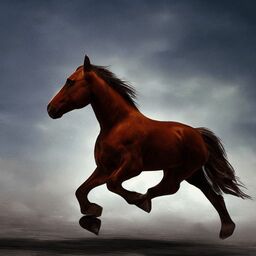}} & {\includegraphics[width=.06\textwidth]{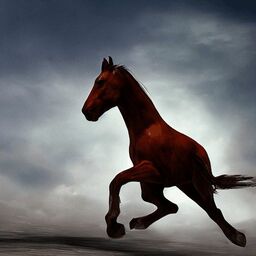}} & {\includegraphics[width=.06\textwidth]{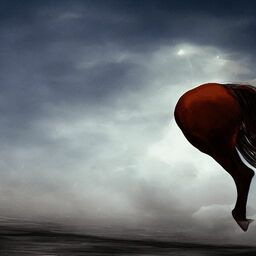}} & {\includegraphics[width=.06\textwidth]{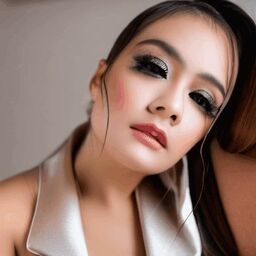}} & {\includegraphics[width=.06\textwidth]{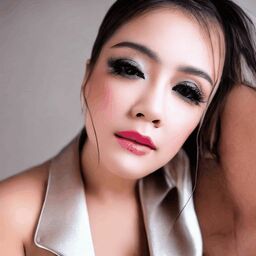}} & {\includegraphics[width=.06\textwidth]{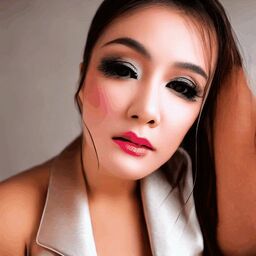}} & {\includegraphics[width=.06\textwidth]{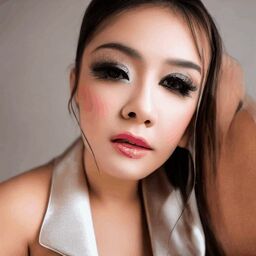}} & {\includegraphics[width=.06\textwidth]{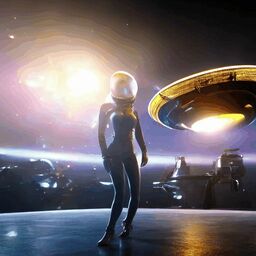}} & {\includegraphics[width=.06\textwidth]{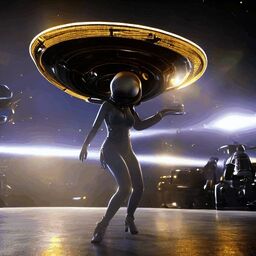}} & {\includegraphics[width=.06\textwidth]{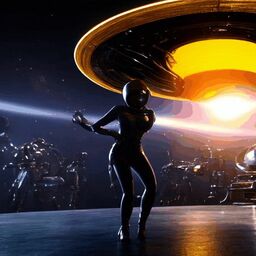}} & {\includegraphics[width=.06\textwidth]{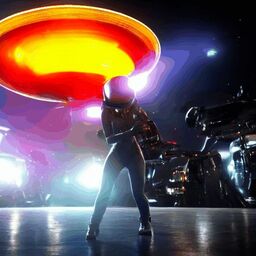}} \\
            \makecell{
              {\fontsize{7}{8}\selectfont \raisebox{0.5ex}{Motion in Latents}} \\[-2pt]
              {\fontsize{7}{8}\selectfont \setlength{\baselineskip}{0.5ex} Cross-Frame Attention}
            } & {\includegraphics[width=.06\textwidth]{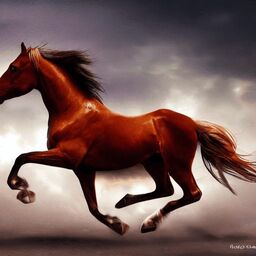}} & {\includegraphics[width=.06\textwidth]{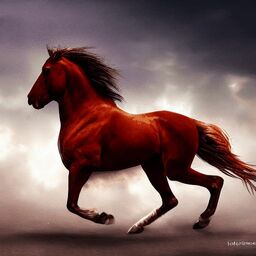}} & {\includegraphics[width=.06\textwidth]{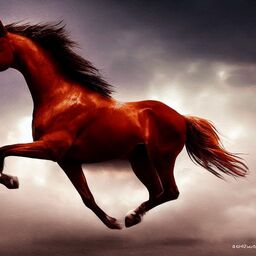}} & {\includegraphics[width=.06\textwidth]{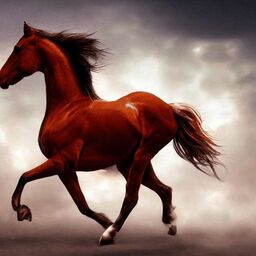}} &{\includegraphics[width=.06\textwidth]{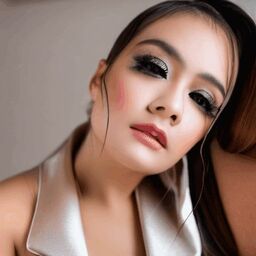}} & {\includegraphics[width=.06\textwidth]{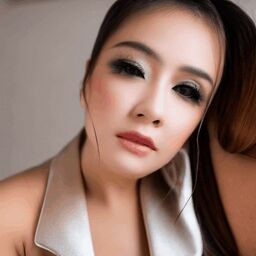}} & {\includegraphics[width=.06\textwidth]{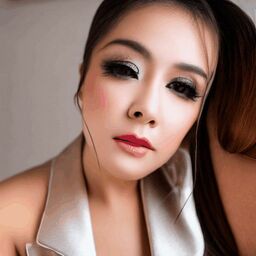}} & {\includegraphics[width=.06\textwidth]{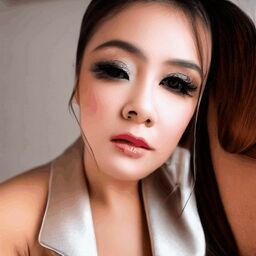}} & {\includegraphics[width=.06\textwidth]{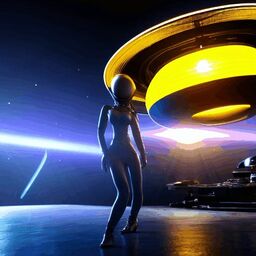}} & {\includegraphics[width=.06\textwidth]{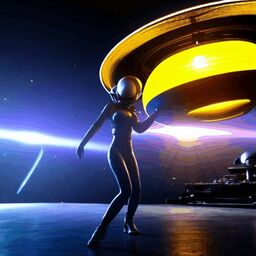}} & {\includegraphics[width=.06\textwidth]{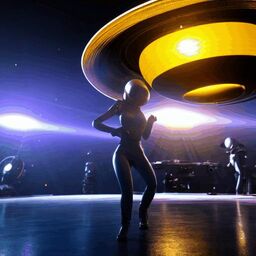}} & {\includegraphics[width=.06\textwidth]{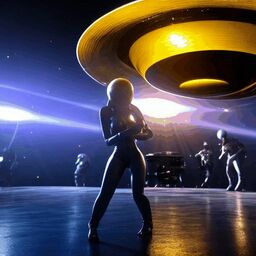}} \\                      
        \end{tabular}
        \caption{Ablation study showing the effect of our proposed components for text-to-video and text-guided video editing. Additional ablation study results are provided in the appendix.}
        \label{fig:abltion_study}
    \end{table*}
\setcounter{figure}{\value{table}} 


\section{Conclusion}

    In this paper, we addressed the problem of zero-shot text-to-video synthesis and proposed a novel method for time-consistent video generation. Our approach does not require any optimization or fine-tuning, making text-to-video generation and its applications affordable for everyone. We demonstrated the effectiveness of our method for various applications, including conditional and specialized video generation, and Video Instruct-Pix2Pix, \ie, instruction-guided video editing. Our contributions to the field include presenting a new problem of zero-shot text-to-video synthesis, showing the utilization of text-to-image diffusion models for generating time-consistent videos, and providing evidence of the effectiveness of our method for various video synthesis applications. We believe that our proposed method will open up new possibilities for video generation and editing, making it accessible and affordable for everyone.

{\small
\bibliographystyle{ieee_fullname}
\bibliography{egbib}
}

\newpage
\clearpage


\section*{\Large Appendix}

This supplementary material provides additional results to show the quality of our text-to-video generation method and its applications, and the importance of individual parts of our approach.

The quality of our text-to-video method without additional conditioning or specialization is investigated further in Sec.~\ref{apendix:unconstrained-text2video}. To this end, qualitative results are presented and compared to the only publicly available state-of-the-art competitor CogVideo \cite{hong2022cogvideo}. In order to analyze the relevance of our proposed procedures, several ablation studies are performed qualitatively.

Sec.~\ref{apendix:t2v_edge_guidance} supplements our paper by  elaborating results for conditional text-to-video generation guided by pose information. 
In Sec.~\ref{apendix:t2v_pose_guidance} we discuss more results of conditional text-to-video generation guided by edge information.
Qualitative results and extensive ablation studies are presented. 

Finally, Sec.~\ref{apendix:t2v_editing} provides additional qualitative results and more comparison with a recent state-of-the art method Tune-A-Video \cite{tune-a-video} for the instruction-guided video editing task and compares to our Video Instruct-Pix2Pix method.

\input{supplemental_chapters/text2video_no_guidance}
\input{supplemental_chapters/text2video_edge_guidance}
\input{supplemental_chapters/text2video_edge_db_guidance}
\input{supplemental_chapters/text2video_pose_guidance}
\input{supplemental_chapters/text2video_guidance_ablation}
\input{supplemental_chapters/text2video_editing}
\input{supplemental_chapters/text2video_editing_comparison}

\end{document}

%% file: supplemental_chapters/text2video_no_guidance.tex
\section{Additional Experiments  for Text-to-Video Unconditional Generation}
\label{apendix:unconstrained-text2video}

\subsection{Qualitative results}
We provide additional qualitative results of our method in the setting of text-to-video unconditional synthesis.
For high-quality generation, we append to each prompt presented in subsequent figures the suffix "high quality, HD, 8K, trending on artstation, high focus". 

\begin{figure*}
    \centering
    \begin{subfigure}{\textwidth}
    \includegraphics[width=0.12\textwidth]{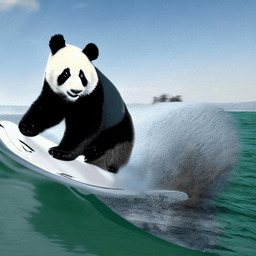} 
    \hfill
    \includegraphics[width=0.12\textwidth]{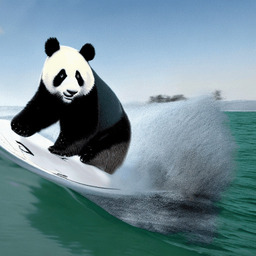} 
    \hfill
    \includegraphics[width=0.12\textwidth]{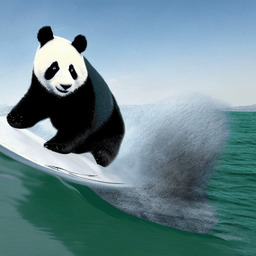} 
    \hfill
    \includegraphics[width=0.12\textwidth]{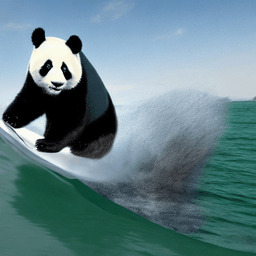} 
    \hfill
    \includegraphics[width=0.12\textwidth]{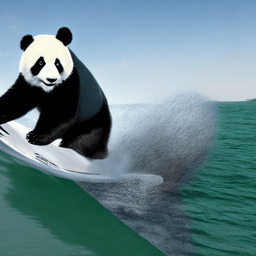} 
    \hfill
    \includegraphics[width=0.12\textwidth]{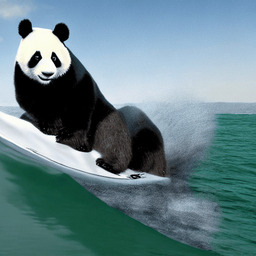} 
    \hfill
    \includegraphics[width=0.12\textwidth]{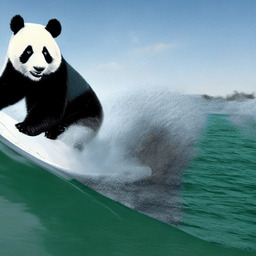} 
    \hfill
    \includegraphics[width=0.12\textwidth]{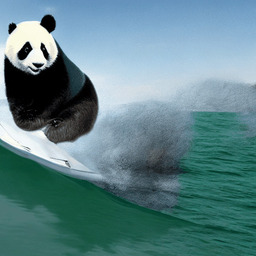} 
    \caption{a high quality realistic photo of a panda surfing on a wakeboard}
    \end{subfigure}
    \vskip\baselineskip
    \begin{subfigure}{\textwidth}
    \includegraphics[width=0.12\textwidth]{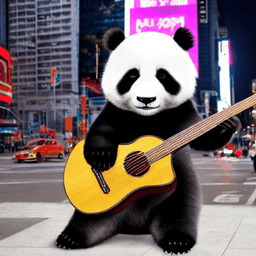} 
    \hfill
    \includegraphics[width=0.12\textwidth]{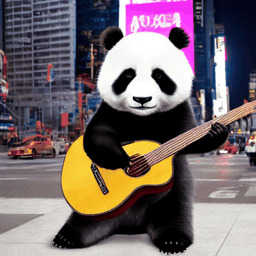} 
    \hfill
    \includegraphics[width=0.12\textwidth]{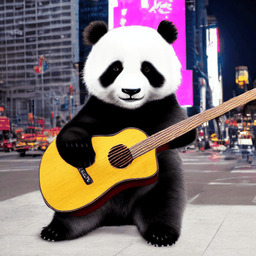} 
    \hfill
    \includegraphics[width=0.12\textwidth]{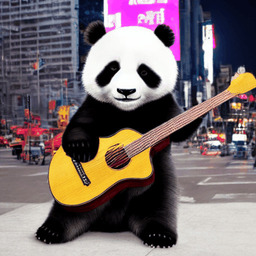} 
    \hfill
    \includegraphics[width=0.12\textwidth]{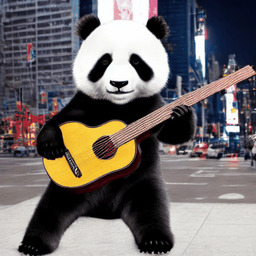} 
    \hfill
    \includegraphics[width=0.12\textwidth]{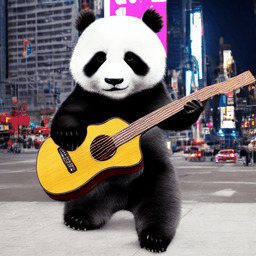} 
    \hfill
    \includegraphics[width=0.12\textwidth]{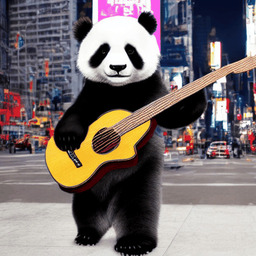} 
    \hfill
    \includegraphics[width=0.12\textwidth]{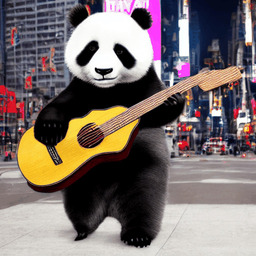} 
    \caption{a high quality realistic photo of a panda playing guitar on times square}
    \end{subfigure}
    \vskip\baselineskip
    \begin{subfigure}{\textwidth}
    \includegraphics[width=0.12\textwidth]{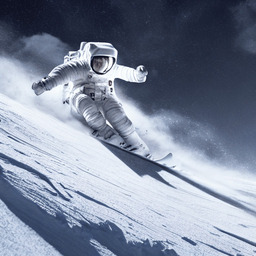} 
    \hfill
    \includegraphics[width=0.12\textwidth]{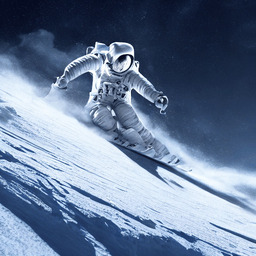} 
    \hfill
    \includegraphics[width=0.12\textwidth]{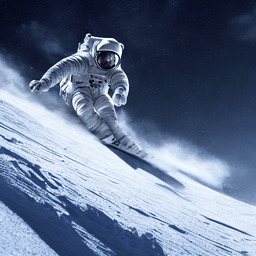} 
    \hfill
    \includegraphics[width=0.12\textwidth]{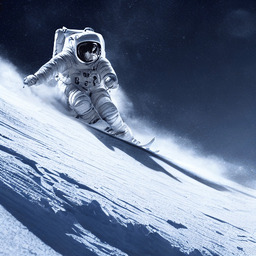} 
    \hfill
    \includegraphics[width=0.12\textwidth]{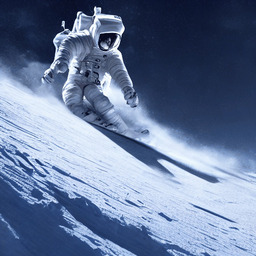} 
    \hfill
    \includegraphics[width=0.12\textwidth]{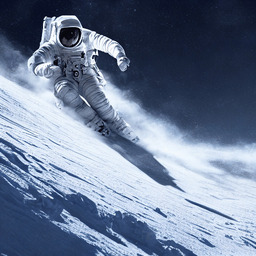} 
    \hfill
    \includegraphics[width=0.12\textwidth]{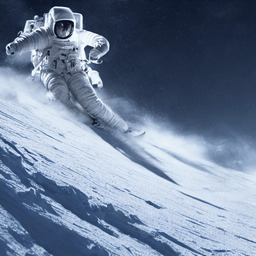} 
    \hfill
    \includegraphics[width=0.12\textwidth]{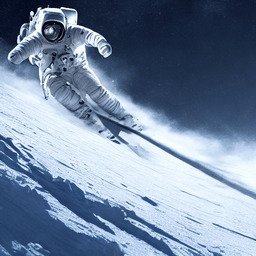} 
    \caption{an astronaut is skiing down a hill}
    \end{subfigure}
    \vskip\baselineskip

    \begin{subfigure}{\textwidth}
    \includegraphics[width=0.12\textwidth]{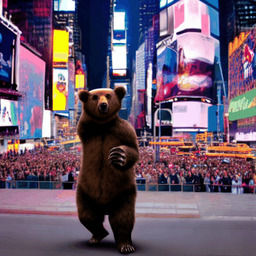} 
    \hfill
    \includegraphics[width=0.12\textwidth]{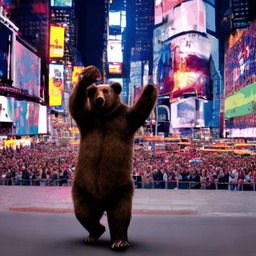} 
    \hfill
    \includegraphics[width=0.12\textwidth]{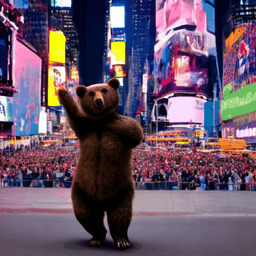} 
    \hfill
    \includegraphics[width=0.12\textwidth]{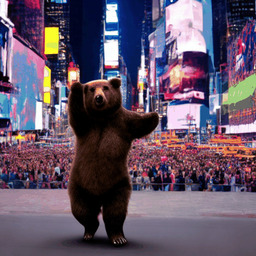} 
    \hfill
    \includegraphics[width=0.12\textwidth]{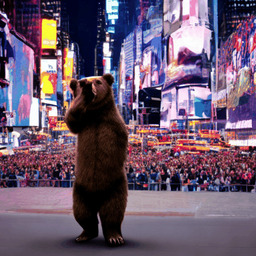} 
    \hfill
    \includegraphics[width=0.12\textwidth]{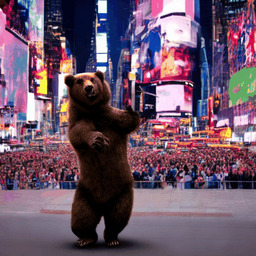} 
    \hfill
    \includegraphics[width=0.12\textwidth]{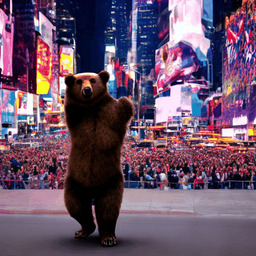} 
    \hfill
    \includegraphics[width=0.12\textwidth]{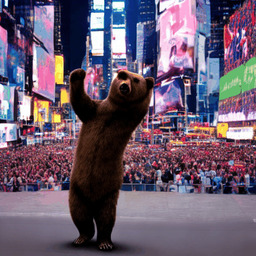} 
    \caption{A bear dancing on times square}
    \end{subfigure}
    \vskip\baselineskip

    \begin{subfigure}{\textwidth}
    \includegraphics[width=0.12\textwidth]{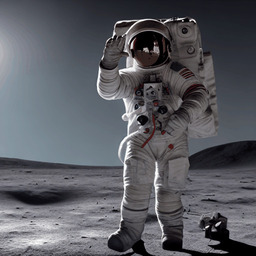} 
    \hfill
    \includegraphics[width=0.12\textwidth]{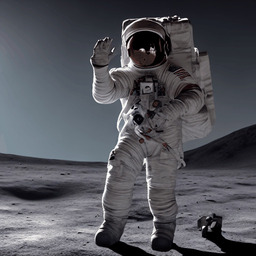} 
    \hfill
    \includegraphics[width=0.12\textwidth]{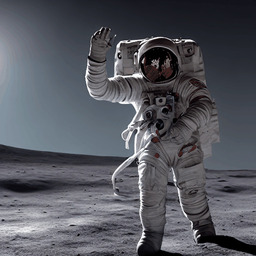} 
    \hfill
    \includegraphics[width=0.12\textwidth]{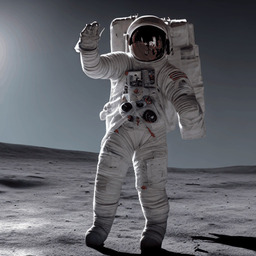} 
    \hfill
    \includegraphics[width=0.12\textwidth]{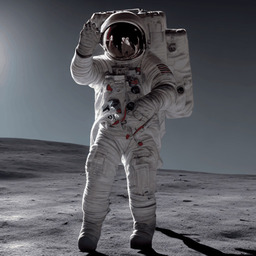} 
    \hfill
    \includegraphics[width=0.12\textwidth]{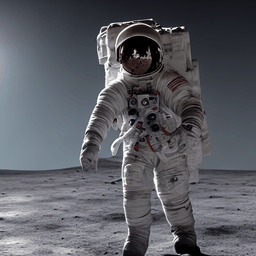} 
    \hfill
    \includegraphics[width=0.12\textwidth]{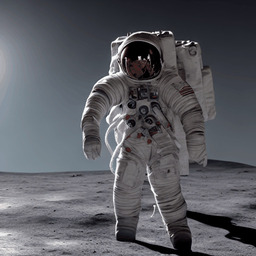} 
    \hfill
    \includegraphics[width=0.12\textwidth]{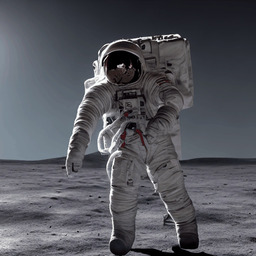} 
    \caption{an astronaut is waving his hands on the moon}
    \end{subfigure}
    \vskip\baselineskip

    \begin{subfigure}{\textwidth}
    \includegraphics[width=0.12\textwidth]{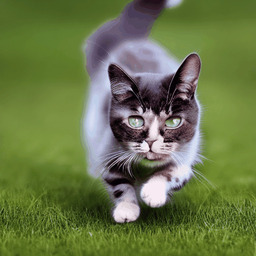} 
    \hfill
    \includegraphics[width=0.12\textwidth]{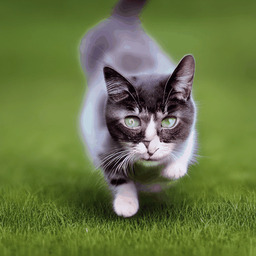} 
    \hfill
    \includegraphics[width=0.12\textwidth]{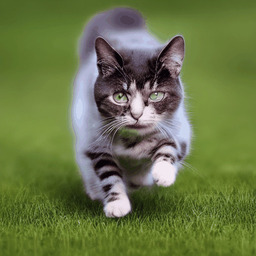} 
    \hfill
    \includegraphics[width=0.12\textwidth]{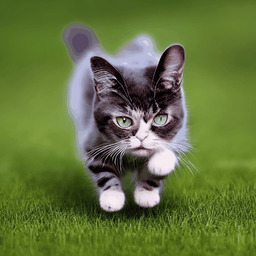} 
    \hfill
    \includegraphics[width=0.12\textwidth]{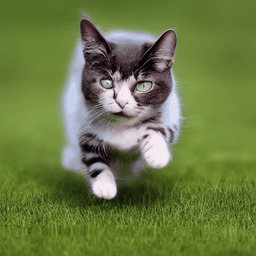} 
    \hfill
    \includegraphics[width=0.12\textwidth]{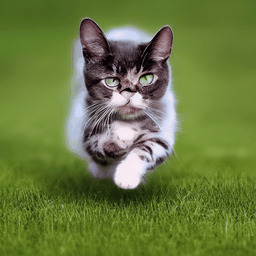} 
    \hfill
    \includegraphics[width=0.12\textwidth]{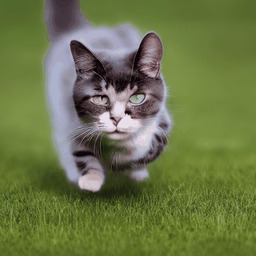} 
    \hfill
    \includegraphics[width=0.12\textwidth]{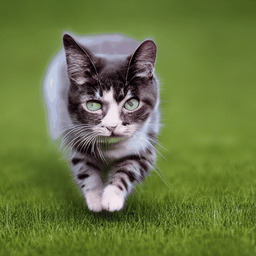} 
    \caption{a high quality realistic photo of a cute cat running on lawn}
    \end{subfigure}
    \vskip\baselineskip

    \caption{Qualitative results of our text-to-video generation method for different prompts.}
    \label{fig:qual_results_full_method}
\end{figure*}

\begin{figure*}
    \centering

     \begin{subfigure}{\textwidth}
    \includegraphics[width=0.12\textwidth]{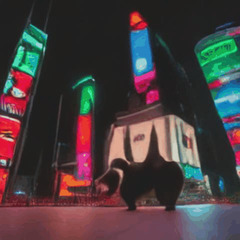} 
    \hfill
    \includegraphics[width=0.12\textwidth]{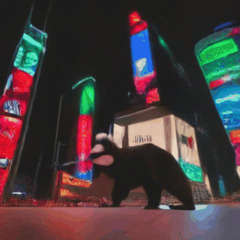} 
    \hfill
    \includegraphics[width=0.12\textwidth]{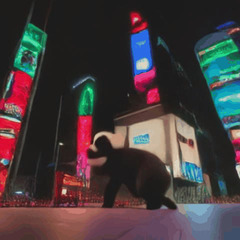} 
    \hfill
    \includegraphics[width=0.12\textwidth]{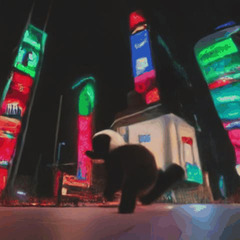} 
    \hfill
    \includegraphics[width=0.12\textwidth]{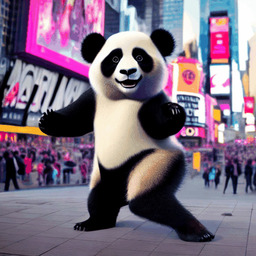} 
    \hfill
    \includegraphics[width=0.12\textwidth]{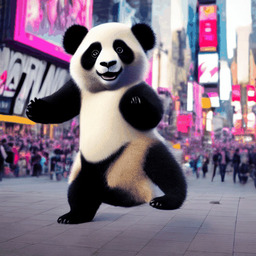}  
    \hfill
    \includegraphics[width=0.12\textwidth]{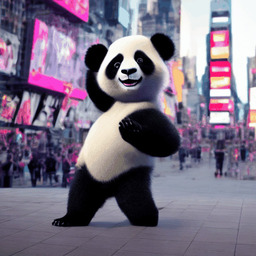} 
    \hfill
    \includegraphics[width=0.12\textwidth]{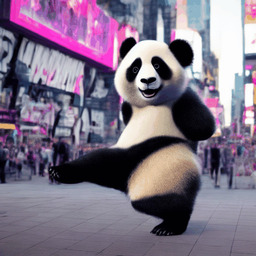}
    \caption{A panda dancing on times square.}

    \end{subfigure}

    \begin{subfigure}{\textwidth}
    \includegraphics[width=0.12\textwidth]{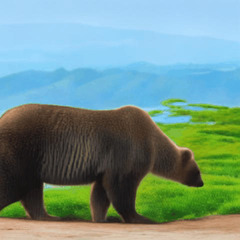} 
    \hfill
    \includegraphics[width=0.12\textwidth]{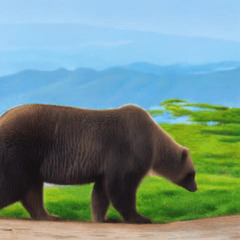} 
    \hfill
    \includegraphics[width=0.12\textwidth]{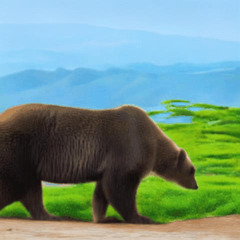} 
    \hfill
    \includegraphics[width=0.12\textwidth]{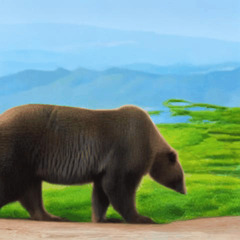} 
    \hfill
    \includegraphics[width=0.12\textwidth]{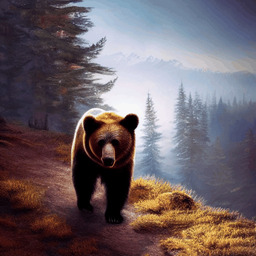} 
    \hfill
    \includegraphics[width=0.12\textwidth]{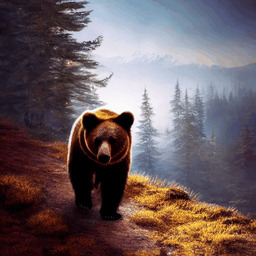}  
    \hfill
    \includegraphics[width=0.12\textwidth]{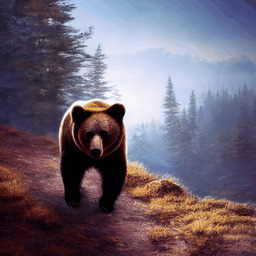} 
    \hfill
    \includegraphics[width=0.12\textwidth]{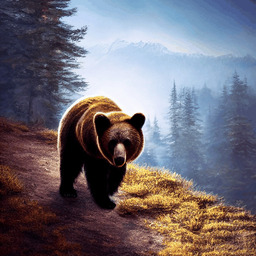} 
    \caption{A bear walking on a mountain.}
    \end{subfigure}

        \begin{subfigure}{\textwidth}
    \includegraphics[width=0.12\textwidth]{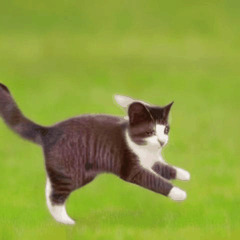} 
    \hfill
    \includegraphics[width=0.12\textwidth]{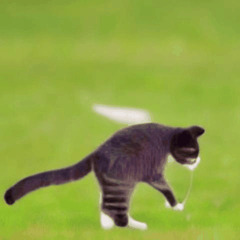} 
    \hfill
    \includegraphics[width=0.12\textwidth]{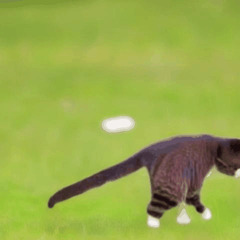} 
    \hfill
    \includegraphics[width=0.12\textwidth]{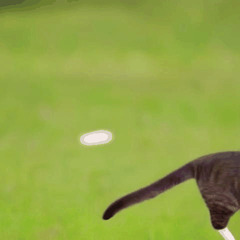} 
    \hfill
    \includegraphics[width=0.12\textwidth]{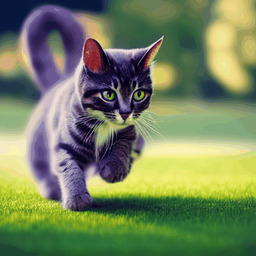} 
    \hfill
    \includegraphics[width=0.12\textwidth]{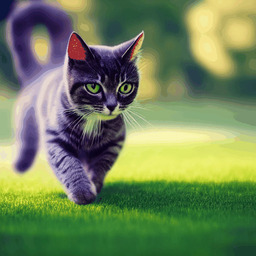}  
    \hfill
    \includegraphics[width=0.12\textwidth]{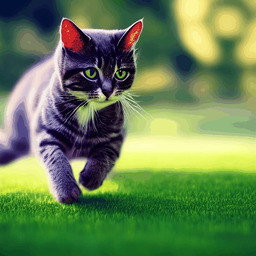} 
    \hfill
    \includegraphics[width=0.12\textwidth]{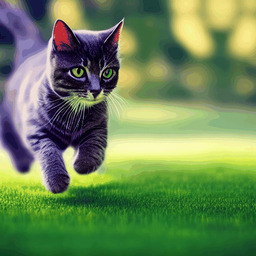}
    \caption{A cat running on the lawn.}

    \end{subfigure}

 \begin{subfigure}{\textwidth}
    \includegraphics[width=0.12\textwidth]{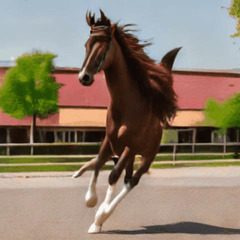} 
    \hfill
    \includegraphics[width=0.12\textwidth]{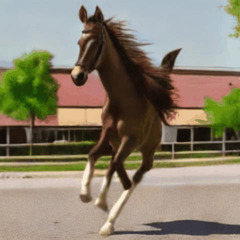} 
    \hfill
    \includegraphics[width=0.12\textwidth]{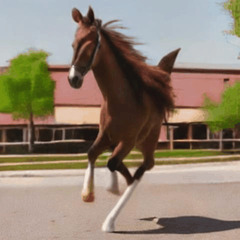} 
    \hfill
    \includegraphics[width=0.12\textwidth]{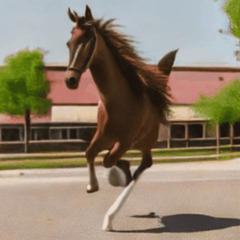} 
    \hfill
    \includegraphics[width=0.12\textwidth]{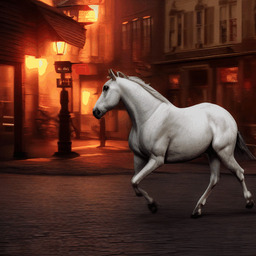} 
    \hfill
    \includegraphics[width=0.12\textwidth]{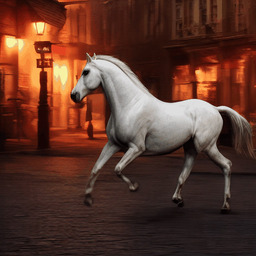}  
    \hfill
    \includegraphics[width=0.12\textwidth]{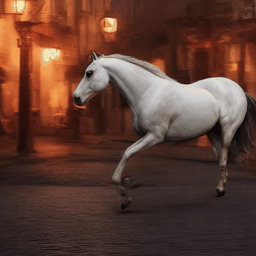} 
    \hfill
    \includegraphics[width=0.12\textwidth]{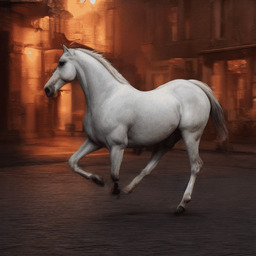}
    \caption{A horse galloping on the street.}

    \end{subfigure}

     \begin{subfigure}{\textwidth}
    \includegraphics[width=0.12\textwidth]{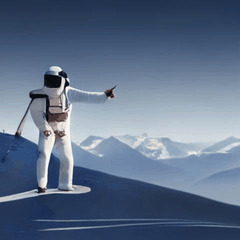} 
    \hfill
    \includegraphics[width=0.12\textwidth]{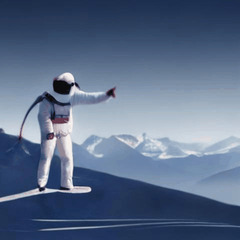} 
    \hfill
    \includegraphics[width=0.12\textwidth]{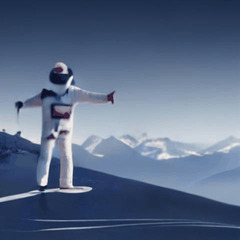} 
    \hfill
    \includegraphics[width=0.12\textwidth]{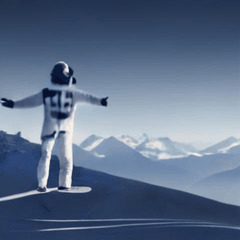} 
    \hfill
    \includegraphics[width=0.12\textwidth]{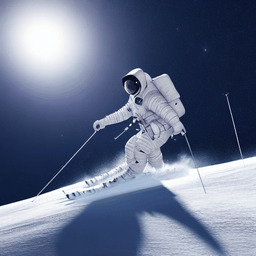} 
    \hfill
    \includegraphics[width=0.12\textwidth]{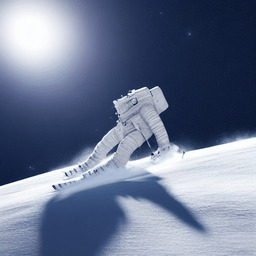}  
    \hfill
    \includegraphics[width=0.12\textwidth]{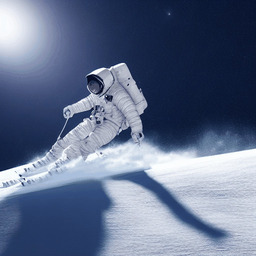} 
    \hfill
    \includegraphics[width=0.12\textwidth]{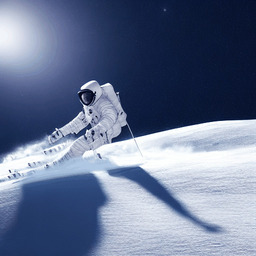}
    \caption{An astronaut skiing down a hill.}

    \end{subfigure}

     \begin{subfigure}{\textwidth}
    \includegraphics[width=0.12\textwidth]{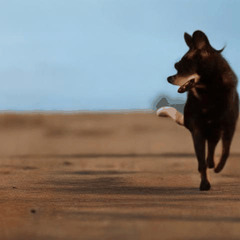} 
    \hfill
    \includegraphics[width=0.12\textwidth]{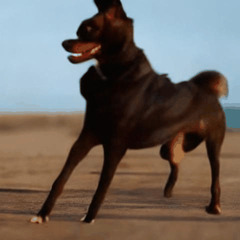} 
    \hfill
    \includegraphics[width=0.12\textwidth]{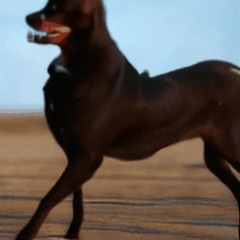} 
    \hfill
    \includegraphics[width=0.12\textwidth]{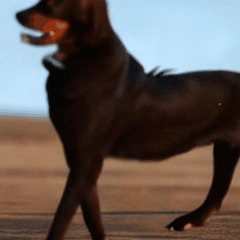} 
    \hfill
    \includegraphics[width=0.12\textwidth]{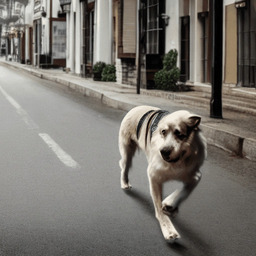} 
    \hfill
    \includegraphics[width=0.12\textwidth]{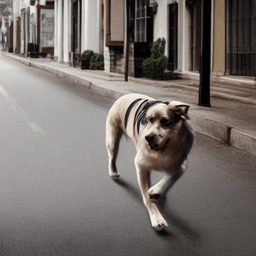}  
    \hfill
    \includegraphics[width=0.12\textwidth]{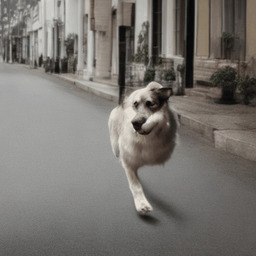} 
    \hfill
    \includegraphics[width=0.12\textwidth]{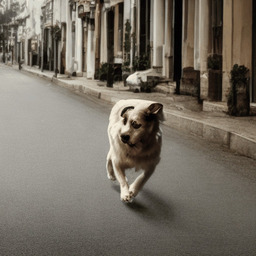}
    \caption{A dog running on the street.}
    \end{subfigure}

     \begin{subfigure}{\textwidth}
    \includegraphics[width=0.12\textwidth]{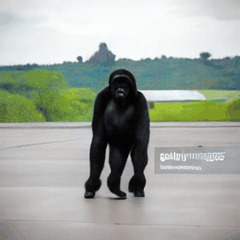} 
    \hfill
    \includegraphics[width=0.12\textwidth]{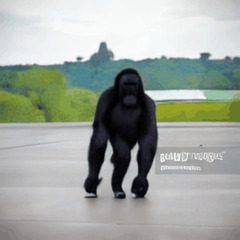} 
    \hfill
    \includegraphics[width=0.12\textwidth]{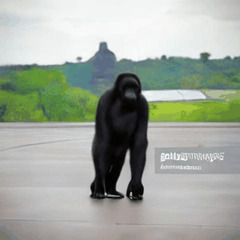} 
    \hfill
    \includegraphics[width=0.12\textwidth]{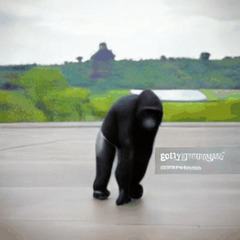} 
    \hfill
    \includegraphics[width=0.12\textwidth]{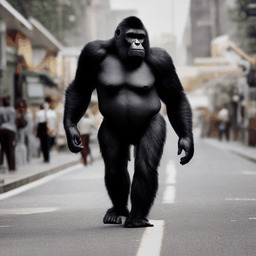} 
    \hfill
    \includegraphics[width=0.12\textwidth]{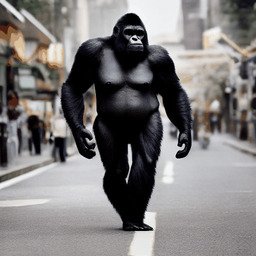}  
    \hfill
    \includegraphics[width=0.12\textwidth]{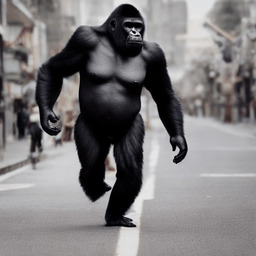} 
    \hfill
    \includegraphics[width=0.12\textwidth]{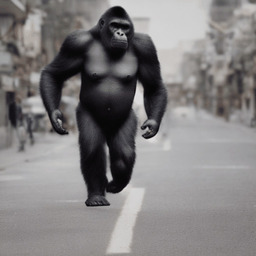}
    \caption{A gorilla walking down the street.}
    \end{subfigure}
    
    \caption{Qualitative comparison between CogVideo \cite{hong2022cogvideo} (frames 1-4 in each row) and our method (frames 5-8 in each row).}
    \label{fig:unconstrained-text2video-ours-vs-cog}
\end{figure*}

Fig.~\ref{fig:qual_results_full_method} shows qualitative results for different actions, e.g. "skiing", "waving" or "dancing".
Thanks to our proposed attention modification, generated frames are consistent across time regarding style and scene.
We obtain plausible motions due to the proposed motion latent approach. 
As can be seen in Fig.~\ref{fig:unconstrained-text2video-ours-vs-cog}, our method performs comparable or sometimes even better than a state-of-the-art approach CogVideo\cite{hong2022cogvideo} which has been trained on a large-scale video data in contrast with our optimization-free approach. 
Fig.~\ref{fig:unconstrained-text2video-ours-vs-cog}(a-b)\&(e) show that generated videos by our method are more text-aligned than CogVideo, regarding the scene. Also the depicted motion is with higher quality in several video generation (e.g. Fig.~\ref{fig:unconstrained-text2video-ours-vs-cog}(a)\&(e)\&(g)).

\subsection{Ablation Studies}
\label{apendix:unconstrained-text2video_ablation}
We conduct additional ablation studies regarding background smoothing, cross-frame attention, latent motion and the number $\Delta t$ of DDPM forward steps. 

\textbf{Timestep to apply motion on latents:} Applying motion on the latent codes $x_{T}$ (corresponding to $\Delta t = 0$)  leads mainly to a global shift without any individual motion of the object, as can be seen for instance at the video of the horse galloping or the gorilla dancing in Fig.~\ref{fig:ablation_ddpm_forward_steps}. It is thus crucial to apply motion on the latents for $T' < T$. We empirically set $\Delta t = 60$ in our method, which provides good object motions (see Fig.~\ref{fig:ablation_ddpm_forward_steps}).

\textbf{Background smoothing:} We visualize the impact of using background smoothing in Fig.~\ref{fig:ablation-background-smoothing-1} and Fig.~\ref{fig:ablation-background-smoothing-2}.
When background smoothing is turned on, $\alpha = 0.6$ is used. When activate, the background is more consistent and better preserved (see e.g. red sign in Fig.~\ref{fig:ablation-background-smoothing-2}).

\textbf{Cross-frame attention and motion latents:} 
Finally, we present additional results where we study the importance of cross-frame attention and motion information on latent codes in Fig.~\ref{fig:ablation-unconditional-attention-motion}. Without cross-frame attention and without motion information on latents the scene differs from frame to frame, and the identity of the main object is not preserved.   
With motion on latents activated, the poses of the objects are better aligned. Yet, the appearance differs between the frames (e.g. when looking at the depicted dog sequence). The identity is much better preserved when cross-frame attention is activated. Also the background scene is more aligned. Finally, we obtain the best results when both, cross-frame attention and motion on latents are activated.

\setcounter{table}{\value{figure}} 
\begin{table*}
\captionsetup{name=Figure}
        \centering
        \begin{tabular}{ cM{18mm} M{ 18mm}M{ 18mm}M{ 18mm}M{ 18mm}M{18mm}M{18mm}M{18mm}M{18mm}}
            \hspace{-0.5cm}\rotatebox[origin=c]{90}{$\Delta t = 0$} &  {\includegraphics[width=.12\textwidth]{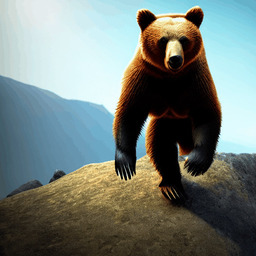}} &  {\includegraphics[width=.12\textwidth]{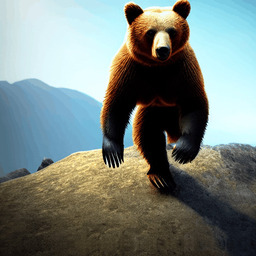}} &  {\includegraphics[width=.12\textwidth]{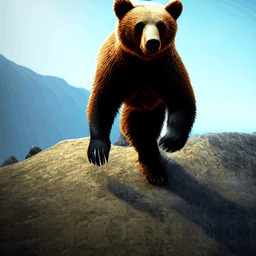}} &  {\includegraphics[width=.12\textwidth]{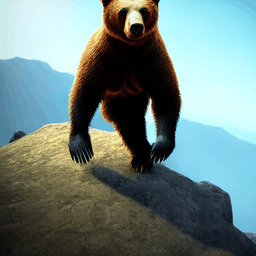}} &  
            {\includegraphics[width=.12\textwidth]{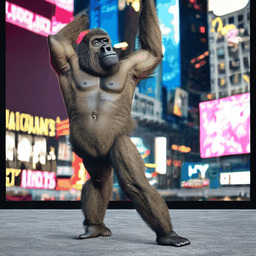}} &  {\includegraphics[width=.12\textwidth]{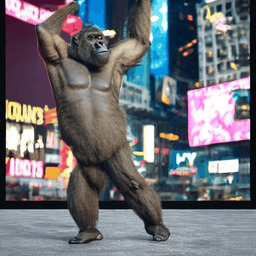}} &  {\includegraphics[width=.12\textwidth]{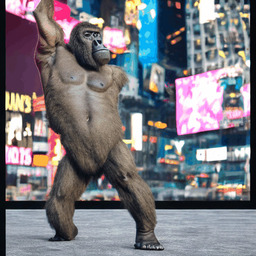}} & 
            {\includegraphics[width=.12\textwidth]{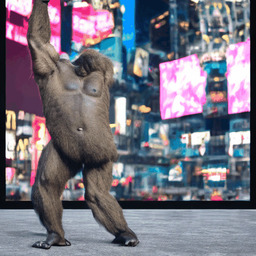}} \\  \hspace{-0.5cm}\rotatebox[origin=c]{90}{$\Delta t = 60 $} &  {\includegraphics[width=.12\textwidth]{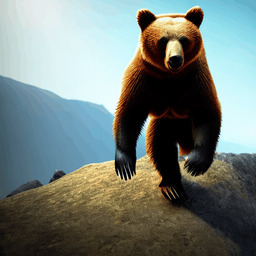}} &  {\includegraphics[width=.12\textwidth]{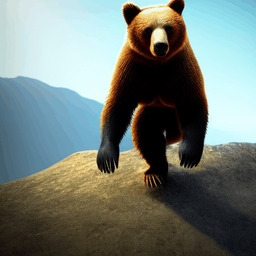}} &  {\includegraphics[width=.12\textwidth]{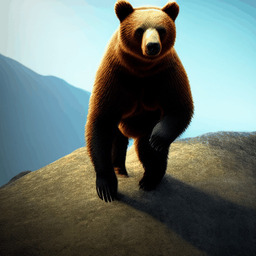}} &  {\includegraphics[width=.12\textwidth]{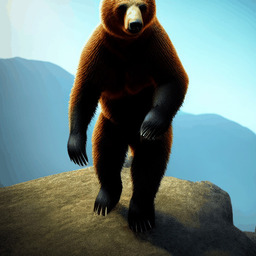}} &  
            {\includegraphics[width=.12\textwidth]{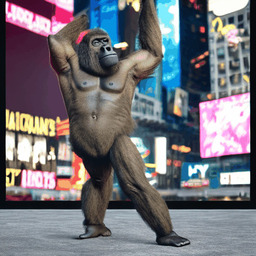}} &  {\includegraphics[width=.12\textwidth]{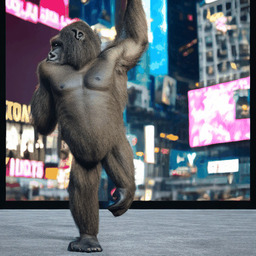}} &  {\includegraphics[width=.12\textwidth]{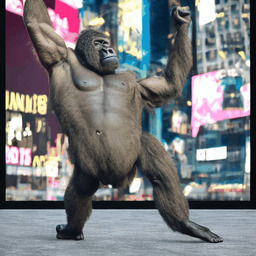}} &  {\includegraphics[width=.12\textwidth]{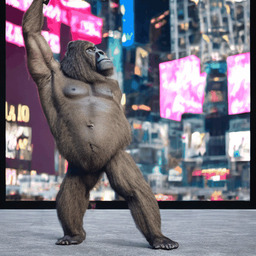}}  \\ 
            \hline
            \hspace{-0.5cm}\rotatebox[origin=c]{90}{$\Delta t = 0$} &  {\includegraphics[width=.12\textwidth]{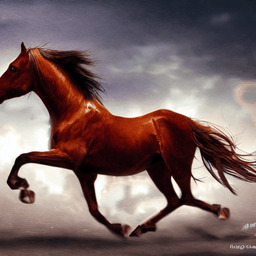}} &  {\includegraphics[width=.12\textwidth]{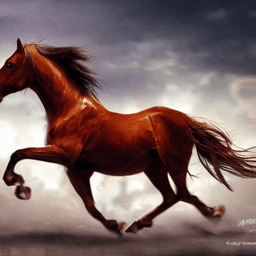}} &  {\includegraphics[width=.12\textwidth]{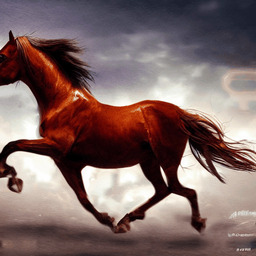}} &  {\includegraphics[width=.12\textwidth]{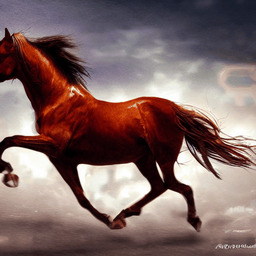}} &  
            {\includegraphics[width=.12\textwidth]{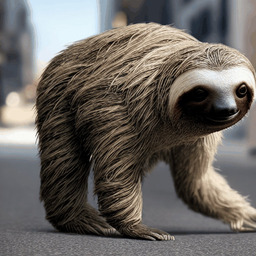}} &  {\includegraphics[width=.12\textwidth]{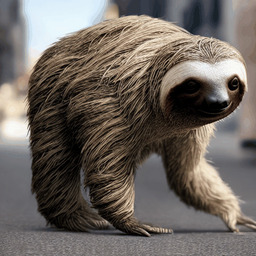}} &  {\includegraphics[width=.12\textwidth]{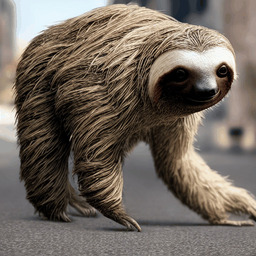}} & 
            {\includegraphics[width=.12\textwidth]{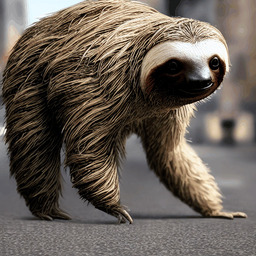}} \\
            \hspace{-0.5cm}\rotatebox[origin=c]{90}{$\Delta t = 60 $} &  {\includegraphics[width=.12\textwidth]{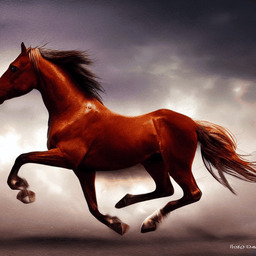}} &  {\includegraphics[width=.12\textwidth]{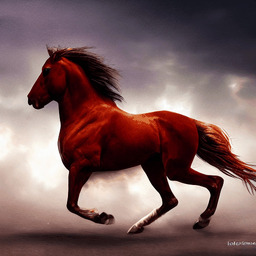}} &  {\includegraphics[width=.12\textwidth]{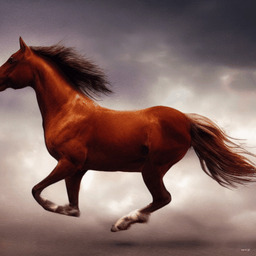}} &  {\includegraphics[width=.12\textwidth]{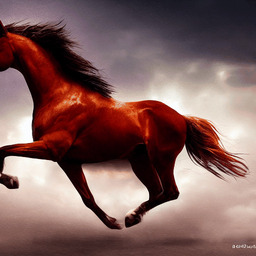}} &  
            {\includegraphics[width=.12\textwidth]{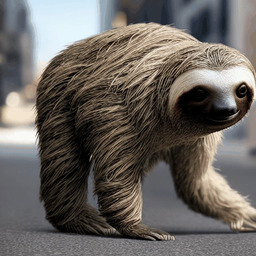}} &  {\includegraphics[width=.12\textwidth]{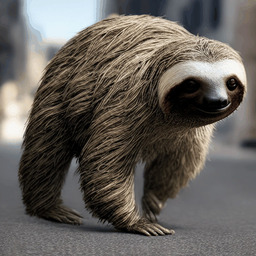}} &  {\includegraphics[width=.12\textwidth]{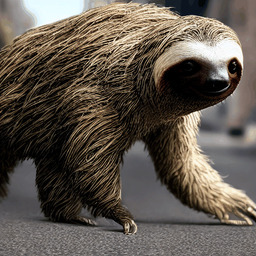}} &  {\includegraphics[width=.12\textwidth]{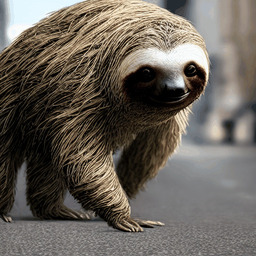}}  \\
            \hspace{-0.5cm}\rotatebox[origin=c]{90}{$\Delta t = 0$} &  {\includegraphics[width=.12\textwidth]{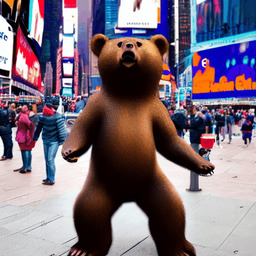}} &  {\includegraphics[width=.12\textwidth]{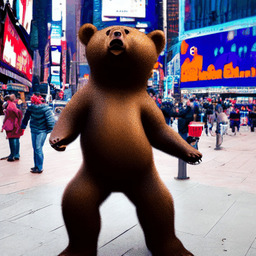}} &  {\includegraphics[width=.12\textwidth]{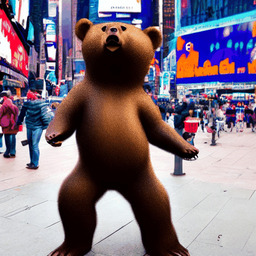}} &  {\includegraphics[width=.12\textwidth]{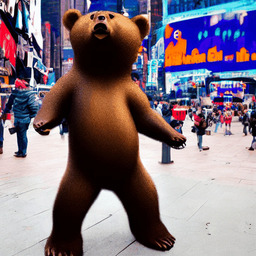}} &  
            {\includegraphics[width=.12\textwidth]{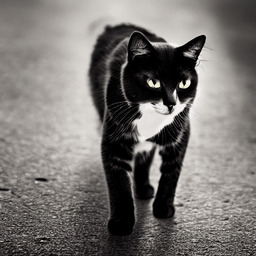}} &  {\includegraphics[width=.12\textwidth]{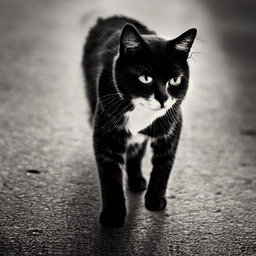}} &  {\includegraphics[width=.12\textwidth]{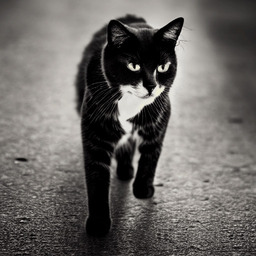}} & 
            {\includegraphics[width=.12\textwidth]{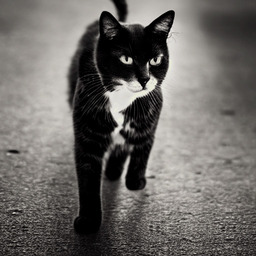}} \\
            \hspace{-0.5cm}\rotatebox[origin=c]{90}{$\Delta t = 60 $} &  {\includegraphics[width=.12\textwidth]{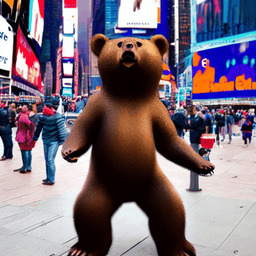}} &  {\includegraphics[width=.12\textwidth]{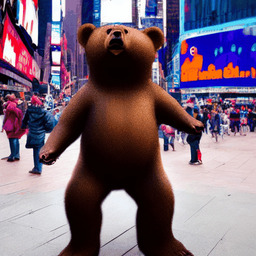}} &  {\includegraphics[width=.12\textwidth]{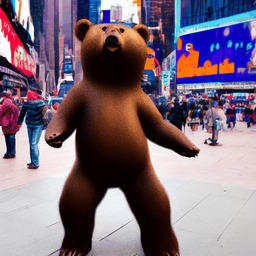}} &  {\includegraphics[width=.12\textwidth]{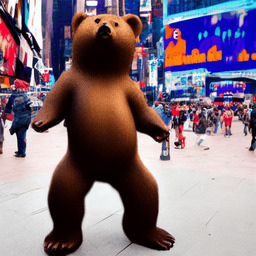}} &  
            {\includegraphics[width=.12\textwidth]{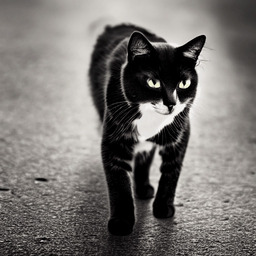}} &  {\includegraphics[width=.12\textwidth]{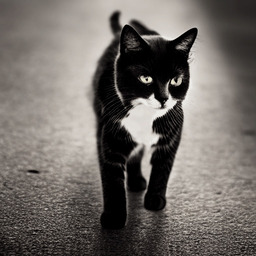}} &  {\includegraphics[width=.12\textwidth]{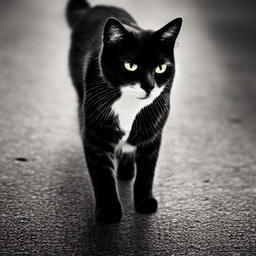}} &  {\includegraphics[width=.12\textwidth]{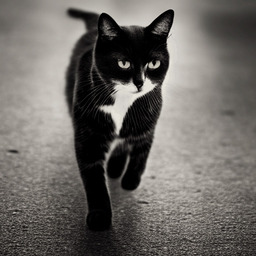}}  \\
        \end{tabular}
        \caption{Ablation study on the number $\Delta t$ of DDPM forward steps. The prompts to our method for text-to-video generation are (from left to right, top to bottom): ``A bear walking on a mountain", ``a gorilla dancing on times square", ``A horse galloping on a street", ``a sloth walking down the street", ``A bear dancing on times square", ``A cat walking down the street."}
        \label{fig:ablation_ddpm_forward_steps}
    \end{table*}
\setcounter{figure}{\value{table}}

\begin{figure*}
    \centering
    \begin{subfigure}{\textwidth}
    \includegraphics[width=0.22\textwidth]{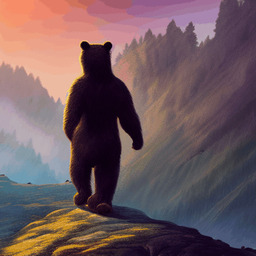} 
    \hfill
    \includegraphics[width=0.22\textwidth]{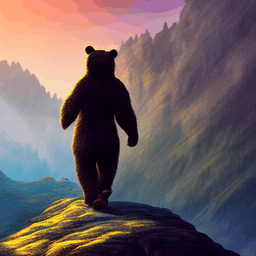} 
    \hfill
    \includegraphics[width=0.22\textwidth]{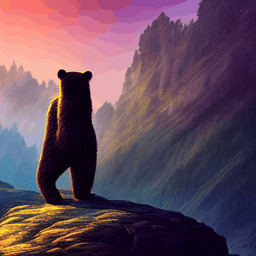} 
    \hfill
    \includegraphics[width=0.22\textwidth]{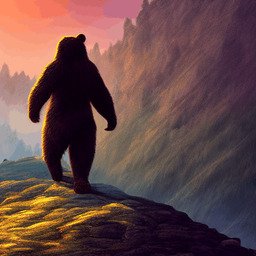} 
    \caption{a bear walking on a mountain, no background smoothing}
    \end{subfigure}
    \begin{subfigure}{\textwidth}
     \includegraphics[width=0.22\textwidth]{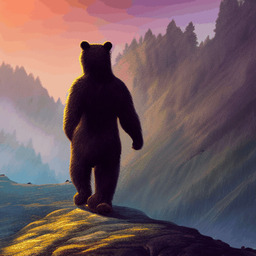} 
    \hfill
    \includegraphics[width=0.22\textwidth]{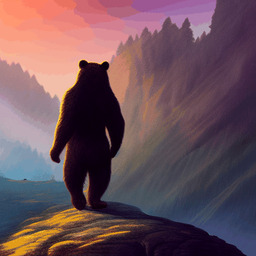} 
    \hfill
    \includegraphics[width=0.22\textwidth]{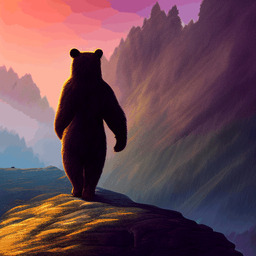} 
    \hfill
    \includegraphics[width=0.22\textwidth]{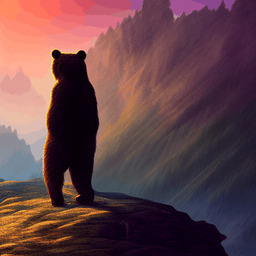} 
    \caption{a bear walking on a mountain, with background smoothing}
    \end{subfigure}
    \begin{subfigure}{\textwidth}
    \includegraphics[width=0.22\textwidth]{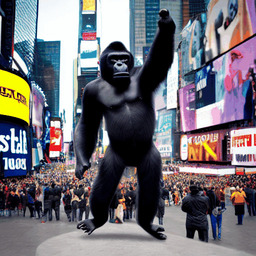} 
    \hfill
    \includegraphics[width=0.22\textwidth]{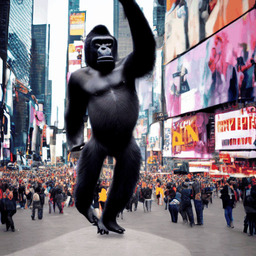} 
    \hfill
    \includegraphics[width=0.22\textwidth]{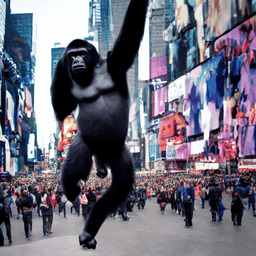} 
    \hfill
    \includegraphics[width=0.22\textwidth]{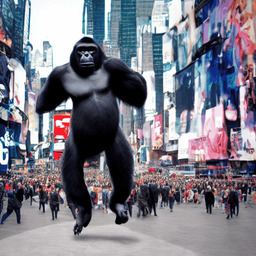} 
    \caption{a gorilla dancing on times square, no background smoothing}
    \end{subfigure}
    \begin{subfigure}{\textwidth}
     \includegraphics[width=0.22\textwidth]{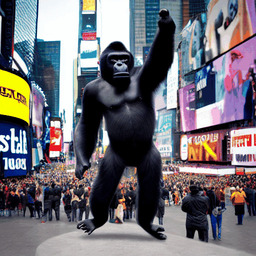} 
    \hfill
    \includegraphics[width=0.22\textwidth]{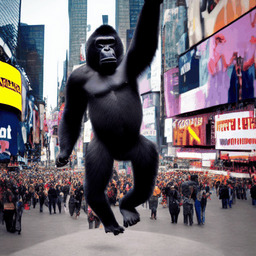} 
    \hfill
    \includegraphics[width=0.22\textwidth]{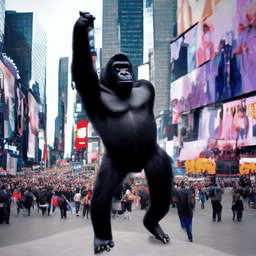} 
    \hfill
    \includegraphics[width=0.22\textwidth]{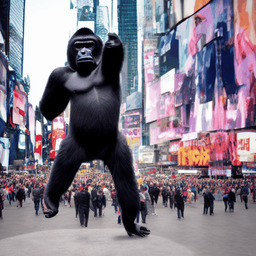} 
    \caption{a gorilla dancing on times square, with background smoothing}
    \end{subfigure}
    \caption{Ablation study on background smoothing.}
    \label{fig:ablation-background-smoothing-1}
\end{figure*}

\begin{figure*}
    \centering
    \begin{subfigure}{\textwidth}
    \includegraphics[width=0.22\textwidth]{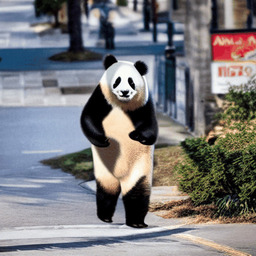} 
    \hfill
    \includegraphics[width=0.22\textwidth]{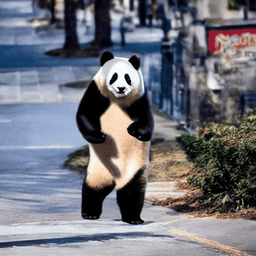} 
    \hfill
    \includegraphics[width=0.22\textwidth]{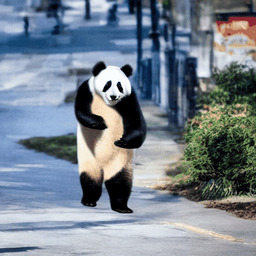} 
    \hfill
    \includegraphics[width=0.22\textwidth]{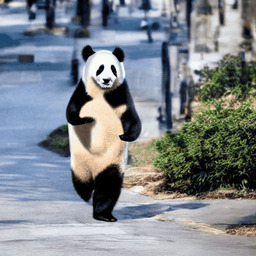} 
    \caption{a panda walking alone down the street, no background smoothing}
    \end{subfigure}
    \begin{subfigure}{\textwidth}
     \includegraphics[width=0.22\textwidth]{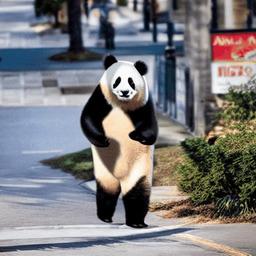} 
    \hfill
    \includegraphics[width=0.22\textwidth]{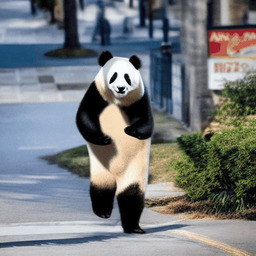} 
    \hfill
    \includegraphics[width=0.22\textwidth]{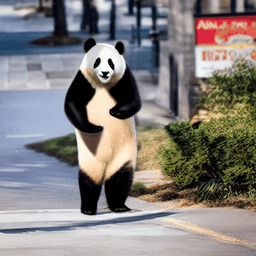} 
    \hfill
    \includegraphics[width=0.22\textwidth]{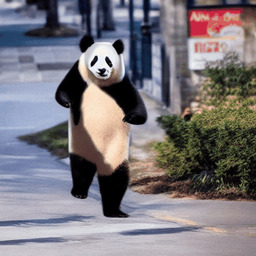} 
    \caption{a panda walking alone down the street, with background smoothing}
    \end{subfigure}
    \begin{subfigure}{\textwidth}
    \includegraphics[width=0.22\textwidth]{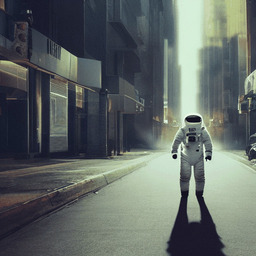} 
    \hfill
    \includegraphics[width=0.22\textwidth]{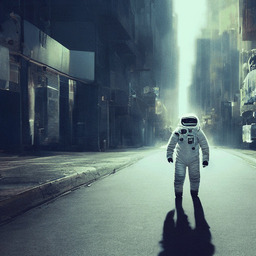} 
    \hfill
    \includegraphics[width=0.22\textwidth]{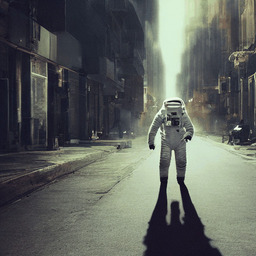} 
    \hfill
    \includegraphics[width=0.22\textwidth]{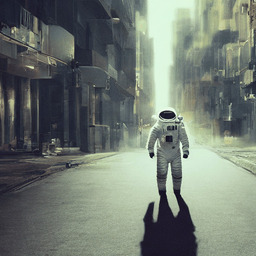} 
    \caption{an astronaut walking on a street, no background smoothing}
    \end{subfigure}
    \begin{subfigure}{\textwidth}
     \includegraphics[width=0.22\textwidth]{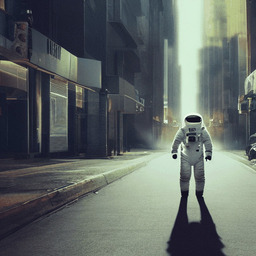} 
    \hfill
    \includegraphics[width=0.22\textwidth]{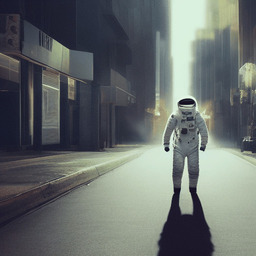} 
    \hfill
    \includegraphics[width=0.22\textwidth]{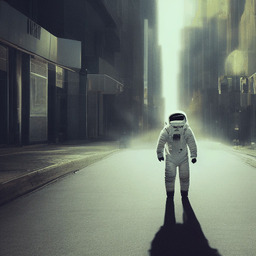} 
    \hfill
    \includegraphics[width=0.22\textwidth]{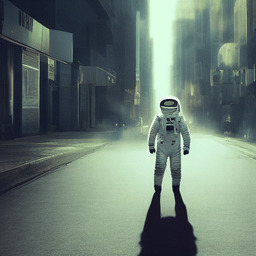} 
    \caption{an astronaut walking on a street, with background smoothing}
    \end{subfigure}
    \caption{Ablation study on background smoothing.}
    \label{fig:ablation-background-smoothing-2}
\end{figure*}



\setcounter{table}{\value{figure}} 
\begin{table*}
\captionsetup{name=Figure}
        \centering
        \begin{tabular}{ cM{18mm} M{ 18mm}M{ 18mm}M{ 18mm}M{ 18mm}M{18mm}M{18mm}M{18mm}M{18mm}}
            \vspace{0.2cm}\hspace{-0.5cm}\rotatebox[origin=c]{90}{\mbox{\begin{varwidth}{5cm} \scriptsize No Motion in latents. \\ No CF-Attention\end{varwidth}}} &  
            {\includegraphics[width=.11\textwidth]{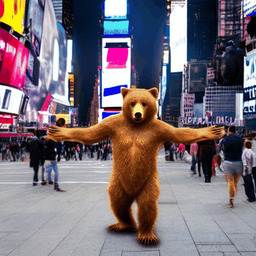}} &  
            {\includegraphics[width=.11\textwidth]{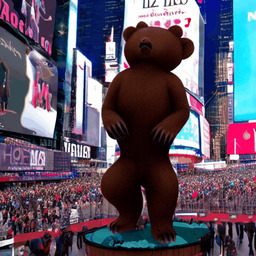}} &  
            {\includegraphics[width=.11\textwidth]{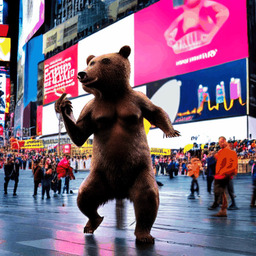}} &  
            {\includegraphics[width=.11\textwidth]{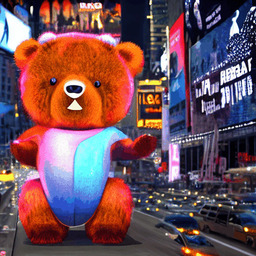}} &  
            
            {\includegraphics[width=.11\textwidth]{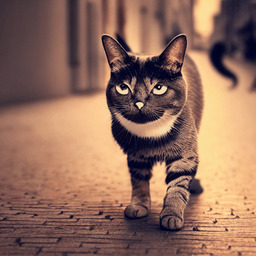}} &  
            {\includegraphics[width=.11\textwidth]{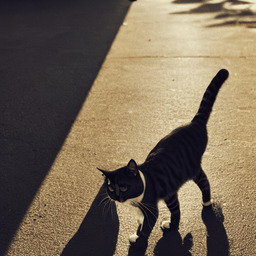}} &  
            {\includegraphics[width=.11\textwidth]{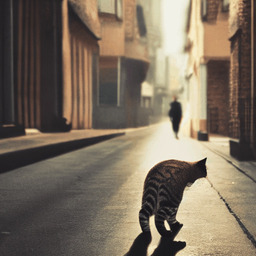}} &  
            {\includegraphics[width=.11\textwidth]{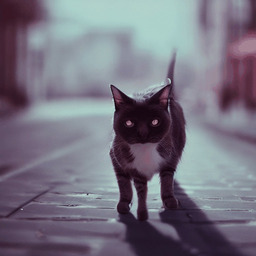}} &   \\  
            \vspace{0.2cm}\hspace{-0.5cm}\rotatebox[origin=c]{90}{\mbox{\begin{varwidth}{5cm} \scriptsize Motion In latents. \\ No CF-Attention\end{varwidth}}} &  
            {\includegraphics[width=.11\textwidth]{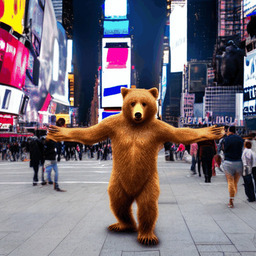}} &  {\includegraphics[width=.11\textwidth]{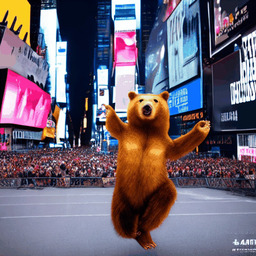}} &  {\includegraphics[width=.11\textwidth]{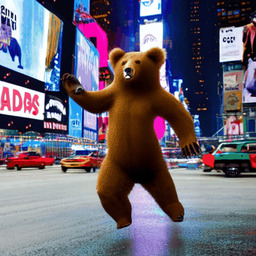}} &  {\includegraphics[width=.11\textwidth]{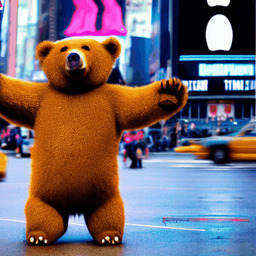}} &  
            {\includegraphics[width=.11\textwidth]{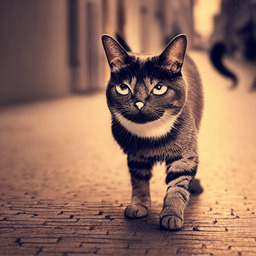}} &  {\includegraphics[width=.11\textwidth]{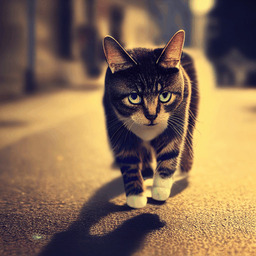}} &  {\includegraphics[width=.11\textwidth]{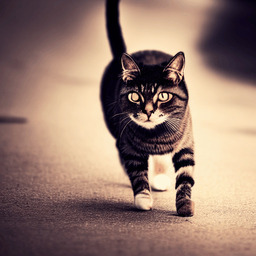}} &  {\includegraphics[width=.11\textwidth]{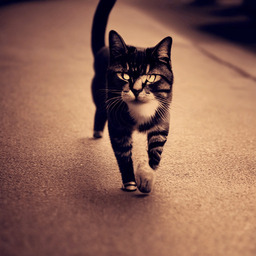}} &   \\ 
            \vspace{0.2cm}\hspace{-0.5cm}\rotatebox[origin=c]{90}{\mbox{\begin{varwidth}{5cm} \scriptsize No Motion In latents. \\  CF-Attention\end{varwidth}}} &  
            {\includegraphics[width=.11\textwidth]{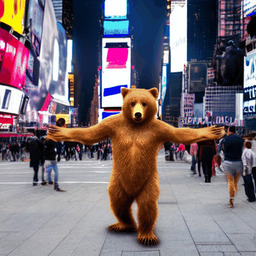}} &  {\includegraphics[width=.11\textwidth]{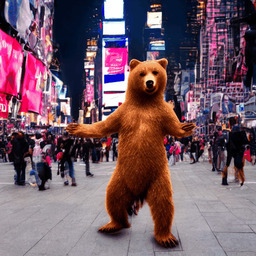}} &  {\includegraphics[width=.11\textwidth]{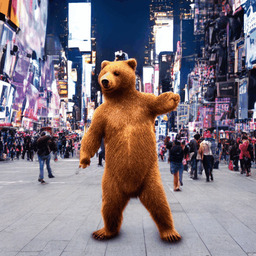}} &  {\includegraphics[width=.11\textwidth]{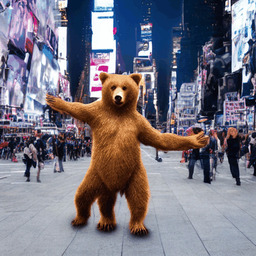}} &  
            {\includegraphics[width=.11\textwidth]{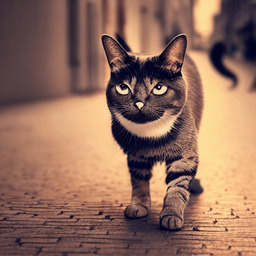}} &  {\includegraphics[width=.11\textwidth]{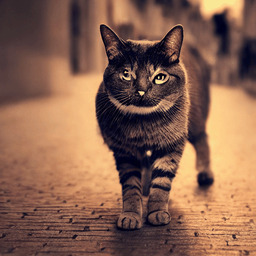}} &  {\includegraphics[width=.11\textwidth]{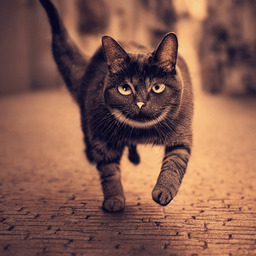}} &  {\includegraphics[width=.11\textwidth]{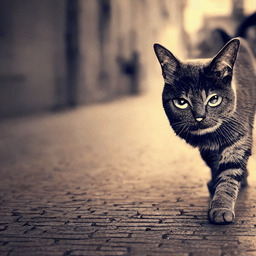}} &    \\ 
            \hspace{-0.5cm}\rotatebox[origin=c]{90}{\mbox{\begin{varwidth}{5cm} \scriptsize Motion In latents. \\ CF-Attention\end{varwidth}}} &  
            {\includegraphics[width=.11\textwidth]{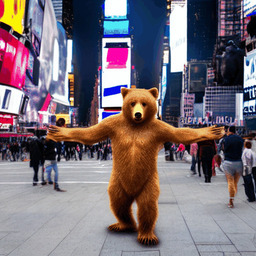}} &  {\includegraphics[width=.11\textwidth]{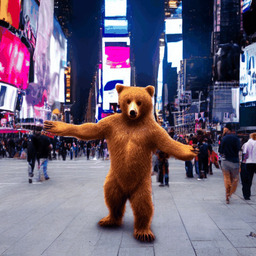}} &  {\includegraphics[width=.11\textwidth]{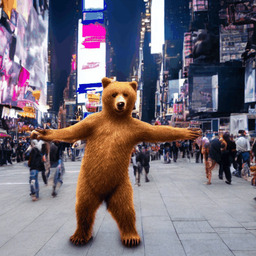}} &  {\includegraphics[width=.11\textwidth]{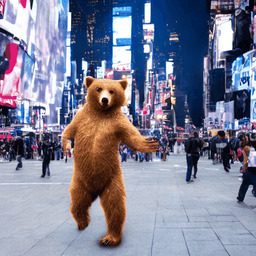}} &  
            {\includegraphics[width=.11\textwidth]{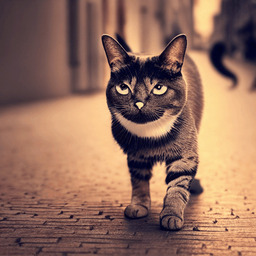}} &  {\includegraphics[width=.11\textwidth]{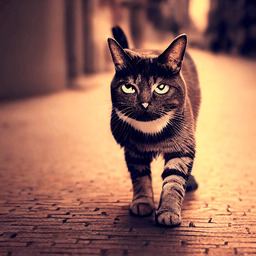}} &  {\includegraphics[width=.11\textwidth]{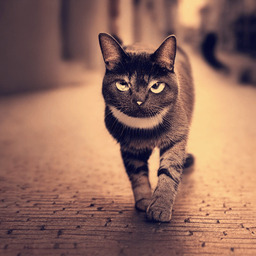}} &  {\includegraphics[width=.11\textwidth]{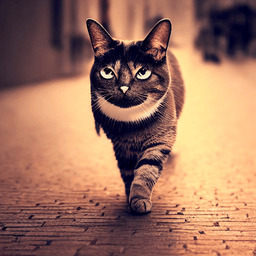}} &  \\

            \vspace{0.2cm}\hspace{-0.5cm}\rotatebox[origin=c]{90}{\mbox{\begin{varwidth}{5cm} \scriptsize No Motion In latents. \\ No CF-Attention\end{varwidth}}} &  
            {\includegraphics[width=.11\textwidth]{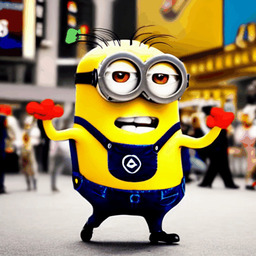}} &  
            {\includegraphics[width=.11\textwidth]{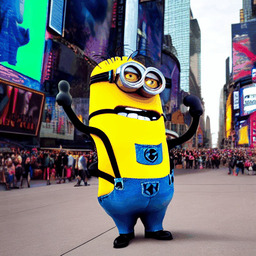}} &  
            {\includegraphics[width=.11\textwidth]{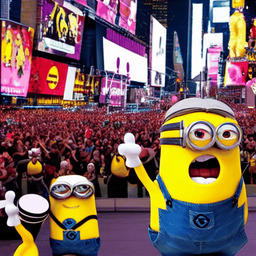}} &  
            {\includegraphics[width=.11\textwidth]{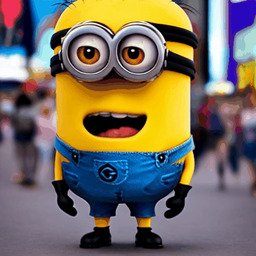}} &  
            
            {\includegraphics[width=.11\textwidth]{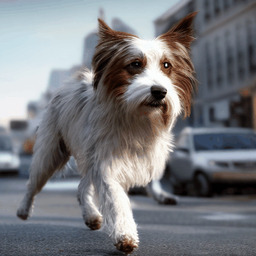}} &  
            {\includegraphics[width=.11\textwidth]{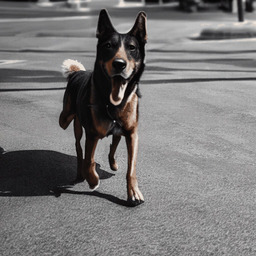}} &  
            {\includegraphics[width=.11\textwidth]{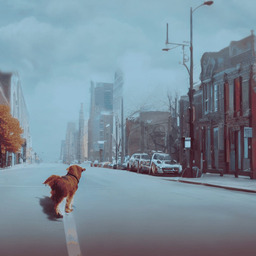}} &  
            {\includegraphics[width=.11\textwidth]{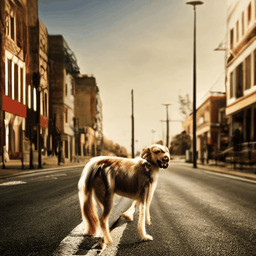}} &   \\  
            \vspace{0.2cm}\hspace{-0.5cm}\rotatebox[origin=c]{90}{\mbox{\begin{varwidth}{5cm} \scriptsize Motion In latents. \\ No CF-Attention\end{varwidth}}} &  
            {\includegraphics[width=.11\textwidth]{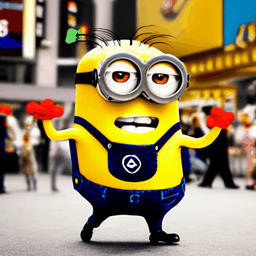}} &  {\includegraphics[width=.11\textwidth]{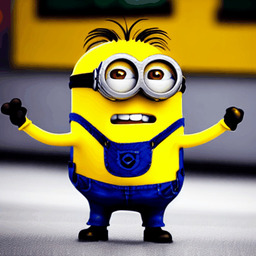}} &  {\includegraphics[width=.11\textwidth]{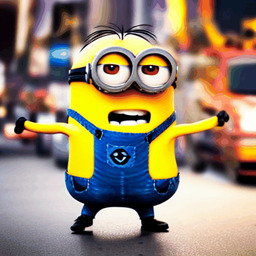}} &  {\includegraphics[width=.11\textwidth]{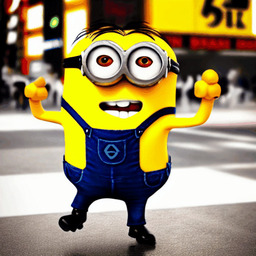}} &  
            {\includegraphics[width=.11\textwidth]{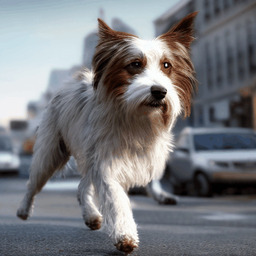}} &  {\includegraphics[width=.11\textwidth]{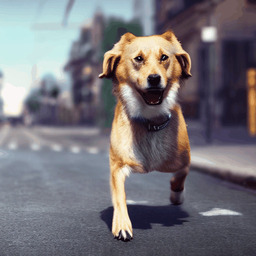}} &  {\includegraphics[width=.11\textwidth]{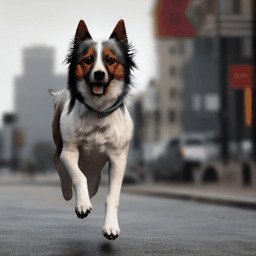}} &  {\includegraphics[width=.11\textwidth]{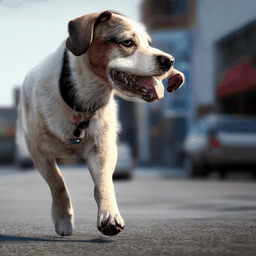}} &   \\ 
            \vspace{0.2cm}\hspace{-0.5cm}\rotatebox[origin=c]{90}{\mbox{\begin{varwidth}{5cm} \scriptsize No Motion In latents. \\  CF-Attention\end{varwidth}}} &  
            {\includegraphics[width=.11\textwidth]{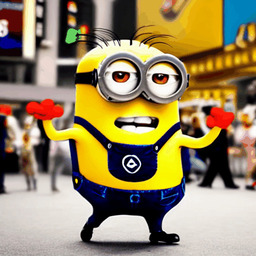}} &  {\includegraphics[width=.11\textwidth]{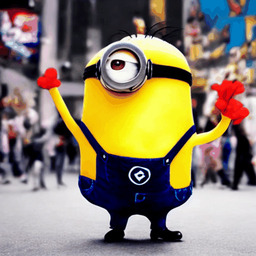}} &  {\includegraphics[width=.11\textwidth]{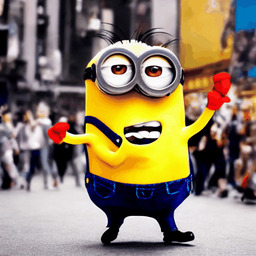}} &  {\includegraphics[width=.11\textwidth]{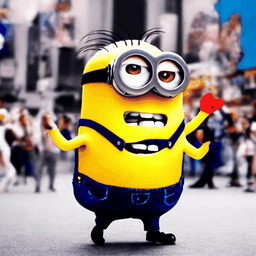}} &  
            {\includegraphics[width=.11\textwidth]{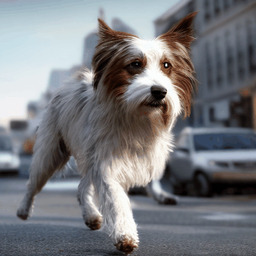}} &  {\includegraphics[width=.11\textwidth]{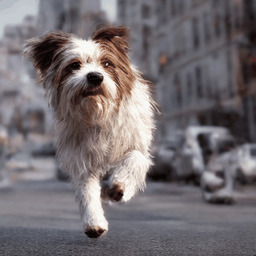}} &  {\includegraphics[width=.11\textwidth]{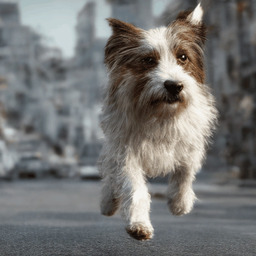}} &  {\includegraphics[width=.11\textwidth]{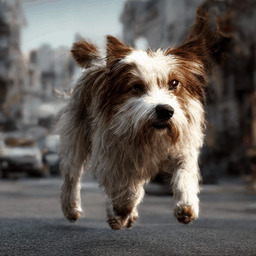}} &    \\ 
            \hspace{-0.5cm}\rotatebox[origin=c]{90}{\mbox{\begin{varwidth}{5cm} \scriptsize Motion In latents. \\ CF-Attention\end{varwidth}}} &  
            {\includegraphics[width=.11\textwidth]{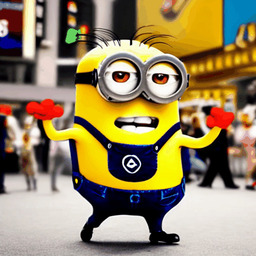}} &  {\includegraphics[width=.11\textwidth]{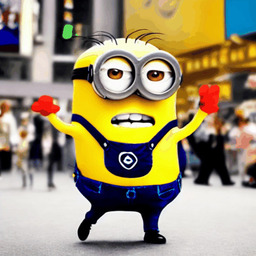}} &  {\includegraphics[width=.11\textwidth]{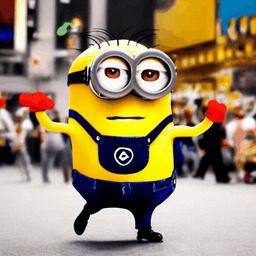}} &  {\includegraphics[width=.11\textwidth]{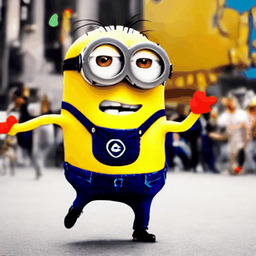}} &  
            {\includegraphics[width=.11\textwidth]{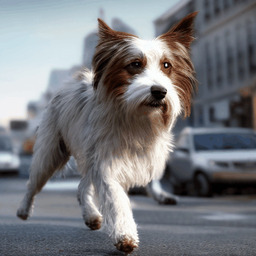}} &  {\includegraphics[width=.11\textwidth]{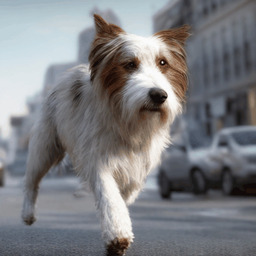}} &  {\includegraphics[width=.11\textwidth]{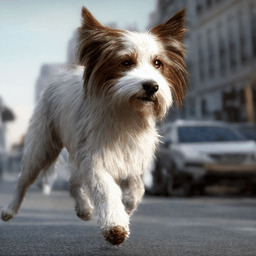}} &  {\includegraphics[width=.11\textwidth]{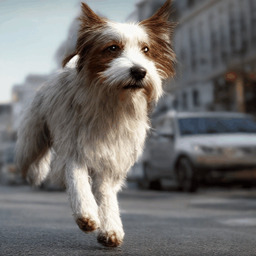}} &  \\ 
        
        \end{tabular}
        \caption{Ablation study on relevance of CF-Attention and Motion In latents. The left half of the rows 1-4 are generated with the prompt ``A bear dancing on times square". The right half of the rows 1-4 are generated with the prompt ``A cat walking on a street".
        The left half or the rows 5-8 are generated with the prompts ``A Minion dancing on times square". The right half of the rows 5 - 8 are generated with the prompt ``A dog walking on a street."}
        \label{fig:ablation-unconditional-attention-motion}
    \end{table*}
\setcounter{figure}{\value{table}} 

%% file: supplemental_chapters/text2video_edge_guidance.tex
\section{Text-to-Video with Edge Guidance}
\label{apendix:t2v_edge_guidance}

\begin{figure*}
    \centering
    \begin{subfigure}{\textwidth}
    \includegraphics[width=0.12\textwidth]{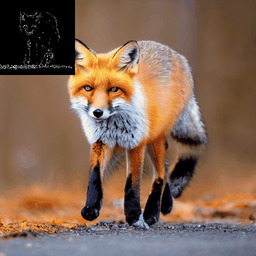} 
    \hfill
    \includegraphics[width=0.12\textwidth]{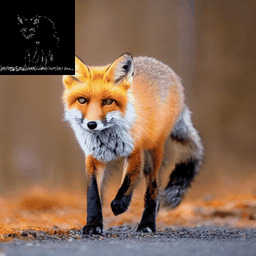} 
    \hfill
    \includegraphics[width=0.12\textwidth]{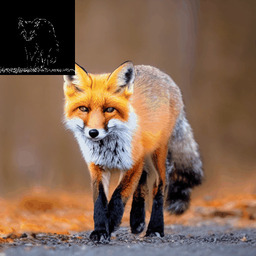} 
    \hfill
    \includegraphics[width=0.12\textwidth]{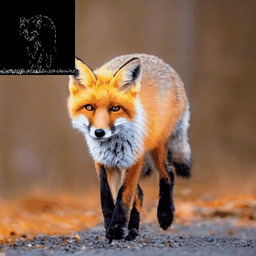} 
    \hfill
    \includegraphics[width=0.12\textwidth]{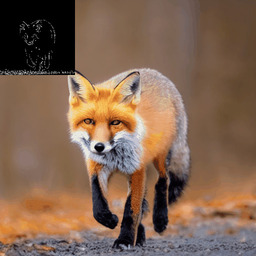} 
    \hfill
    \includegraphics[width=0.12\textwidth]{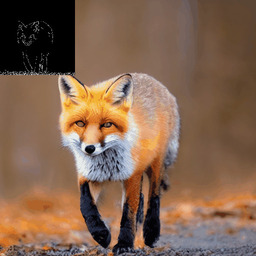} 
    \hfill
    \includegraphics[width=0.12\textwidth]{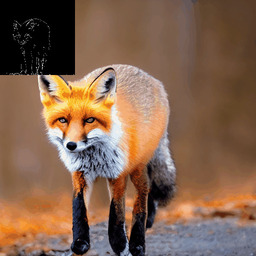} 
    \hfill
    \includegraphics[width=0.12\textwidth]{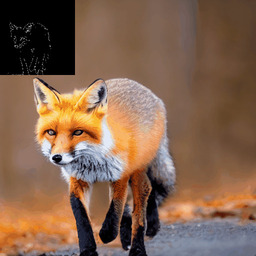} 
    \caption{wild fox is walking, a high-quality, detailed and professional photo}
    \end{subfigure}
    \vskip\baselineskip
    \begin{subfigure}{\textwidth}
    \includegraphics[width=0.12\textwidth]{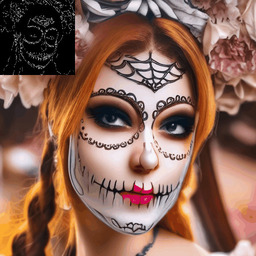} 
    \hfill
    \includegraphics[width=0.12\textwidth]{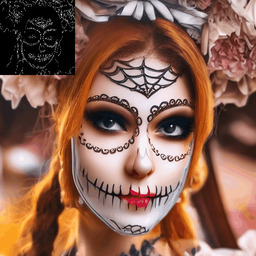} 
    \hfill
    \includegraphics[width=0.12\textwidth]{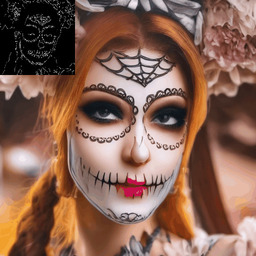} 
    \hfill
    \includegraphics[width=0.12\textwidth]{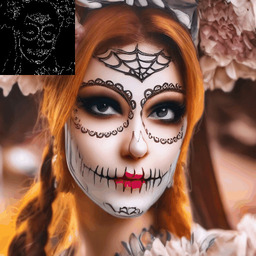} 
    \hfill
    \includegraphics[width=0.12\textwidth]{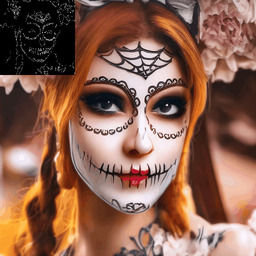} 
    \hfill
    \includegraphics[width=0.12\textwidth]{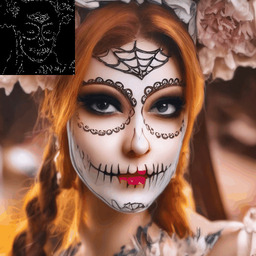} 
    \hfill
    \includegraphics[width=0.12\textwidth]{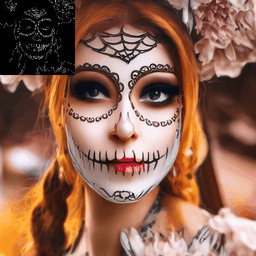} 
    \hfill
    \includegraphics[width=0.12\textwidth]{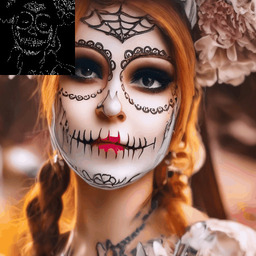} 
    \caption{beautiful girl Halloween style, a high-quality, detailed and professional photo}
    \end{subfigure}
    \vskip\baselineskip
    \begin{subfigure}{\textwidth}
    \includegraphics[width=0.12\textwidth]{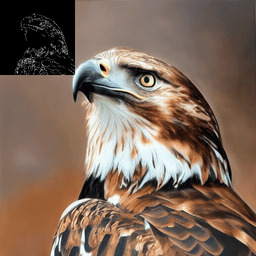} 
    \hfill
    \includegraphics[width=0.12\textwidth]{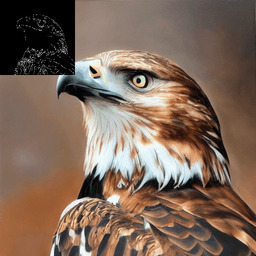} 
    \hfill
    \includegraphics[width=0.12\textwidth]{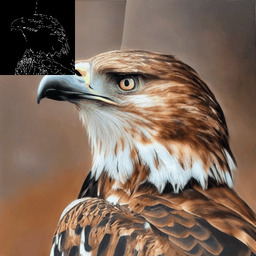} 
    \hfill
    \includegraphics[width=0.12\textwidth]{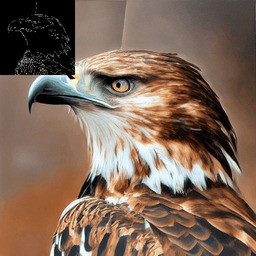} 
    \hfill
    \includegraphics[width=0.12\textwidth]{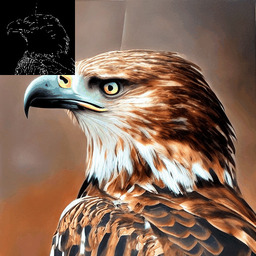} 
    \hfill
    \includegraphics[width=0.12\textwidth]{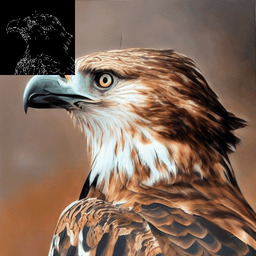} 
    \hfill
    \includegraphics[width=0.12\textwidth]{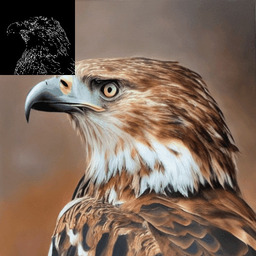} 
    \hfill
    \includegraphics[width=0.12\textwidth]{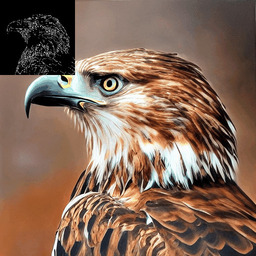} 
    \caption{a hawk, a high-quality, detailed and professional photo}
    \end{subfigure}
    \vskip\baselineskip

    \begin{subfigure}{\textwidth}
    \includegraphics[width=0.12\textwidth]{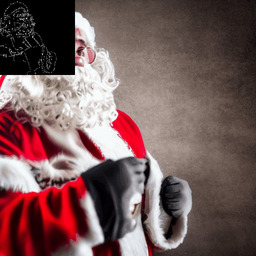} 
    \hfill
    \includegraphics[width=0.12\textwidth]{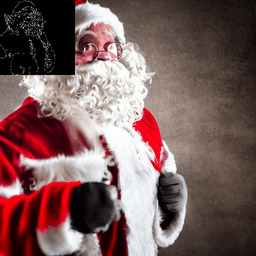} 
    \hfill
    \includegraphics[width=0.12\textwidth]{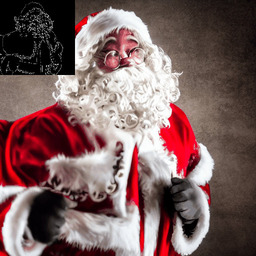} 
    \hfill
    \includegraphics[width=0.12\textwidth]{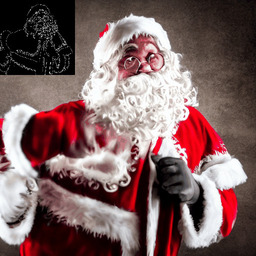} 
    \hfill
    \includegraphics[width=0.12\textwidth]{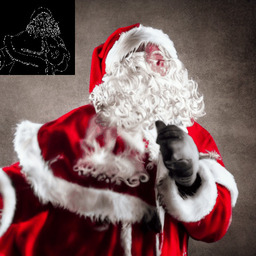} 
    \hfill
    \includegraphics[width=0.12\textwidth]{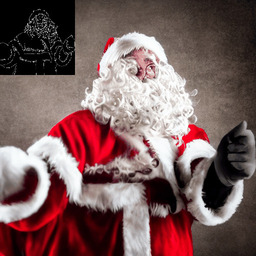} 
    \hfill
    \includegraphics[width=0.12\textwidth]{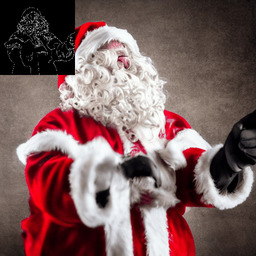} 
    \hfill
    \includegraphics[width=0.12\textwidth]{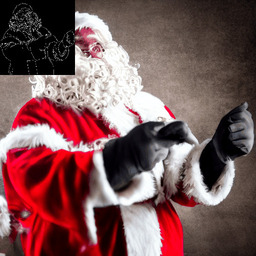} 
    \caption{a santa claus, a high-quality, detailed and professional photo}
    \end{subfigure}
    \vskip\baselineskip

    \begin{subfigure}{\textwidth}
    \includegraphics[width=0.12\textwidth]{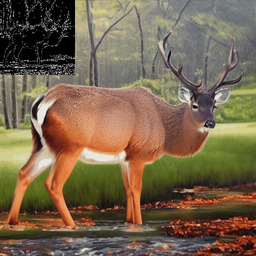} 
    \hfill
    \includegraphics[width=0.12\textwidth]{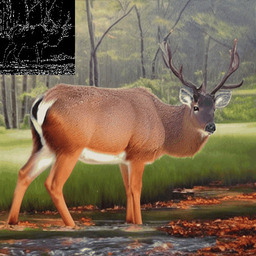} 
    \hfill
    \includegraphics[width=0.12\textwidth]{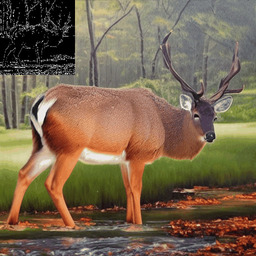} 
    \hfill
    \includegraphics[width=0.12\textwidth]{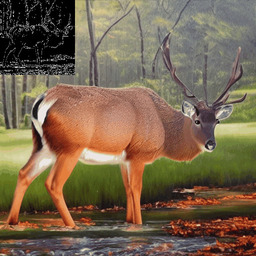} 
    \hfill
    \includegraphics[width=0.12\textwidth]{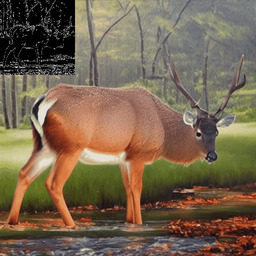} 
    \hfill
    \includegraphics[width=0.12\textwidth]{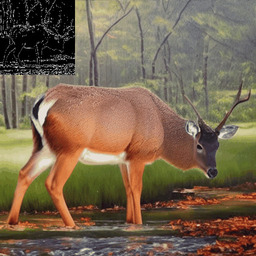} 
    \hfill
    \includegraphics[width=0.12\textwidth]{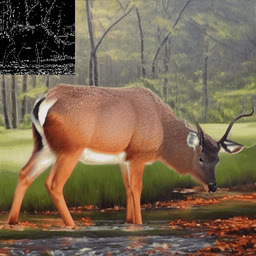} 
    \hfill
    \includegraphics[width=0.12\textwidth]{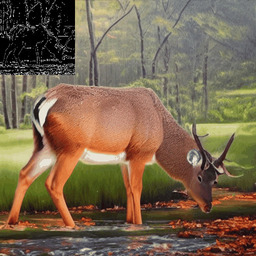} 
    \caption{oil painting of a deer, a high-quality, detailed and professional photo}
    \end{subfigure}
    \vskip\baselineskip

    \begin{subfigure}{\textwidth}
    \includegraphics[width=0.12\textwidth]{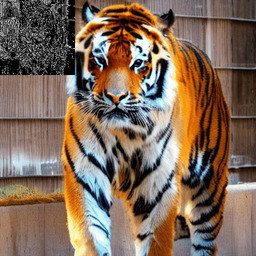} 
    \hfill
    \includegraphics[width=0.12\textwidth]{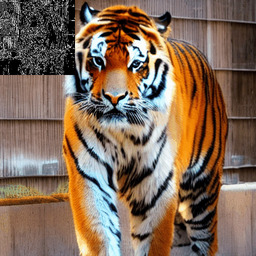} 
    \hfill
    \includegraphics[width=0.12\textwidth]{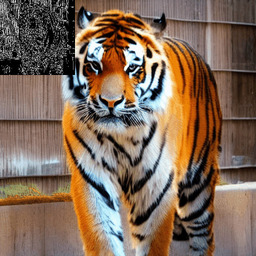} 
    \hfill
    \includegraphics[width=0.12\textwidth]{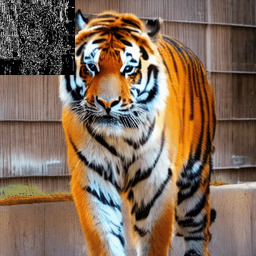} 
    \hfill
    \includegraphics[width=0.12\textwidth]{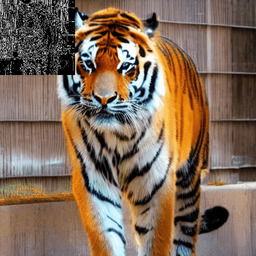} 
    \hfill
    \includegraphics[width=0.12\textwidth]{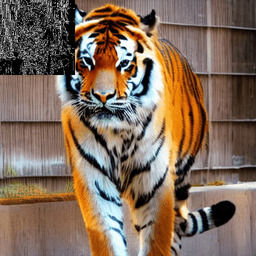} 
    \hfill
    \includegraphics[width=0.12\textwidth]{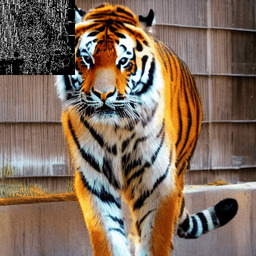} 
    \hfill
    \includegraphics[width=0.12\textwidth]{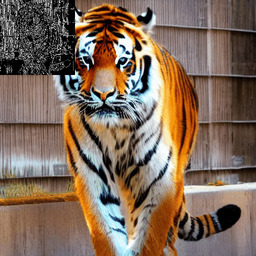} 
    \caption{a tiger, a high-quality, detailed and professional photo}
    \end{subfigure}
    \vskip\baselineskip

    \begin{subfigure}{\textwidth}
    \includegraphics[width=0.12\textwidth]{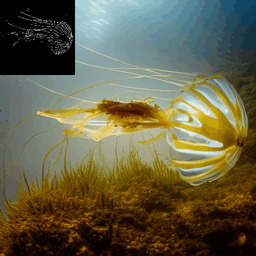} 
    \hfill
    \includegraphics[width=0.12\textwidth]{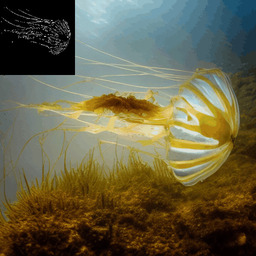} 
    \hfill
    \includegraphics[width=0.12\textwidth]{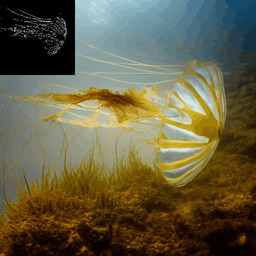} 
    \hfill
    \includegraphics[width=0.12\textwidth]{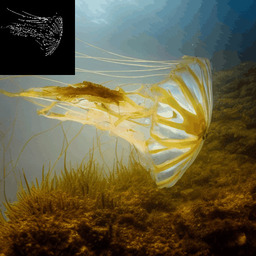} 
    \hfill
    \includegraphics[width=0.12\textwidth]{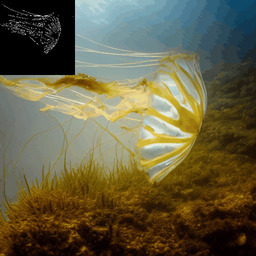} 
    \hfill
    \includegraphics[width=0.12\textwidth]{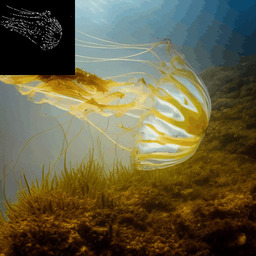} 
    \hfill
    \includegraphics[width=0.12\textwidth]{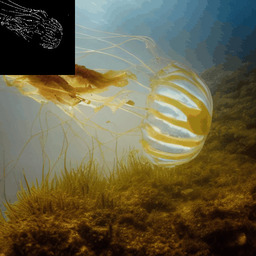} 
    \hfill
    \includegraphics[width=0.12\textwidth]{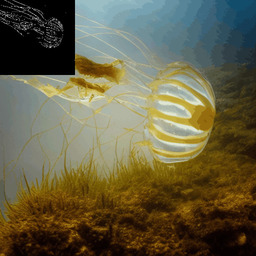} 
    \caption{a jellyfish, a high-quality, detailed and professional photo}
    \end{subfigure}
    \vskip\baselineskip

    \caption{Conditional generation of our method with edge control.}
    \label{fig:qual_results_full_method_edge_guidance}
\end{figure*}

In Fig.~\ref{fig:qual_results_full_method_edge_guidance} we present more video generation results by guiding our method with edge information. 
In Fig.~\ref{fig:table_of_edge_study} we show the effect of our cross-frame attention and motion in latents for text-to-video generation with edge guidance.
As can be noticed when using CF-Attn layer the generation preserves the identity of the person better, and using motion in latents further improves the temporal consistency.


%% file: supplemental_chapters/text2video_edge_db_guidance.tex
\begin{figure*}
    \centering
    \begin{subfigure}{\textwidth}
    \includegraphics[width=0.12\textwidth]{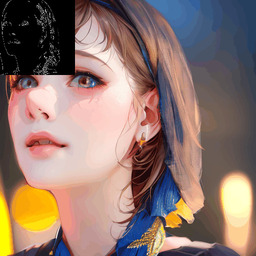} 
    \hfill
    \includegraphics[width=0.12\textwidth]{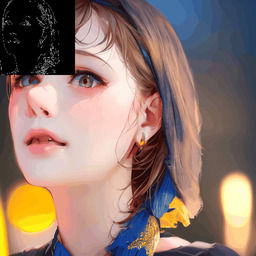} 
    \hfill
    \includegraphics[width=0.12\textwidth]{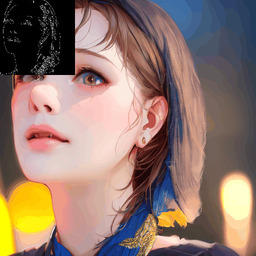} 
    \hfill
    \includegraphics[width=0.12\textwidth]{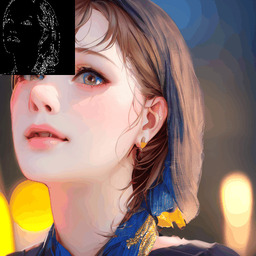} 
    \hfill
    \includegraphics[width=0.12\textwidth]{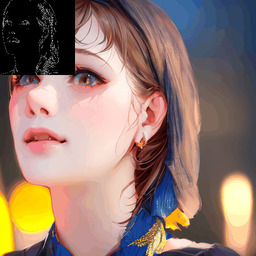} 
    \hfill
    \includegraphics[width=0.12\textwidth]{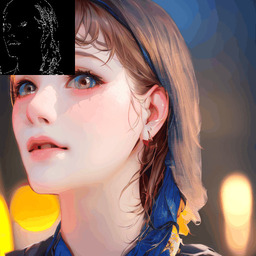} 
    \hfill
    \includegraphics[width=0.12\textwidth]{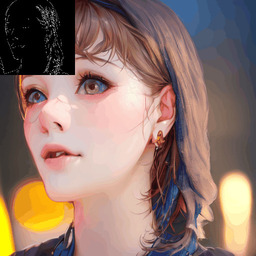} 
    \hfill
    \includegraphics[width=0.12\textwidth]{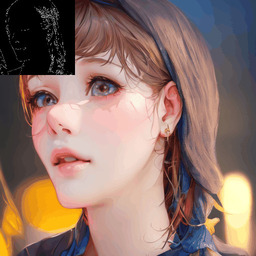} 
    \caption{professional photograph of 1girl style, ((detailed face)), (High Detail), Sharp, 8k, ((bokeh))}
    \end{subfigure}
    \vskip\baselineskip

    \begin{subfigure}{\textwidth}
    \includegraphics[width=0.12\textwidth]{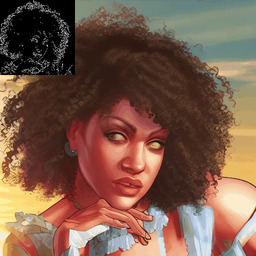} 
    \hfill
    \includegraphics[width=0.12\textwidth]{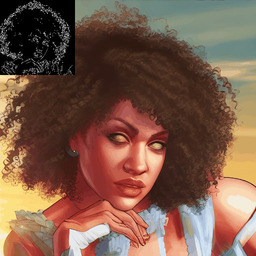} 
    \hfill
    \includegraphics[width=0.12\textwidth]{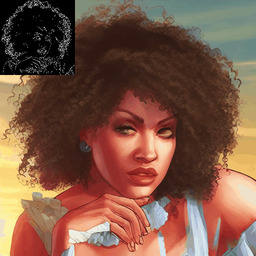} 
    \hfill
    \includegraphics[width=0.12\textwidth]{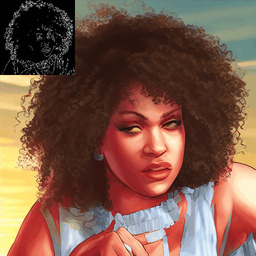} 
    \hfill
    \includegraphics[width=0.12\textwidth]{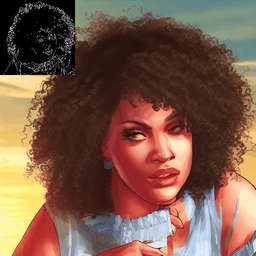} 
    \hfill
    \includegraphics[width=0.12\textwidth]{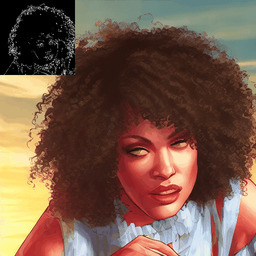} 
    \hfill
    \includegraphics[width=0.12\textwidth]{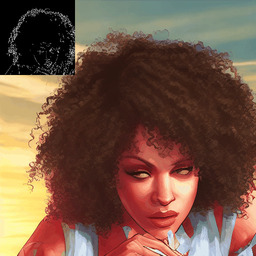} 
    \hfill
    \includegraphics[width=0.12\textwidth]{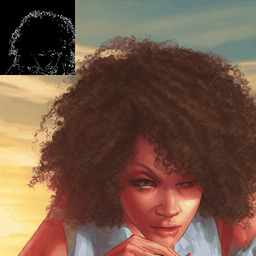} 
    \caption{gtav style}
    \end{subfigure}
    \vskip\baselineskip

    \begin{subfigure}{\textwidth}
    \includegraphics[width=0.12\textwidth]{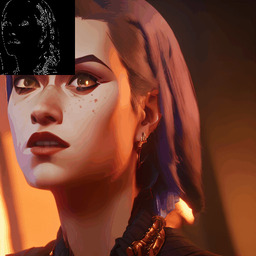} 
    \hfill
    \includegraphics[width=0.12\textwidth]{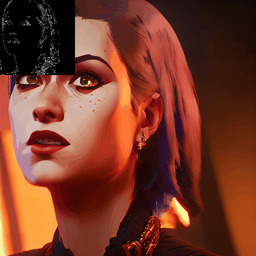} 
    \hfill
    \includegraphics[width=0.12\textwidth]{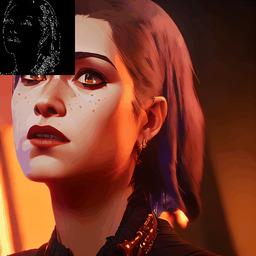} 
    \hfill
    \includegraphics[width=0.12\textwidth]{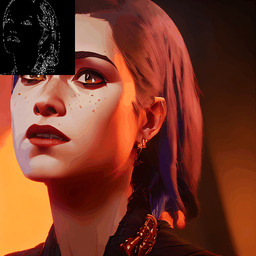} 
    \hfill
    \includegraphics[width=0.12\textwidth]{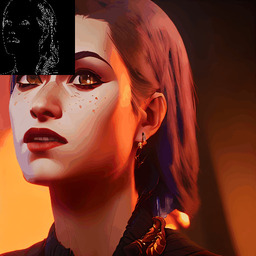} 
    \hfill
    \includegraphics[width=0.12\textwidth]{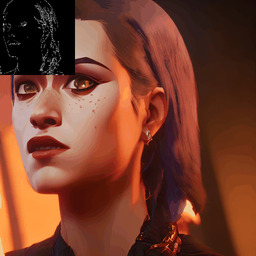} 
    \hfill
    \includegraphics[width=0.12\textwidth]{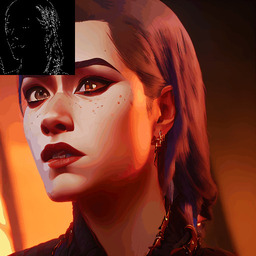} 
    \hfill
    \includegraphics[width=0.12\textwidth]{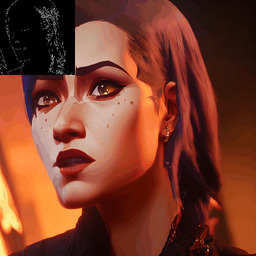} 
    \caption{professional photograph of arcane style, ((detailed face)), (High Detail), Sharp, 8k, ((bokeh))}
    \end{subfigure}
    \vskip\baselineskip

    \begin{subfigure}{\textwidth}
    \includegraphics[width=0.12\textwidth]{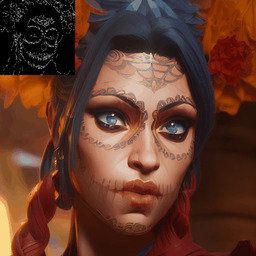} 
    \hfill
    \includegraphics[width=0.12\textwidth]{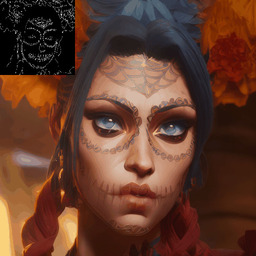} 
    \hfill
    \includegraphics[width=0.12\textwidth]{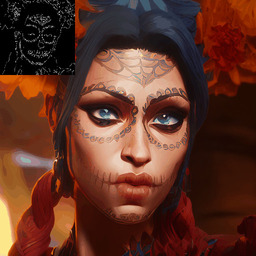} 
    \hfill
    \includegraphics[width=0.12\textwidth]{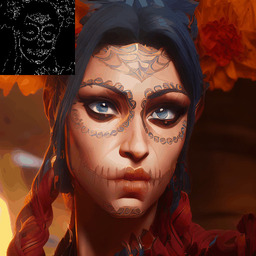} 
    \hfill
    \includegraphics[width=0.12\textwidth]{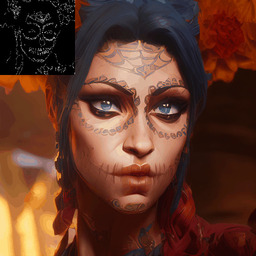} 
    \hfill
    \includegraphics[width=0.12\textwidth]{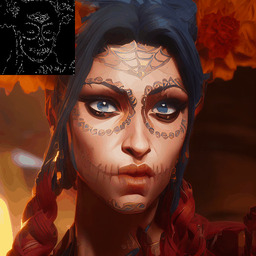} 
    \hfill
    \includegraphics[width=0.12\textwidth]{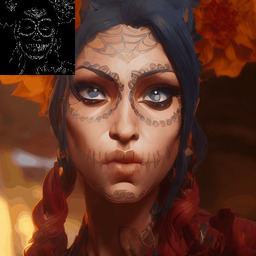} 
    \hfill
    \includegraphics[width=0.12\textwidth]{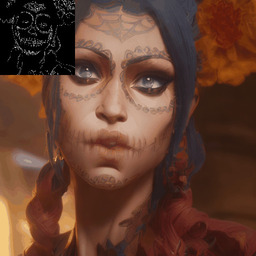} 
    \caption{A stunning intricate full color portrait of arcane style, epic character composition, sharp focus, natural lighting, subsurface scattering, f2, 35mm, film grain}
    \end{subfigure}
    \vskip\baselineskip

    \caption{Conditional generation with edge control and DreamBooth \cite{ruiz2022dreambooth} models. The keywords ``1girl style", ``gtav", ``arcane style", and ``arcane style" correspond to the personalized models of \href{https://civitai.com/models/8740/superanime-viper}{Anime Style}, \href{https://civitai.com/models/1309/gta5-artwork-diffusion}{GTA style}, and \href{https://civitai.com/models/23/arcane-diffusion}{Arcane Style}.}
    \label{fig:qual_results_full_method_edge_guidance_db}
\end{figure*}

%% file: supplemental_chapters/text2video_pose_guidance.tex
\section{Text-to-Video with Pose Guidance}
\label{apendix:t2v_pose_guidance}

\begin{figure*}
    \centering
    \begin{subfigure}{\textwidth}
    \includegraphics[width=0.12\textwidth]{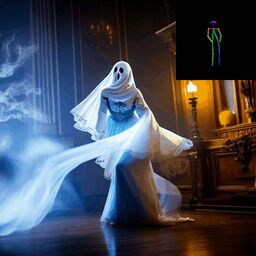} 
    \hfill
    \includegraphics[width=0.12\textwidth]{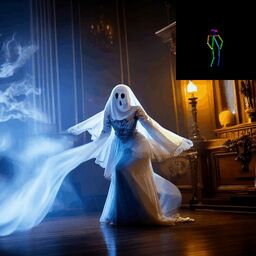} 
    \hfill
    \includegraphics[width=0.12\textwidth]{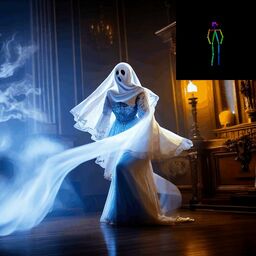} 
    \hfill
    \includegraphics[width=0.12\textwidth]{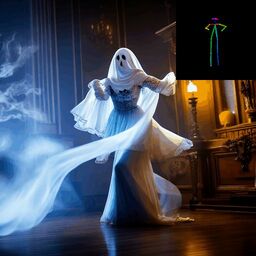} 
    \hfill
    \includegraphics[width=0.12\textwidth]{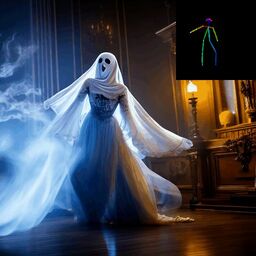} 
    \hfill
    \includegraphics[width=0.12\textwidth]{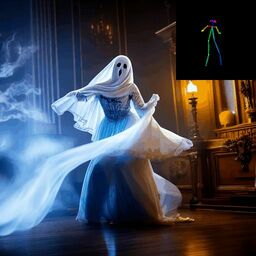} 
    \hfill
    \includegraphics[width=0.12\textwidth]{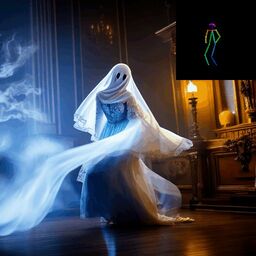} 
    \hfill
    \includegraphics[width=0.12\textwidth]{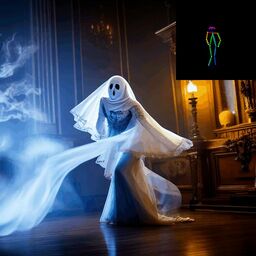} 
    \caption{a ghost dancing in a haunted mansion}
    \end{subfigure}
    \vskip\baselineskip

    \begin{subfigure}{\textwidth}
    \includegraphics[width=0.12\textwidth]{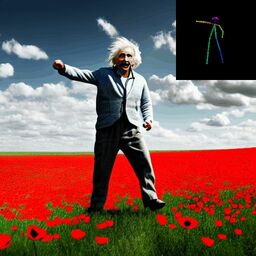} 
    \hfill
    \includegraphics[width=0.12\textwidth]{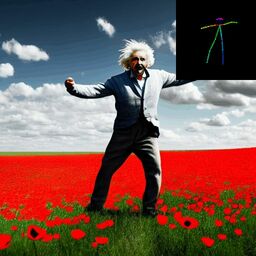} 
    \hfill
    \includegraphics[width=0.12\textwidth]{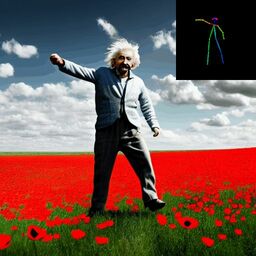} 
    \hfill
    \includegraphics[width=0.12\textwidth]{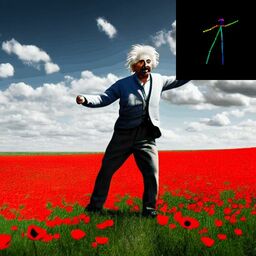} 
    \hfill
    \includegraphics[width=0.12\textwidth]{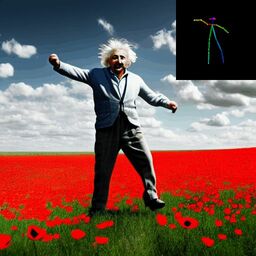} 
    \hfill
    \includegraphics[width=0.12\textwidth]{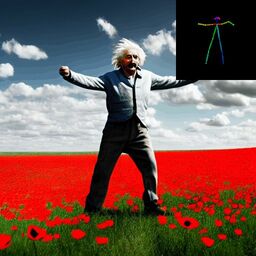} 
    \hfill
    \includegraphics[width=0.12\textwidth]{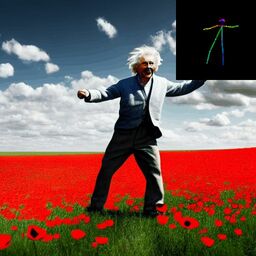} 
    \hfill
    \includegraphics[width=0.12\textwidth]{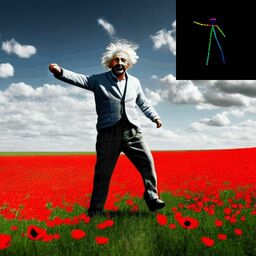} 
    \caption{Albert Einstein dancing in a field of poppies}
    \end{subfigure}
    \vskip\baselineskip

    \begin{subfigure}{\textwidth}
    \includegraphics[width=0.12\textwidth]{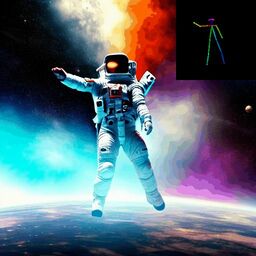} 
    \hfill
    \includegraphics[width=0.12\textwidth]{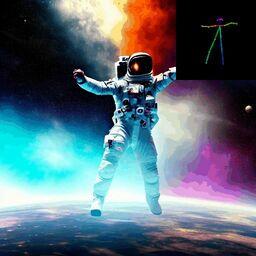} 
    \hfill
    \includegraphics[width=0.12\textwidth]{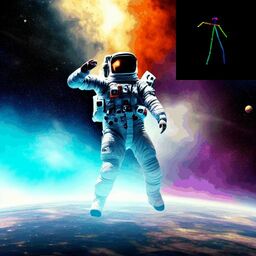} 
    \hfill
    \includegraphics[width=0.12\textwidth]{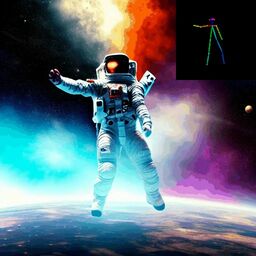} 
    \hfill
    \includegraphics[width=0.12\textwidth]{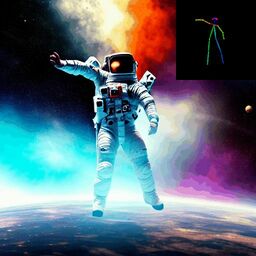} 
    \hfill
    \includegraphics[width=0.12\textwidth]{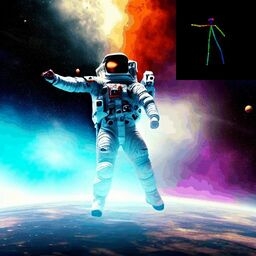} 
    \hfill
    \includegraphics[width=0.12\textwidth]{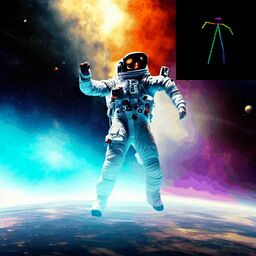} 
    \hfill
    \includegraphics[width=0.12\textwidth]{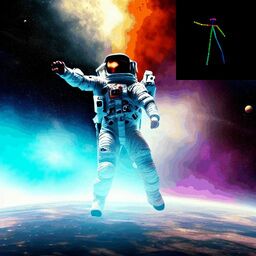} 
    \caption{an astronaut dancing in the outer space}
    \end{subfigure}
    \vskip\baselineskip


    \begin{subfigure}{\textwidth}
    \includegraphics[width=0.12\textwidth]{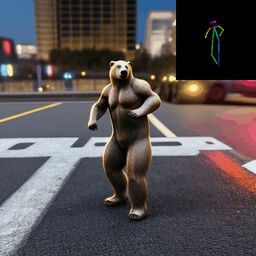} 
    \hfill
    \includegraphics[width=0.12\textwidth]{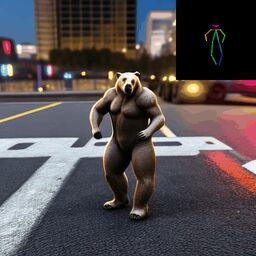} 
    \hfill
    \includegraphics[width=0.12\textwidth]{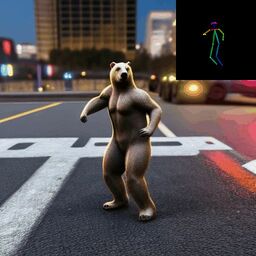} 
    \hfill
    \includegraphics[width=0.12\textwidth]{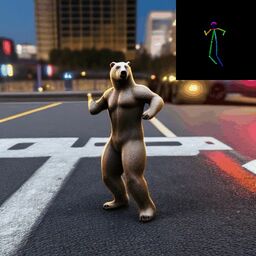} 
    \hfill
    \includegraphics[width=0.12\textwidth]{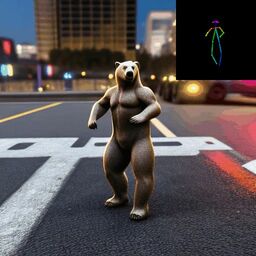} 
    \hfill
    \includegraphics[width=0.12\textwidth]{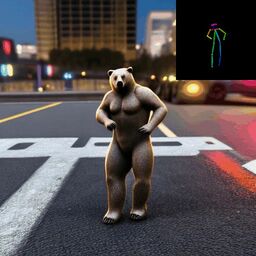} 
    \hfill
    \includegraphics[width=0.12\textwidth]{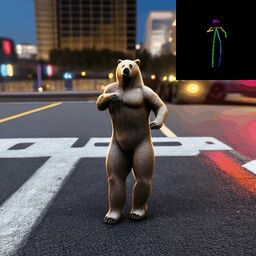} 
    \hfill
    \includegraphics[width=0.12\textwidth]{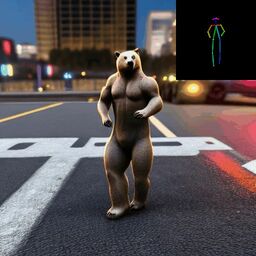} 
    \caption{a bear dancing on the concrete}
    \end{subfigure}
    \vskip\baselineskip

    \begin{subfigure}{\textwidth}
    \includegraphics[width=0.12\textwidth]{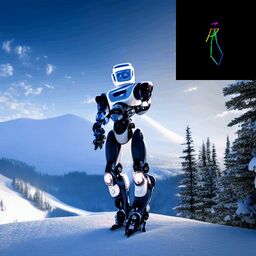} 
    \hfill
    \includegraphics[width=0.12\textwidth]{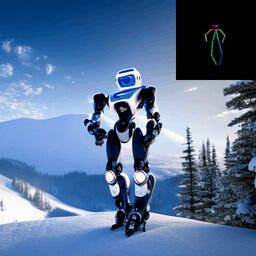} 
    \hfill
    \includegraphics[width=0.12\textwidth]{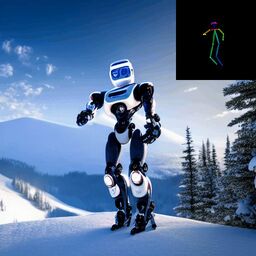} 
    \hfill
    \includegraphics[width=0.12\textwidth]{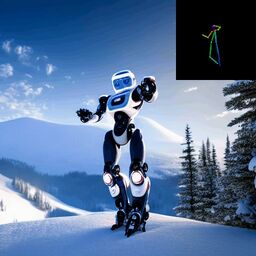} 
    \hfill
    \includegraphics[width=0.12\textwidth]{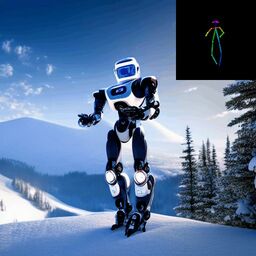} 
    \hfill
    \includegraphics[width=0.12\textwidth]{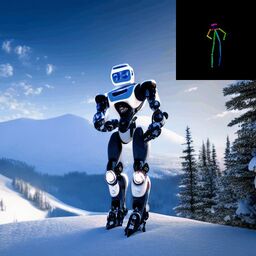} 
    \hfill
    \includegraphics[width=0.12\textwidth]{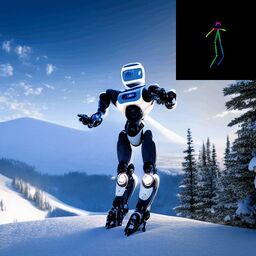} 
    \hfill
    \includegraphics[width=0.12\textwidth]{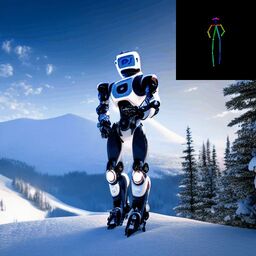} 
    \caption{a robot dancing on top of a snowy mountain}
    \end{subfigure}
    \vskip\baselineskip


    \caption{Conditional generation with pose guidance.}
    \label{fig:qual_results_pose}
\end{figure*}

In Fig.~\ref{fig:qual_results_pose} we present additional results of our method guided by pose information. 
In Fig.~\ref{fig:table_of_pose_study} we show the effect of our cross-frame attention and motion information in latents.

%% file: supplemental_chapters/text2video_guidance_ablation.tex
\setcounter{table}{\value{figure}}  
\begin{table*}
\captionsetup{name=Figure}
        \centering
        \begin{tabular}{ cM{18mm} M{ 18mm}M{ 18mm}M{ 18mm}M{ 18mm}M{18mm}M{18mm}M{18mm}M{18mm}}
            \vspace{0.2cm}\hspace{-0.5cm}\rotatebox[origin=c]{90}{\mbox{\begin{varwidth}{5cm} \scriptsize No Motion In latents. \\ No CF-Attention\end{varwidth}}} &  
            {\includegraphics[width=.11\textwidth]{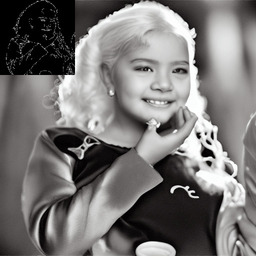}} &  
            {\includegraphics[width=.11\textwidth]{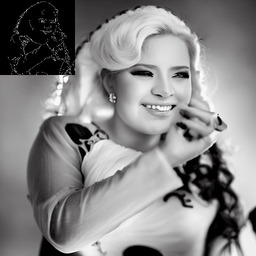}} &  
            {\includegraphics[width=.11\textwidth]{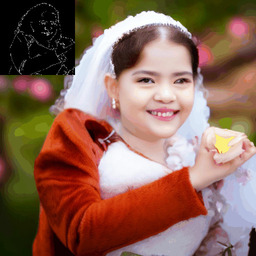}} &  
            {\includegraphics[width=.11\textwidth]{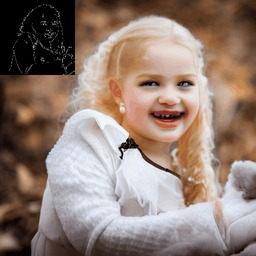}} &  
            {\includegraphics[width=.11\textwidth]{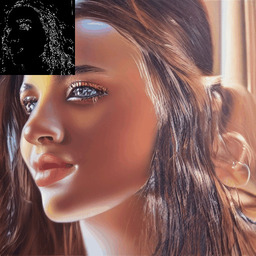}} &  
            {\includegraphics[width=.11\textwidth]{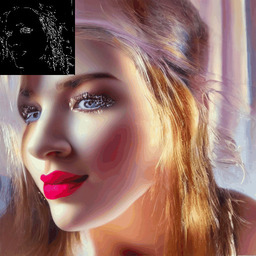}} &  
            {\includegraphics[width=.11\textwidth]{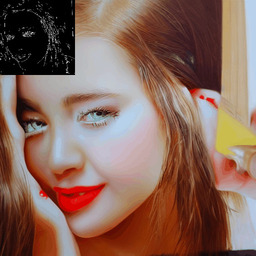}} &  
            {\includegraphics[width=.11\textwidth]{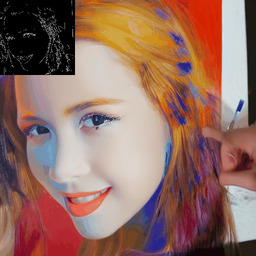}} &   \\  

            \vspace{0.2cm}\hspace{-0.5cm}\rotatebox[origin=c]{90}{\mbox{\begin{varwidth}{5cm} \scriptsize Motion In latents. \\ No CF-Attention\end{varwidth}}} &  
            {\includegraphics[width=.11\textwidth]{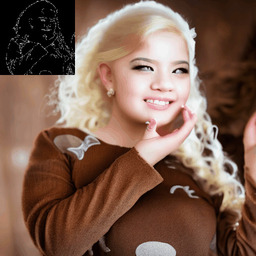}} &  
            {\includegraphics[width=.11\textwidth]{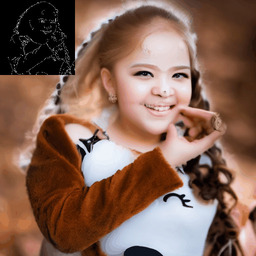}} &  
            {\includegraphics[width=.11\textwidth]{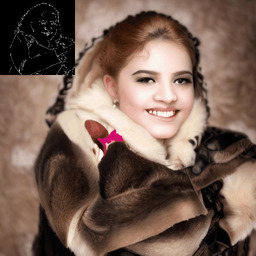}} &  
            {\includegraphics[width=.11\textwidth]{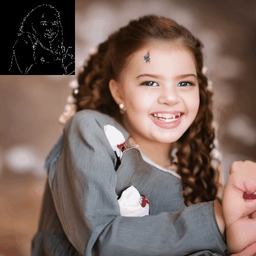}} &  
            {\includegraphics[width=.11\textwidth]{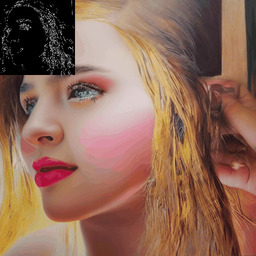}} &  
            {\includegraphics[width=.11\textwidth]{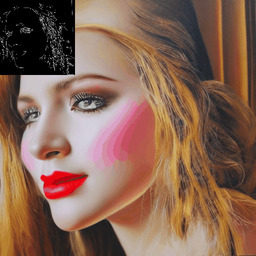}} &  
            {\includegraphics[width=.11\textwidth]{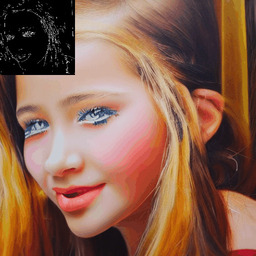}} &  
            {\includegraphics[width=.11\textwidth]{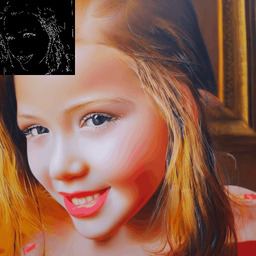}} &   \\ 

            \vspace{0.2cm}\hspace{-0.5cm}\rotatebox[origin=c]{90}{\mbox{\begin{varwidth}{5cm} \scriptsize No Motion In latents. \\  CF-Attention\end{varwidth}}} &  
            {\includegraphics[width=.11\textwidth]{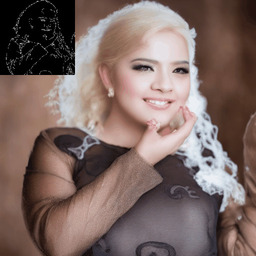}} &  
            {\includegraphics[width=.11\textwidth]{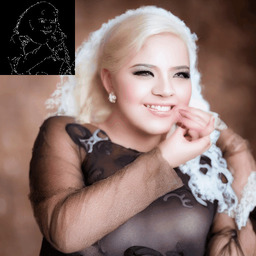}} &  
            {\includegraphics[width=.11\textwidth]{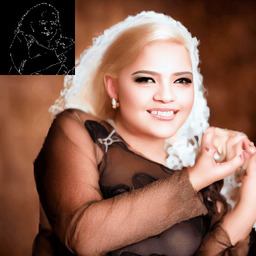}} &  
            {\includegraphics[width=.11\textwidth]{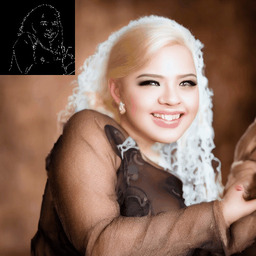}} &  
            {\includegraphics[width=.11\textwidth]{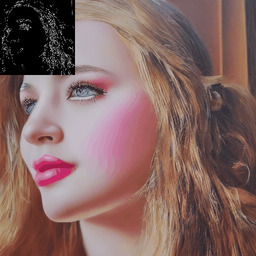}} &  
            {\includegraphics[width=.11\textwidth]{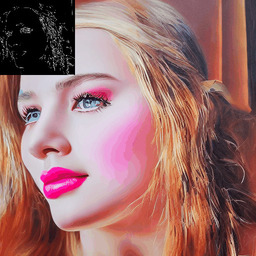}} &  
            {\includegraphics[width=.11\textwidth]{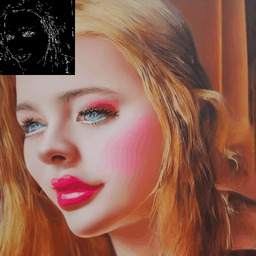}} &  
            {\includegraphics[width=.11\textwidth]{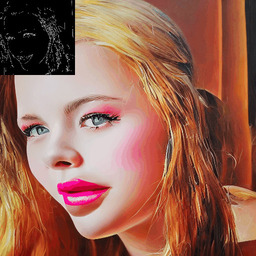}} &    \\ 

            \hspace{-0.5cm}\rotatebox[origin=c]{90}{\mbox{\begin{varwidth}{5cm} \scriptsize Motion In latents. \\ CF-Attention\end{varwidth}}} &  
            {\includegraphics[width=.11\textwidth]{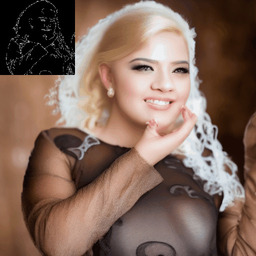}} &  
            {\includegraphics[width=.11\textwidth]{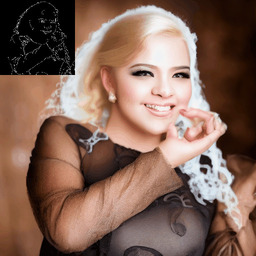}} &  
            {\includegraphics[width=.11\textwidth]{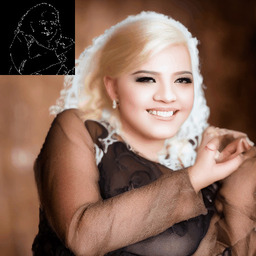}} &  
            {\includegraphics[width=.11\textwidth]{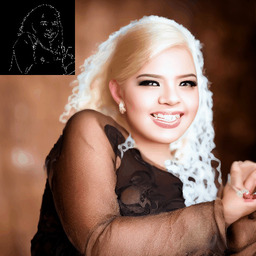}} &  
            {\includegraphics[width=.11\textwidth]{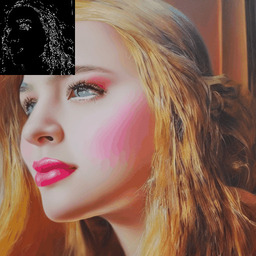}} &  
            {\includegraphics[width=.11\textwidth]{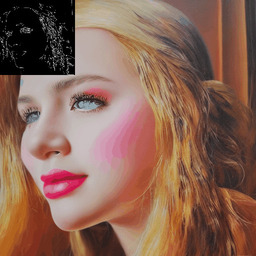}} &  
            {\includegraphics[width=.11\textwidth]{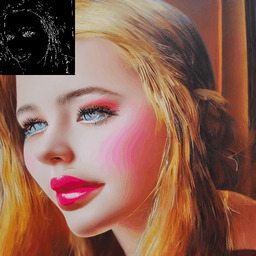}} &  
            {\includegraphics[width=.11\textwidth]{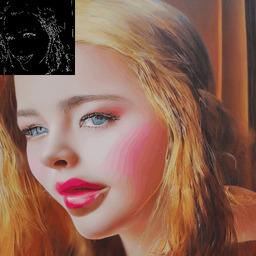}} &  \\ 
        
        \end{tabular}
        \caption{Ablation study on our CF-Attn block and adding motion information in latents for edge-guided video generation. The left half of each row is generated with the text prompt ``A beautiful girl". The right half of each row is generated with the text prompt ``oil painting of a girl."}
        \label{fig:table_of_edge_study}
    \end{table*}
\setcounter{figure}{\value{table}}

\begin{table*}
\captionsetup{name=Figure}
        \centering
        \begin{tabular}{ cM{18mm} M{ 18mm}M{ 18mm}M{ 18mm}M{ 18mm}M{18mm}M{18mm}M{18mm}M{18mm}}
            \vspace{0.2cm}\hspace{-0.5cm}\rotatebox[origin=c]{90}{\mbox{\begin{varwidth}{5cm} \scriptsize No Motion In latents. \\ No CF-Attention\end{varwidth}}} &  
            {\includegraphics[width=.11\textwidth]{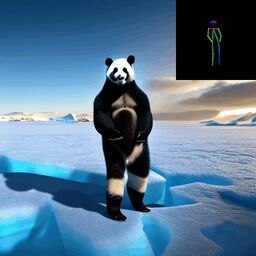}} &  
            {\includegraphics[width=.11\textwidth]{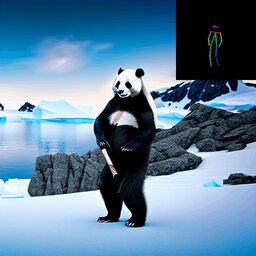}} &  
            {\includegraphics[width=.11\textwidth]{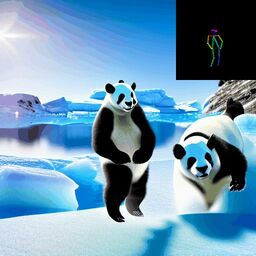}} &  
            {\includegraphics[width=.11\textwidth]{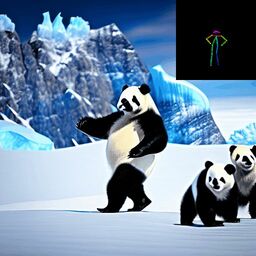}} &  
            {\includegraphics[width=.11\textwidth]{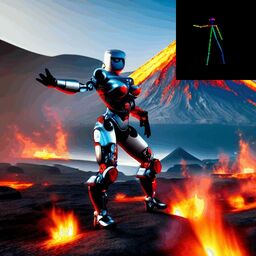}} &  
            {\includegraphics[width=.11\textwidth]{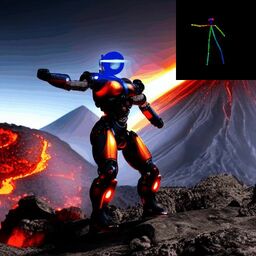}} &  
            {\includegraphics[width=.11\textwidth]{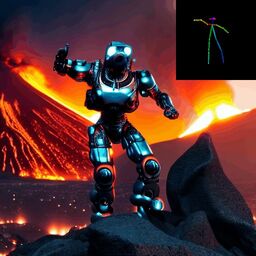}} &  
            {\includegraphics[width=.11\textwidth]{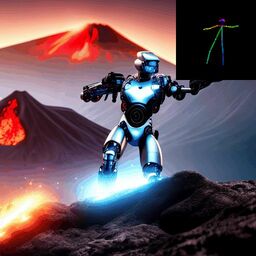}} &   \\  

            \vspace{0.2cm}\hspace{-0.5cm}\rotatebox[origin=c]{90}{\mbox{\begin{varwidth}{5cm} \scriptsize Motion In latents. \\ No CF-Attention\end{varwidth}}} &  
            {\includegraphics[width=.11\textwidth]{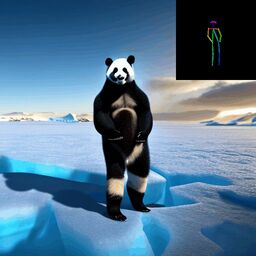}} &  
            {\includegraphics[width=.11\textwidth]{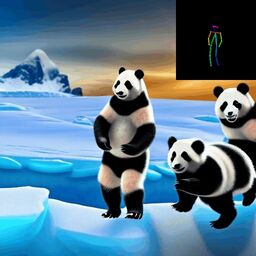}} &  
            {\includegraphics[width=.11\textwidth]{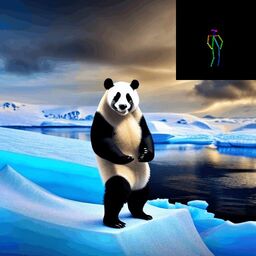}} &  
            {\includegraphics[width=.11\textwidth]{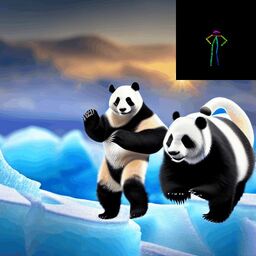}} &  
            {\includegraphics[width=.11\textwidth]{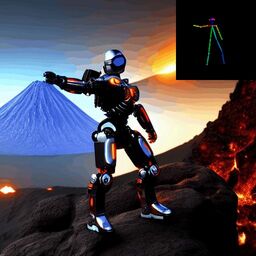}} &  
            {\includegraphics[width=.11\textwidth]{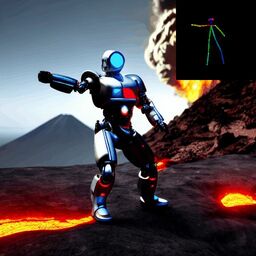}} &  
            {\includegraphics[width=.11\textwidth]{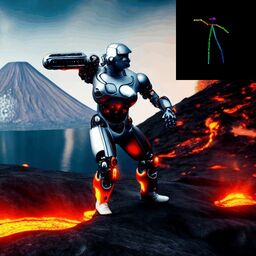}} &  
            {\includegraphics[width=.11\textwidth]{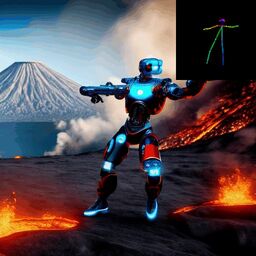}} &   \\ 

            \vspace{0.2cm}\hspace{-0.5cm}\rotatebox[origin=c]{90}{\mbox{\begin{varwidth}{5cm} \scriptsize No Motion In latents. \\  CF-Attention\end{varwidth}}} &  
            {\includegraphics[width=.11\textwidth]{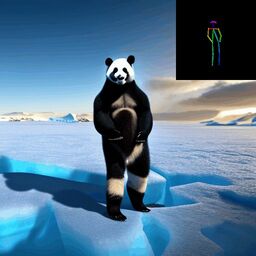}} &  
            {\includegraphics[width=.11\textwidth]{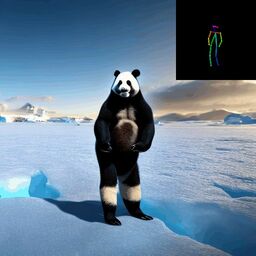}} &  
            {\includegraphics[width=.11\textwidth]{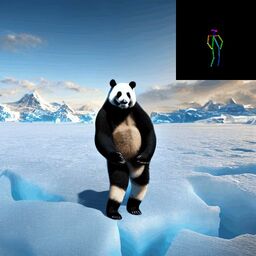}} &  
            {\includegraphics[width=.11\textwidth]{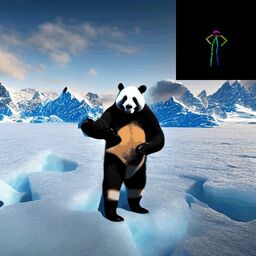}} &  
            {\includegraphics[width=.11\textwidth]{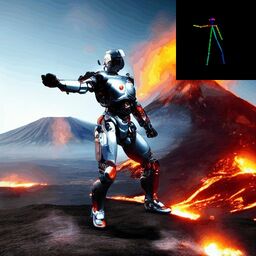}} &  
            {\includegraphics[width=.11\textwidth]{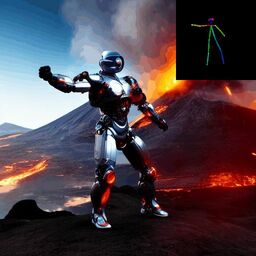}} &  
            {\includegraphics[width=.11\textwidth]{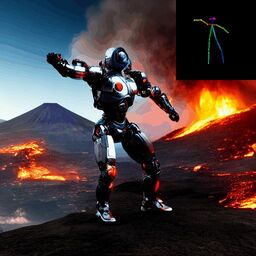}} &  
            {\includegraphics[width=.11\textwidth]{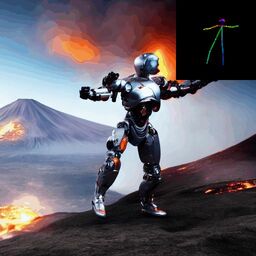}} &    \\ 

            \hspace{-0.5cm}\rotatebox[origin=c]{90}{\mbox{\begin{varwidth}{5cm} \scriptsize Motion In latents. \\ CF-Attention\end{varwidth}}} &  
            {\includegraphics[width=.11\textwidth]{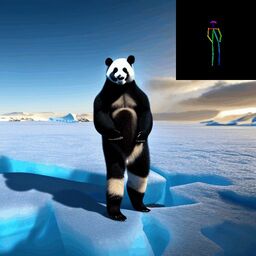}} &  
            {\includegraphics[width=.11\textwidth]{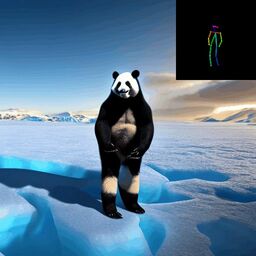}} &  
            {\includegraphics[width=.11\textwidth]{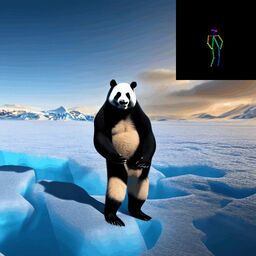}} &  
            {\includegraphics[width=.11\textwidth]{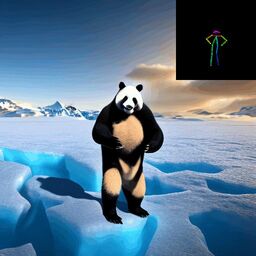}} &  
            {\includegraphics[width=.11\textwidth]{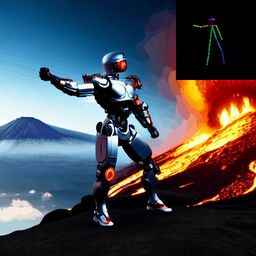}} &  
            {\includegraphics[width=.11\textwidth]{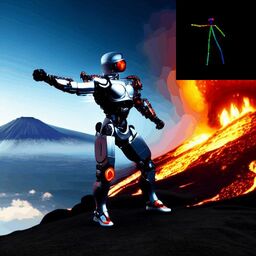}} &  
            {\includegraphics[width=.11\textwidth]{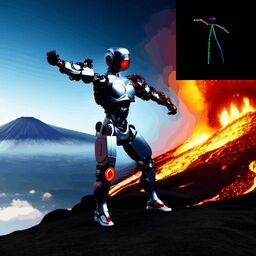}} &  
            {\includegraphics[width=.11\textwidth]{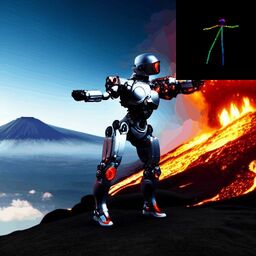}} &  \\ 
        
        \end{tabular}
        \caption{
        Ablation study on our CF-Attn block and adding motion information in latents for pose-guided video generation. The left half of each row is generated with the text prompt ``a panda dancing in Antarctica". The right half of each row is generated with the text prompt ``a cyborg dancing near a volcano".
        }
        \label{fig:table_of_pose_study}
    \end{table*}
\setcounter{figure}{\value{table}}

%% file: supplemental_chapters/text2video_editing.tex
\section{Video Instruct-Pix2Pix}
\label{apendix:t2v_editing}

\begin{figure*}
    \centering
    \begin{subfigure}{\textwidth}     \centering
    \includegraphics[width=0.11\textwidth]{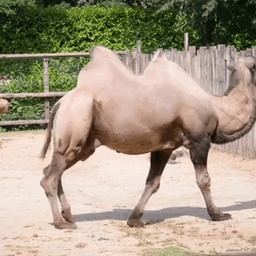}
    \includegraphics[width=0.11\textwidth]{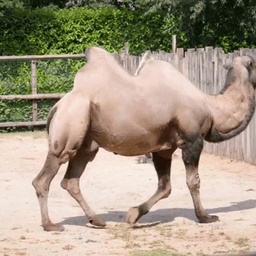}
    \includegraphics[width=0.11\textwidth]{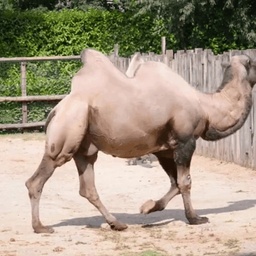}
    \includegraphics[width=0.11\textwidth]{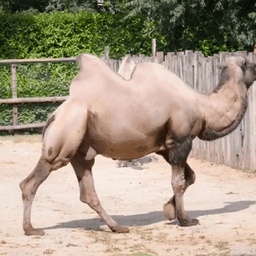}
    \includegraphics[width=0.11\textwidth]{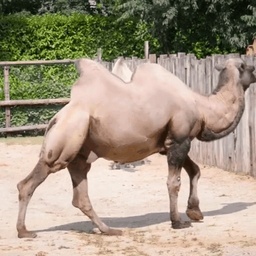}
    \includegraphics[width=0.11\textwidth]{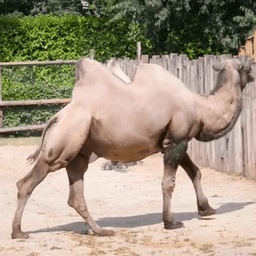}
    \includegraphics[width=0.11\textwidth]{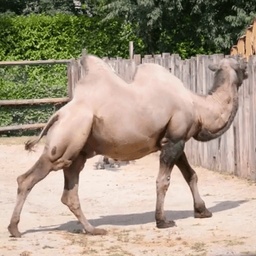}
    \includegraphics[width=0.11\textwidth]{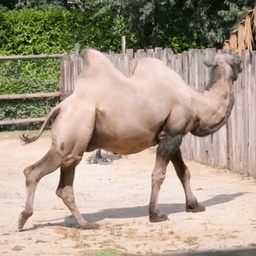} 
    \end{subfigure}
    
    \begin{subfigure}{\textwidth}     \centering
    \includegraphics[width=0.11\textwidth]{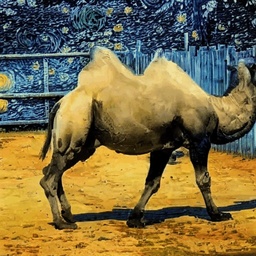}
    \includegraphics[width=0.11\textwidth]{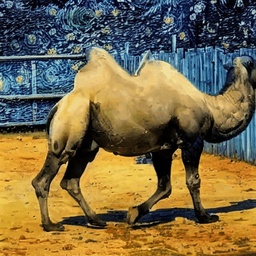}
    \includegraphics[width=0.11\textwidth]{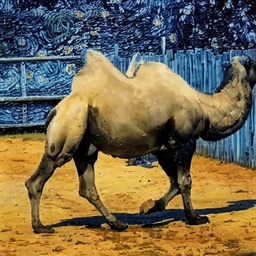}
    \includegraphics[width=0.11\textwidth]{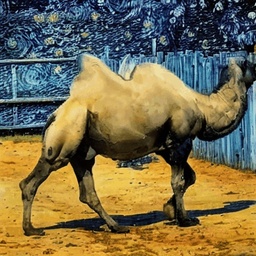}
    \includegraphics[width=0.11\textwidth]{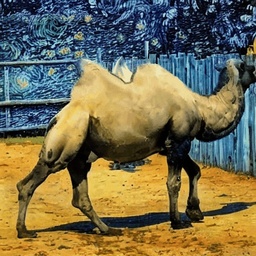}
    \includegraphics[width=0.11\textwidth]{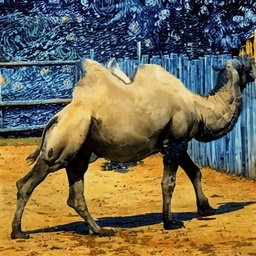}
    \includegraphics[width=0.11\textwidth]{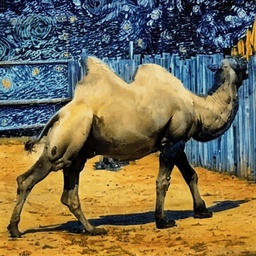}
    \includegraphics[width=0.11\textwidth]{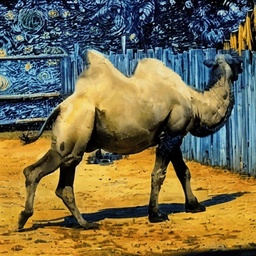} 
    \caption{Instruction: ``make it Van Gogh Starry Night style''}
    \end{subfigure}
    
    \begin{subfigure}{\textwidth}     \centering
    \includegraphics[width=0.11\textwidth]{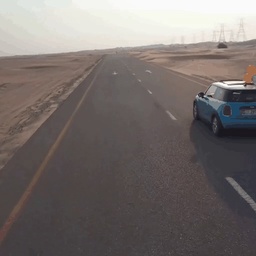}
    \includegraphics[width=0.11\textwidth]{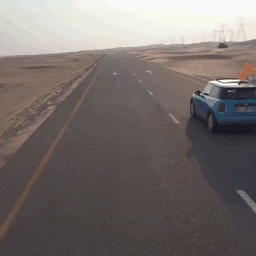}
    \includegraphics[width=0.11\textwidth]{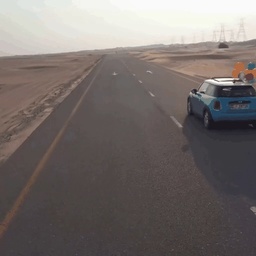}
    \includegraphics[width=0.11\textwidth]{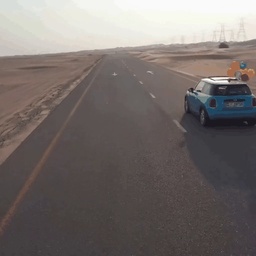}
    \includegraphics[width=0.11\textwidth]{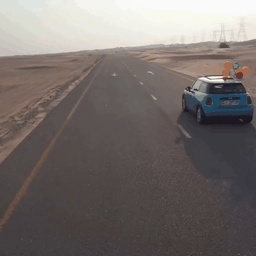}
    \includegraphics[width=0.11\textwidth]{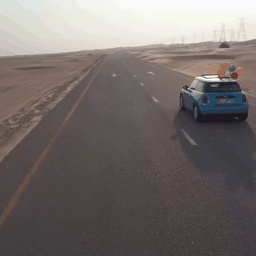}
    \includegraphics[width=0.11\textwidth]{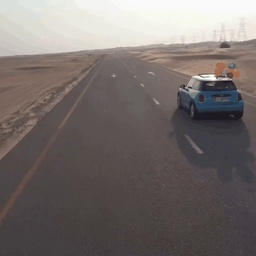}
    \includegraphics[width=0.11\textwidth]{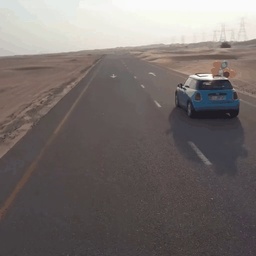}
    
    \end{subfigure}
    
    \begin{subfigure}{\textwidth}     \centering
    \includegraphics[width=0.11\textwidth]{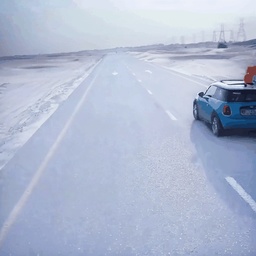}
    \includegraphics[width=0.11\textwidth]{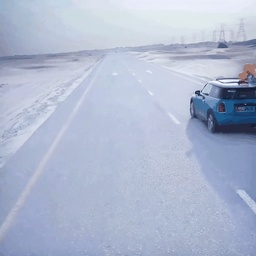}
    \includegraphics[width=0.11\textwidth]{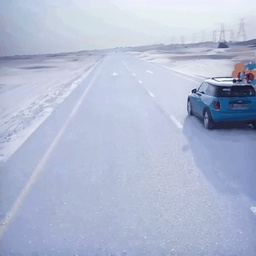}
    \includegraphics[width=0.11\textwidth]{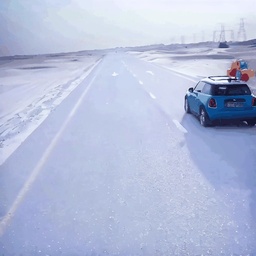}
    \includegraphics[width=0.11\textwidth]{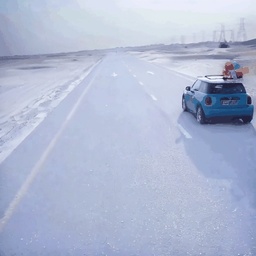}
    \includegraphics[width=0.11\textwidth]{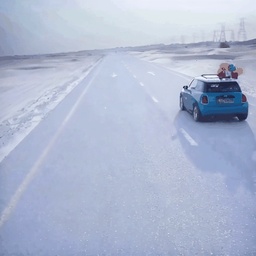}
    \includegraphics[width=0.11\textwidth]{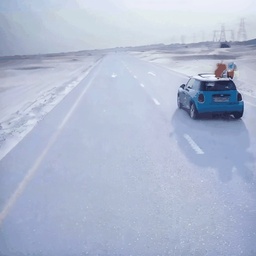}
    \includegraphics[width=0.11\textwidth]{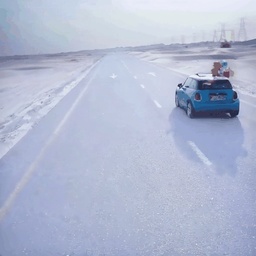}
    \caption{Instruction: ``make it snowy''}
    \end{subfigure}
    
    \begin{subfigure}{\textwidth}     \centering
    \includegraphics[width=0.11\textwidth]{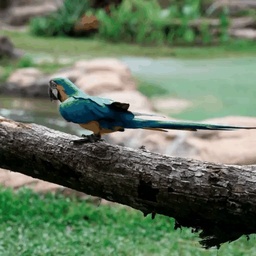}
    \includegraphics[width=0.11\textwidth]{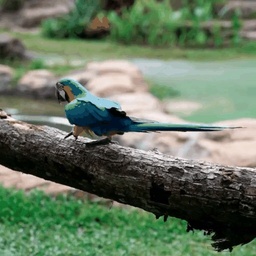}
    \includegraphics[width=0.11\textwidth]{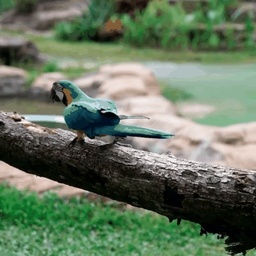}
    \includegraphics[width=0.11\textwidth]{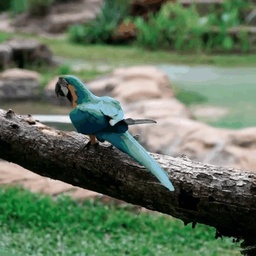}
    \includegraphics[width=0.11\textwidth]{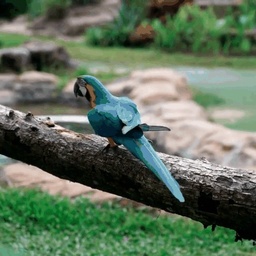}
    \includegraphics[width=0.11\textwidth]{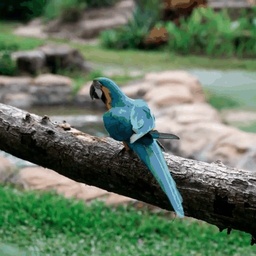}
    \includegraphics[width=0.11\textwidth]{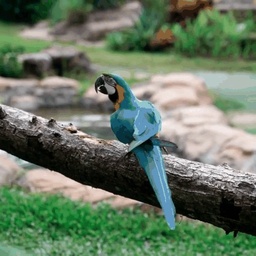}
    \includegraphics[width=0.11\textwidth]{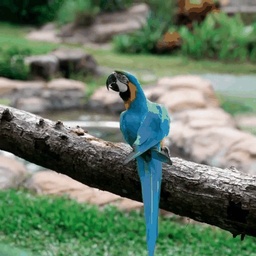}
    \end{subfigure}
    
    \begin{subfigure}{\textwidth}     \centering
    \includegraphics[width=0.11\textwidth]{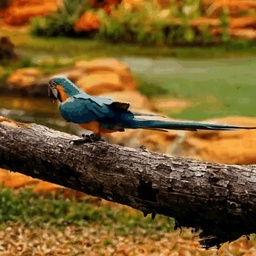}
    \includegraphics[width=0.11\textwidth]{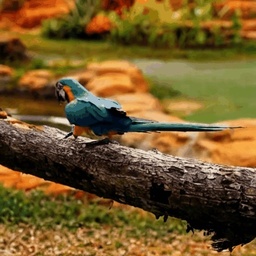}
    \includegraphics[width=0.11\textwidth]{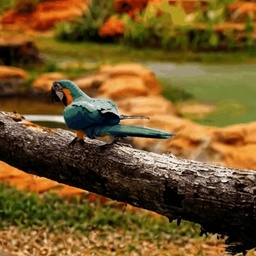}
    \includegraphics[width=0.11\textwidth]{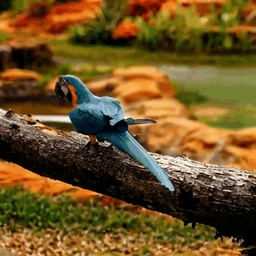}
    \includegraphics[width=0.11\textwidth]{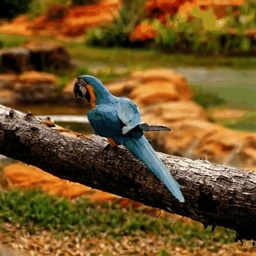}
    \includegraphics[width=0.11\textwidth]{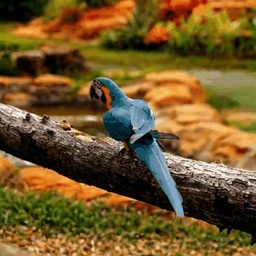}
    \includegraphics[width=0.11\textwidth]{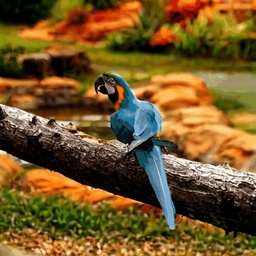}
    \includegraphics[width=0.11\textwidth]{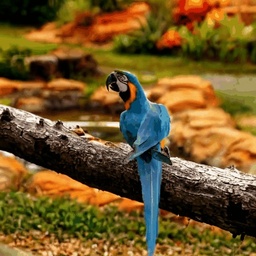}
    \caption{Instruction: ``make it autumn''}
    \end{subfigure}
    
    \begin{subfigure}{\textwidth}     \centering
    \includegraphics[width=0.11\textwidth]{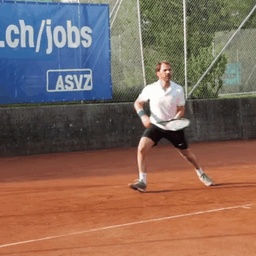}
    \includegraphics[width=0.11\textwidth]{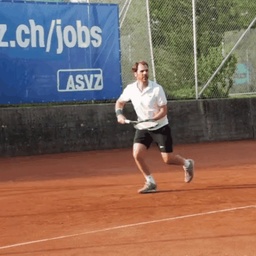}
    \includegraphics[width=0.11\textwidth]{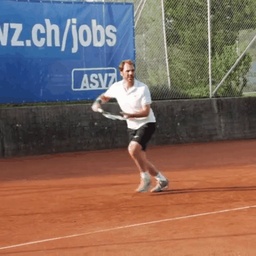}
    \includegraphics[width=0.11\textwidth]{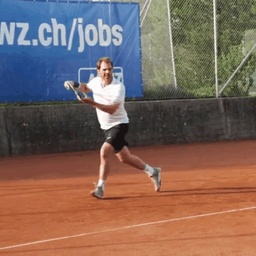}
    \includegraphics[width=0.11\textwidth]{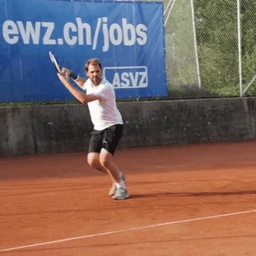}
    \includegraphics[width=0.11\textwidth]{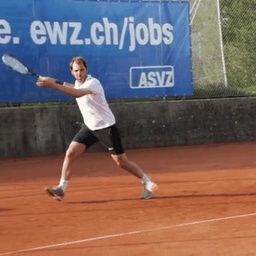}
    \includegraphics[width=0.11\textwidth]{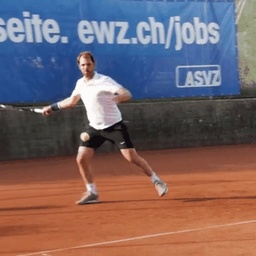}
    \includegraphics[width=0.11\textwidth]{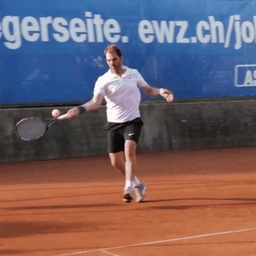} 
    \end{subfigure}
    
    \begin{subfigure}{\textwidth}     \centering
    \includegraphics[width=0.11\textwidth]{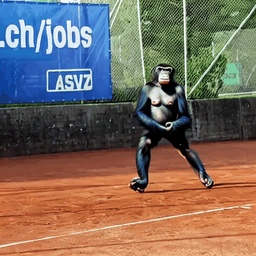}
    \includegraphics[width=0.11\textwidth]{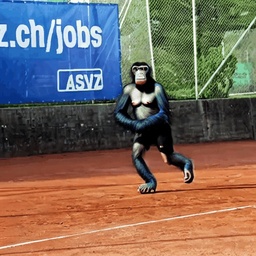}
    \includegraphics[width=0.11\textwidth]{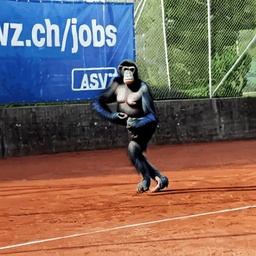}
    \includegraphics[width=0.11\textwidth]{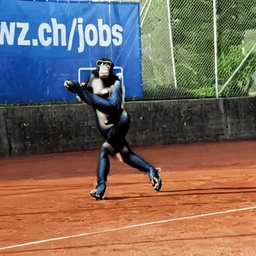}
    \includegraphics[width=0.11\textwidth]{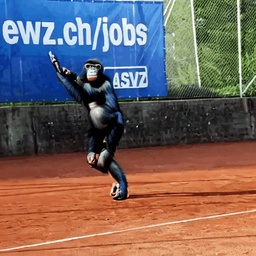}
    \includegraphics[width=0.11\textwidth]{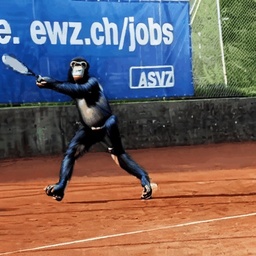}
    \includegraphics[width=0.11\textwidth]{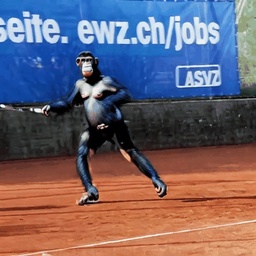}
    \includegraphics[width=0.11\textwidth]{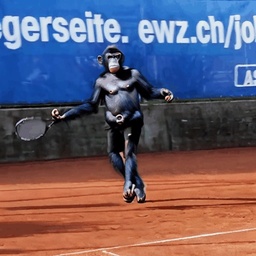} 
    \caption{Instruction: ``replace man with chimpanzee''}
    \end{subfigure}
    
    \begin{subfigure}{\textwidth}     \centering
    \includegraphics[width=0.11\textwidth]{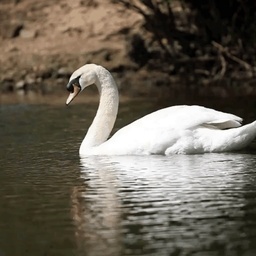}
    \includegraphics[width=0.11\textwidth]{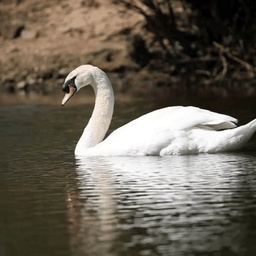}
    \includegraphics[width=0.11\textwidth]{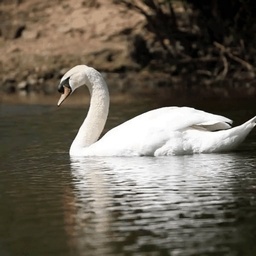}
    \includegraphics[width=0.11\textwidth]{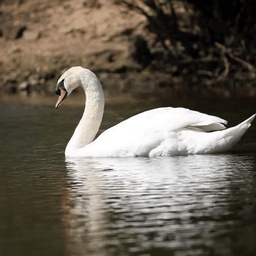}
    \includegraphics[width=0.11\textwidth]{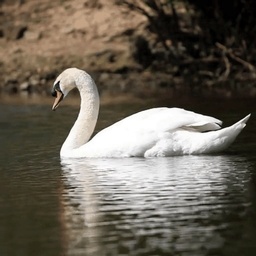}
    \includegraphics[width=0.11\textwidth]{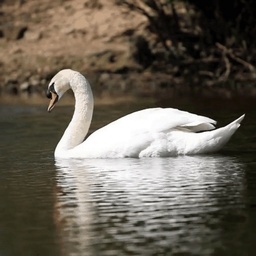}
    \includegraphics[width=0.11\textwidth]{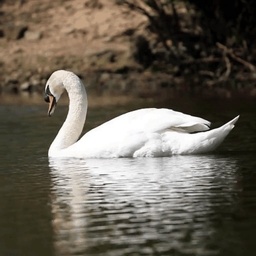}
    \includegraphics[width=0.11\textwidth]{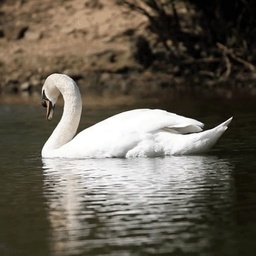}
    

    \end{subfigure}
    
    \begin{subfigure}{\textwidth}     \centering
    \includegraphics[width=0.11\textwidth]{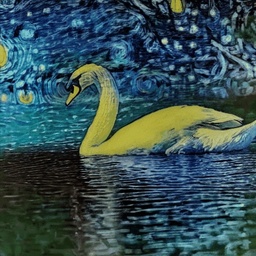}
    \includegraphics[width=0.11\textwidth]{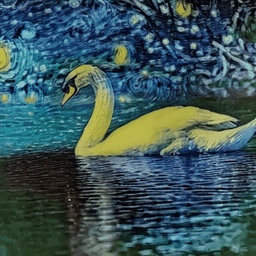}
    \includegraphics[width=0.11\textwidth]{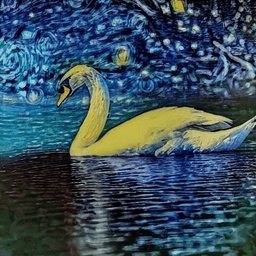}
    \includegraphics[width=0.11\textwidth]{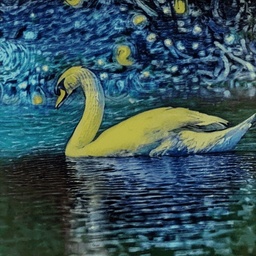}
    \includegraphics[width=0.11\textwidth]{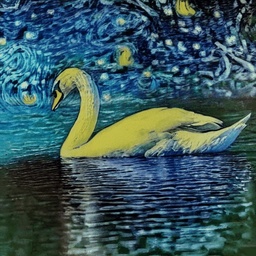}
    \includegraphics[width=0.11\textwidth]{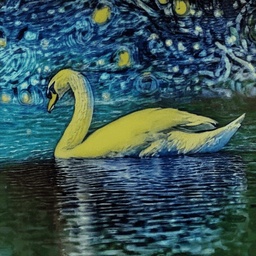}
    \includegraphics[width=0.11\textwidth]{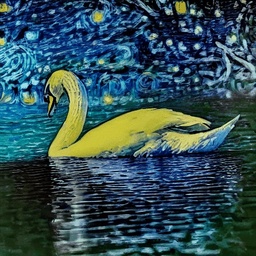}
    \includegraphics[width=0.11\textwidth]{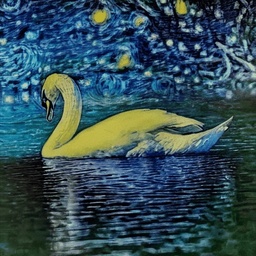}
    \caption{Instruction: ``make it Van Gogh Starry Night style"}
    \end{subfigure}
    \caption{Text-guided video editing using our model combined with Instruct-Pix2Pix \cite{brooks2022instructpix2pix}. Instructions for editing are described under each video sequence.}
    \label{fig:video_editing_results}
\end{figure*}

In Fig.~\ref{fig:video_editing_results} we present additional results of instruct-guided video editing by using our approach combined with Instruct-Pix2Pix \cite{brooks2022instructpix2pix}. 
As shown in Figures \ref{fig:table_of_EDITING/comparison1} and \ref{fig:table_of_EDITING/comparison2} our method outperforms naive per-frame approach of Instruct-Pix2Pix and a recent state-of-the-art method Tune-A-Video \cite{tune-a-video}.
Particularly, while being semantically aware of text-guided edits, Tune-A-Video has limitations in localized editing, and struggles to transfer the style and color information.
On the other hand Instruct-Pix2Pix makes visually plausible edits on image level but has issues with temporal consistency.
In contrast with the mentioned approaches our method preserves the temporal consistency when editing videos by given prompts.

%% file: supplemental_chapters/text2video_editing_comparison.tex
\setcounter{table}{\value{figure}}  
\begin{table*}
\captionsetup{name=Figure}
        \centering
        \begin{tabular}{ cM{18mm} M{ 18mm}M{ 18mm}M{ 18mm}M{ 18mm}M{18mm}M{18mm}M{18mm}M{18mm}}
            \vspace{0.2cm}\hspace{-0.5cm}\rotatebox[origin=c]{90}{\mbox{\begin{varwidth}{5cm} \scriptsize Original\end{varwidth}}} &  
            {\includegraphics[width=.11\textwidth]{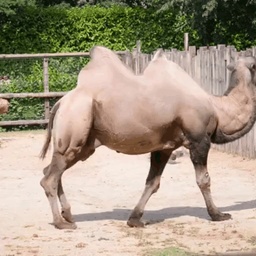}} &  
            {\includegraphics[width=.11\textwidth]{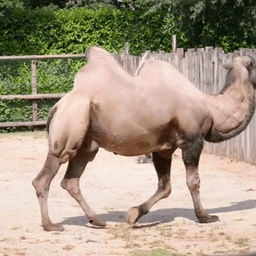}} &  
            {\includegraphics[width=.11\textwidth]{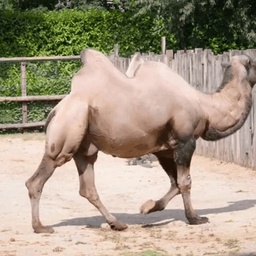}} &  
            {\includegraphics[width=.11\textwidth]{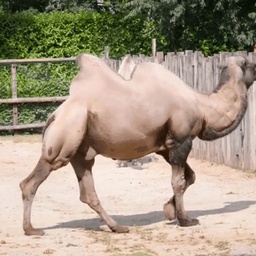}} &  
            {\includegraphics[width=.11\textwidth]{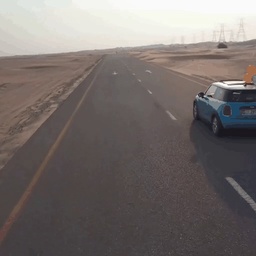}} &  
            {\includegraphics[width=.11\textwidth]{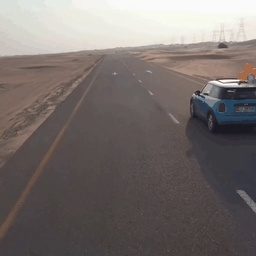}} &  
            {\includegraphics[width=.11\textwidth]{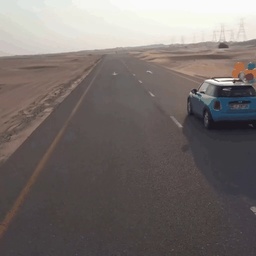}} &  
            {\includegraphics[width=.11\textwidth]{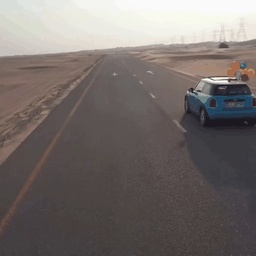}} &   \\  

            \vspace{0.2cm}\hspace{-0.5cm}\rotatebox[origin=c]{90}{\mbox{\begin{varwidth}{5cm} \scriptsize Our\end{varwidth}}} &  
            {\includegraphics[width=.11\textwidth]{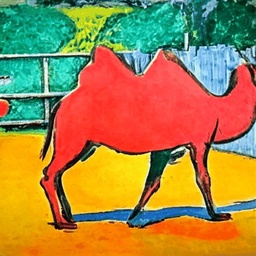}} &  
            {\includegraphics[width=.11\textwidth]{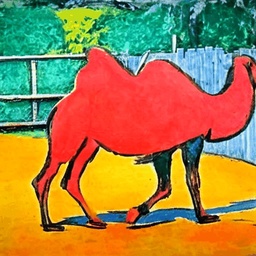}} &  
            {\includegraphics[width=.11\textwidth]{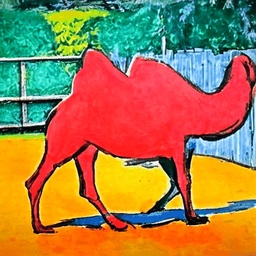}} &  
            {\includegraphics[width=.11\textwidth]{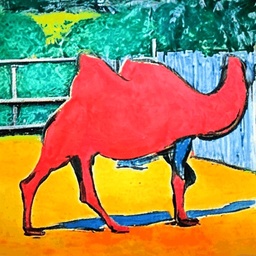}} &  
            {\includegraphics[width=.11\textwidth]{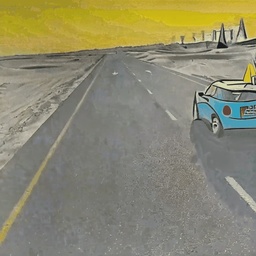}} &  
            {\includegraphics[width=.11\textwidth]{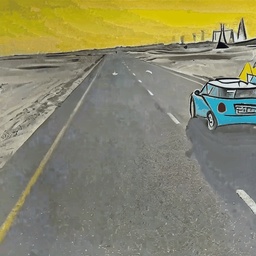}} &  
            {\includegraphics[width=.11\textwidth]{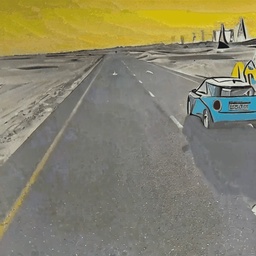}} &  
            {\includegraphics[width=.11\textwidth]{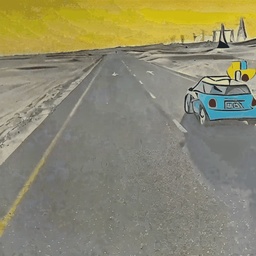}} &   \\  

            \vspace{0.2cm}\hspace{-0.5cm}\rotatebox[origin=c]{90}{\mbox{\begin{varwidth}{5cm} \scriptsize Instruct-Pix2Pix\end{varwidth}}} &  
            {\includegraphics[width=.11\textwidth]{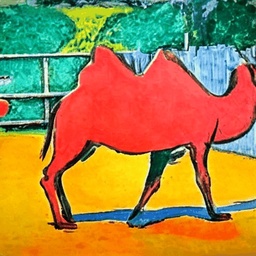}} &  
            {\includegraphics[width=.11\textwidth]{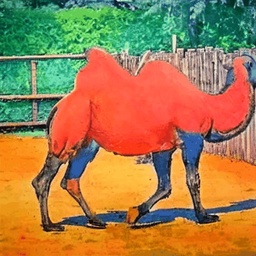}} &  
            {\includegraphics[width=.11\textwidth]{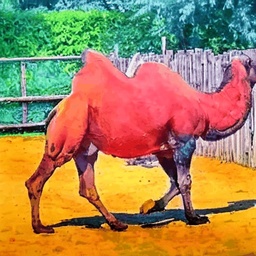}} &  
            {\includegraphics[width=.11\textwidth]{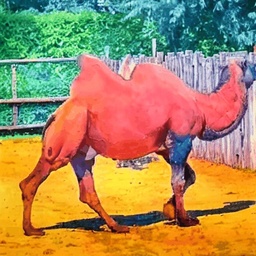}} &  
            {\includegraphics[width=.11\textwidth]{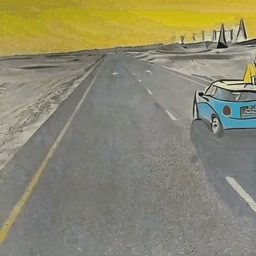}} &  
            {\includegraphics[width=.11\textwidth]{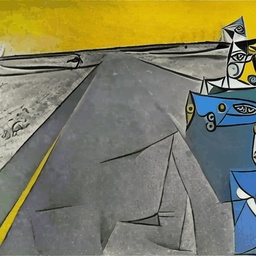}} &  
            {\includegraphics[width=.11\textwidth]{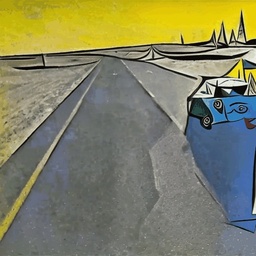}} &  
            {\includegraphics[width=.11\textwidth]{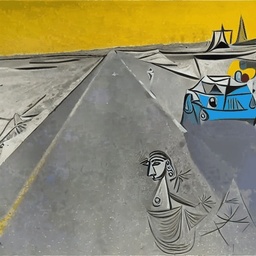}} &   \\  

            \hspace{-0.5cm}\rotatebox[origin=c]{90}{\mbox{\begin{varwidth}{5cm} \scriptsize Tune-A-Video\end{varwidth}}} &  
            {\includegraphics[width=.11\textwidth]{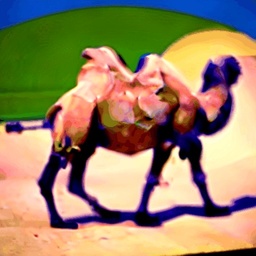}} &  
            {\includegraphics[width=.11\textwidth]{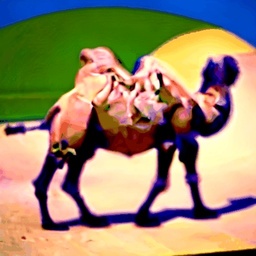}} &  
            {\includegraphics[width=.11\textwidth]{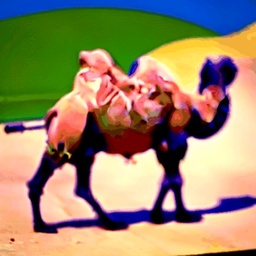}} &  
            {\includegraphics[width=.11\textwidth]{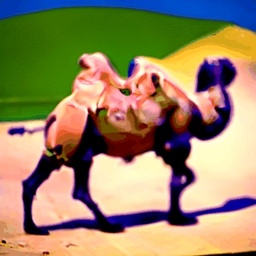}} &  
            {\includegraphics[width=.11\textwidth]{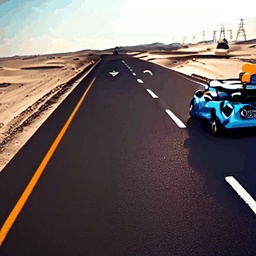}} &  
            {\includegraphics[width=.11\textwidth]{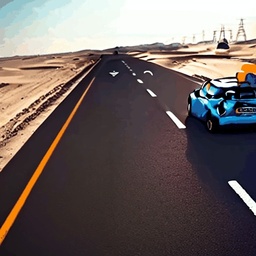}} &  
            {\includegraphics[width=.11\textwidth]{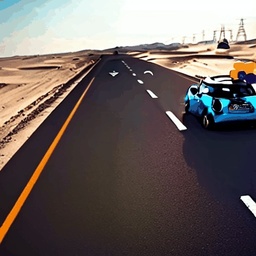}} &  
            {\includegraphics[width=.11\textwidth]{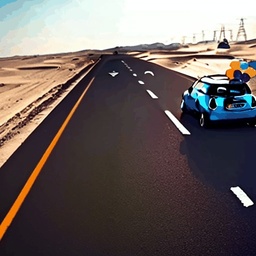}} &   \\  
        
        \end{tabular}
        \caption{Text guided video editing using our method compared to Instruct-Pix2Pix \cite{brooks2022instructpix2pix} frame-by-frame and Tune-A-Video \cite{tune-a-video}. The left half of each row is generated with the instruction ``make it Expressionism style" for our method and Instruct-Pix2Pix and ``a camel walking, \textbf{in Expressionism style}" for Tune-A-Video. The right half of each row is generated with the text prompt ``make it Picasso style" and ``a mini cooper riding on a road, \textbf{in Picasso style}" respectively.}
        \label{fig:table_of_EDITING/comparison1}
    \end{table*}
\setcounter{figure}{\value{table}}

\begin{table*}
\captionsetup{name=Figure}
        \centering
        \begin{tabular}{ cM{18mm} M{ 18mm}M{ 18mm}M{ 18mm}M{ 18mm}M{18mm}M{18mm}M{18mm}M{18mm}}
            \vspace{0.2cm}\hspace{-0.5cm}\rotatebox[origin=c]{90}{\mbox{\begin{varwidth}{5cm} \scriptsize Original\end{varwidth}}} &  
            {\includegraphics[width=.11\textwidth]{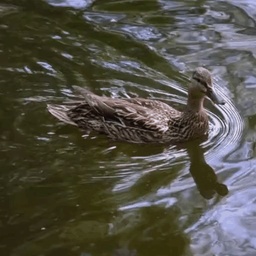}} &  
            {\includegraphics[width=.11\textwidth]{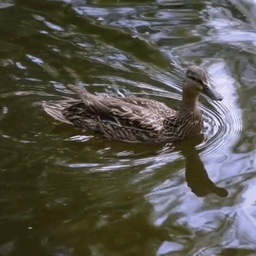}} &  
            {\includegraphics[width=.11\textwidth]{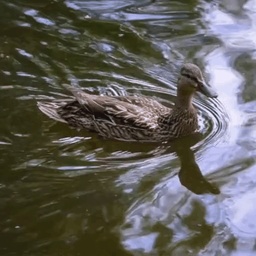}} &  
            {\includegraphics[width=.11\textwidth]{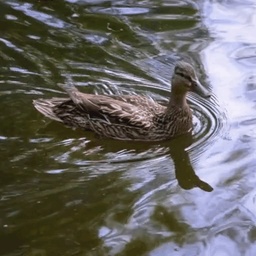}} &  
            {\includegraphics[width=.11\textwidth]{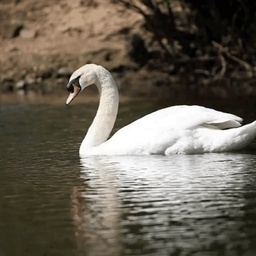}} &  
            {\includegraphics[width=.11\textwidth]{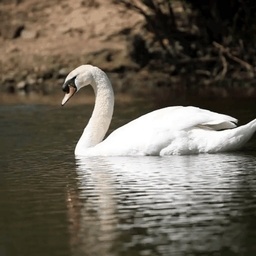}} &  
            {\includegraphics[width=.11\textwidth]{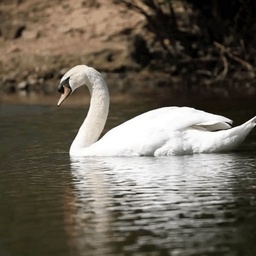}} &  
            {\includegraphics[width=.11\textwidth]{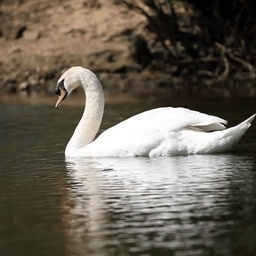}} &   \\  

            \vspace{0.2cm}\hspace{-0.5cm}\rotatebox[origin=c]{90}{\mbox{\begin{varwidth}{5cm} \scriptsize Our\end{varwidth}}} &  
            {\includegraphics[width=.11\textwidth]{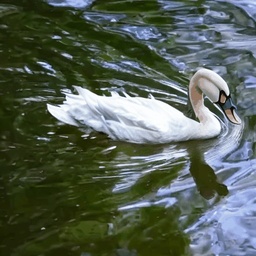}} &  
            {\includegraphics[width=.11\textwidth]{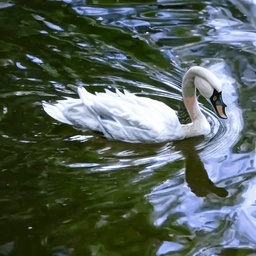}} &  
            {\includegraphics[width=.11\textwidth]{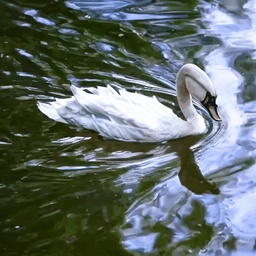}} &  
            {\includegraphics[width=.11\textwidth]{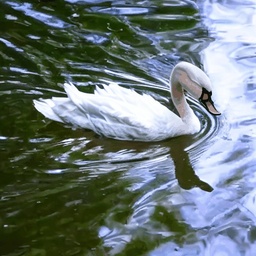}} &  
            {\includegraphics[width=.11\textwidth]{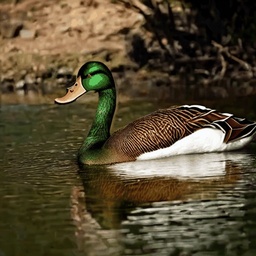}} &  
            {\includegraphics[width=.11\textwidth]{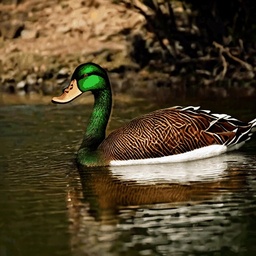}} &  
            {\includegraphics[width=.11\textwidth]{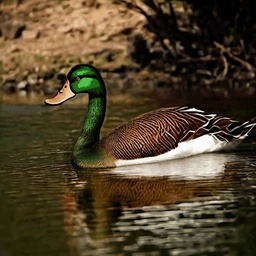}} &  
            {\includegraphics[width=.11\textwidth]{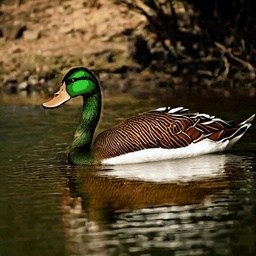}} &   \\  

            \vspace{0.2cm}\hspace{-0.5cm}\rotatebox[origin=c]{90}{\mbox{\begin{varwidth}{5cm} \scriptsize Instruct-Pix2Pix\end{varwidth}}} &  
            {\includegraphics[width=.11\textwidth]{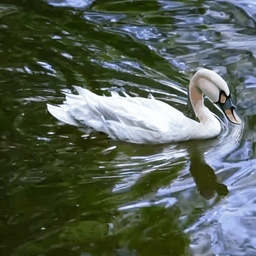}} &  
            {\includegraphics[width=.11\textwidth]{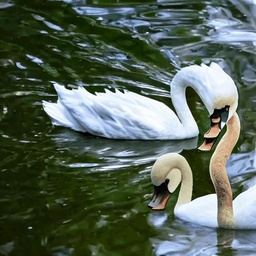}} &  
            {\includegraphics[width=.11\textwidth]{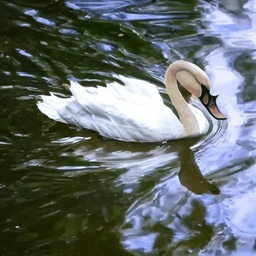}} &  
            {\includegraphics[width=.11\textwidth]{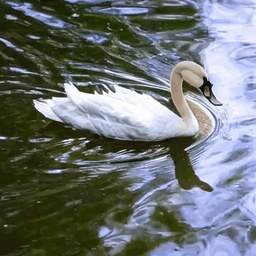}} &  
            {\includegraphics[width=.11\textwidth]{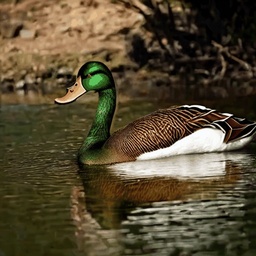}} &  
            {\includegraphics[width=.11\textwidth]{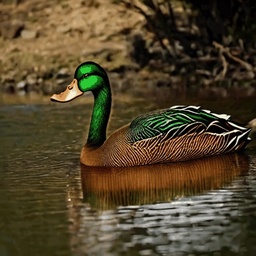}} &  
            {\includegraphics[width=.11\textwidth]{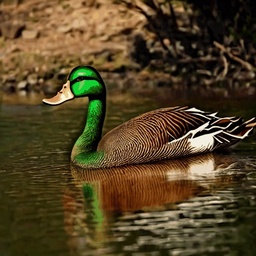}} &  
            {\includegraphics[width=.11\textwidth]{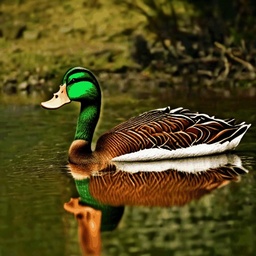}} &   \\  

            \hspace{-0.5cm}\rotatebox[origin=c]{90}{\mbox{\begin{varwidth}{5cm} \scriptsize Tune-A-Video\end{varwidth}}} &  
            {\includegraphics[width=.11\textwidth]{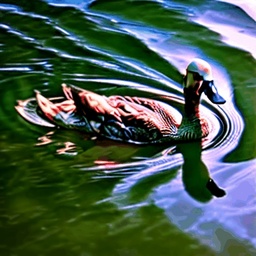}} &  
            {\includegraphics[width=.11\textwidth]{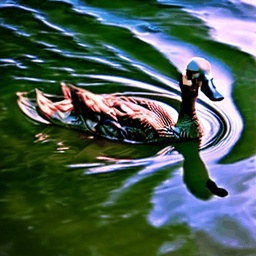}} &  
            {\includegraphics[width=.11\textwidth]{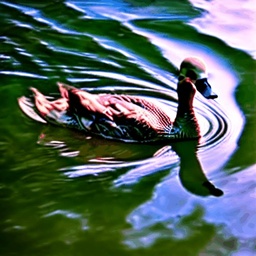}} &  
            {\includegraphics[width=.11\textwidth]{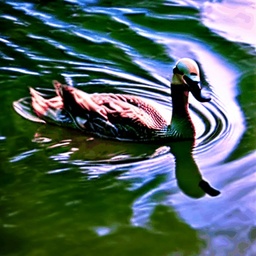}} &  
            {\includegraphics[width=.11\textwidth]{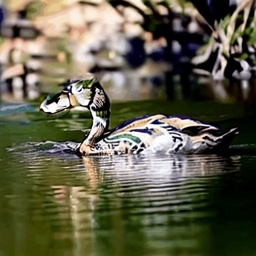}} &  
            {\includegraphics[width=.11\textwidth]{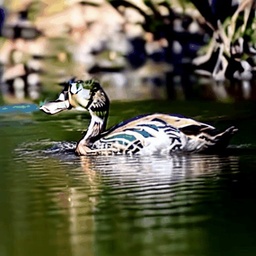}} &  
            {\includegraphics[width=.11\textwidth]{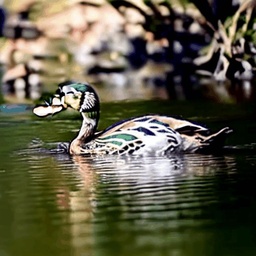}} &  
            {\includegraphics[width=.11\textwidth]{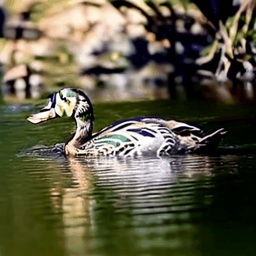}} &   \\  
        
        \end{tabular}
        \caption{Text guided video editing using our method compared to Instruct-Pix2Pix \cite{brooks2022instructpix2pix} frame-by-frame and Tune-A-Video \cite{tune-a-video}. The left half of each row is generated with the instruction ``replace mallard with swan" for our method and Instruct-Pix2Pix and ``a \textbf{swan} swimming on water" for Tune-A-Video. The right half of each row is generated with the text prompt ``replace swan with mallard" and ``a \textbf{mallard} swimming on water" respectively.}
        \label{fig:table_of_EDITING/comparison2}
    \end{table*}
\setcounter{figure}{\value{table}}

%% file: Text2Video-Zero.bbl
\begin{thebibliography}{10}\itemsep=-1pt

\bibitem{bar2022text2live}
Omer Bar-Tal, Dolev Ofri-Amar, Rafail Fridman, Yoni Kasten, and Tali Dekel.
\newblock Text2live: Text-driven layered image and video editing.
\newblock In {\em Computer Vision--ECCV 2022: 17th European Conference, Tel
  Aviv, Israel, October 23--27, 2022, Proceedings, Part XV}, pages 707--723.
  Springer, 2022.

\bibitem{brooks2022instructpix2pix}
Tim Brooks, Aleksander Holynski, and Alexei~A Efros.
\newblock Instructpix2pix: Learning to follow image editing instructions.
\newblock {\em arXiv preprint arXiv:2211.09800}, 2022.

\bibitem{dhariwal2021diffusion}
Prafulla Dhariwal and Alexander Nichol.
\newblock Diffusion models beat gans on image synthesis.
\newblock {\em Advances in Neural Information Processing Systems},
  34:8780--8794, 2021.

\bibitem{ding2022cogview2}
Ming Ding, Wendi Zheng, Wenyi Hong, and Jie Tang.
\newblock Cogview2: Faster and better text-to-image generation via hierarchical
  transformers.
\newblock {\em arXiv preprint arXiv:2204.14217}, 2022.

\bibitem{gen1_paper}
Patrick Esser, Johnathan Chiu, Parmida Atighehchian, Jonathan Granskog, and
  Anastasis Germanidis.
\newblock Structure and content-guided video synthesis with diffusion models.
\newblock {\em arXiv preprint arXiv:2302.03011}, 2023.

\bibitem{VQ_GAN_paper}
Patrick Esser, Robin Rombach, and Bjorn Ommer.
\newblock Taming transformers for high-resolution image synthesis.
\newblock In {\em Proceedings of the IEEE/CVF conference on computer vision and
  pattern recognition}, pages 12873--12883, 2021.

\bibitem{make-a-scene}
Oran Gafni, Adam Polyak, Oron Ashual, Shelly Sheynin, Devi Parikh, and Yaniv
  Taigman.
\newblock Make-a-scene: Scene-based text-to-image generation with human priors.
\newblock In {\em Computer Vision--ECCV 2022: 17th European Conference, Tel
  Aviv, Israel, October 23--27, 2022, Proceedings, Part XV}, pages 89--106.
  Springer, 2022.

\bibitem{goodfellow2020generative}
Ian Goodfellow, Jean Pouget-Abadie, Mehdi Mirza, Bing Xu, David Warde-Farley,
  Sherjil Ozair, Aaron Courville, and Yoshua Bengio.
\newblock Generative adversarial networks.
\newblock {\em Communications of the ACM}, 63(11):139--144, 2020.

\bibitem{prompt2prompt}
Amir Hertz, Ron Mokady, Jay Tenenbaum, Kfir Aberman, Yael Pritch, and Daniel
  Cohen-Or.
\newblock Prompt-to-prompt image editing with cross attention control.
\newblock {\em arXiv preprint arXiv:2208.01626}, 2022.

\bibitem{hessel2021clipscore}
Jack Hessel, Ari Holtzman, Maxwell Forbes, Ronan~Le Bras, and Yejin Choi.
\newblock Clipscore: A reference-free evaluation metric for image captioning.
\newblock {\em arXiv preprint arXiv:2104.08718}, 2021.

\bibitem{imagen_video}
Jonathan Ho, William Chan, Chitwan Saharia, Jay Whang, Ruiqi Gao, Alexey
  Gritsenko, Diederik~P Kingma, Ben Poole, Mohammad Norouzi, David~J Fleet,
  et~al.
\newblock Imagen video: High definition video generation with diffusion models.
\newblock {\em arXiv preprint arXiv:2210.02303}, 2022.

\bibitem{DDPM_paper}
Jonathan Ho, Ajay Jain, and Pieter Abbeel.
\newblock Denoising diffusion probabilistic models.
\newblock {\em Advances in Neural Information Processing Systems},
  33:6840--6851, 2020.

\bibitem{ho2022classifier}
Jonathan Ho and Tim Salimans.
\newblock Classifier-free diffusion guidance.
\newblock {\em arXiv preprint arXiv:2207.12598}, 2022.

\bibitem{video-diffusion-models}
Jonathan Ho, Tim Salimans, Alexey Gritsenko, William Chan, Mohammad Norouzi,
  and David~J Fleet.
\newblock Video diffusion models.
\newblock {\em arXiv preprint arXiv:2204.03458}, 2022.

\bibitem{hong2022cogvideo}
Wenyi Hong, Ming Ding, Wendi Zheng, Xinghan Liu, and Jie Tang.
\newblock Cogvideo: Large-scale pretraining for text-to-video generation via
  transformers.
\newblock {\em arXiv preprint arXiv:2205.15868}, 2022.

\bibitem{lee2023shape}
Yao-Chih Lee, Ji-Ze~Genevieve Jang, Yi-Ting Chen, Elizabeth Qiu, and Jia-Bin
  Huang.
\newblock Shape-aware text-driven layered video editing.
\newblock {\em arXiv e-prints}, pages arXiv--2301, 2023.

\bibitem{liu2023more}
Xihui Liu, Dong~Huk Park, Samaneh Azadi, Gong Zhang, Arman Chopikyan, Yuxiao
  Hu, Humphrey Shi, Anna Rohrbach, and Trevor Darrell.
\newblock More control for free! image synthesis with semantic diffusion
  guidance.
\newblock In {\em Proceedings of the IEEE/CVF Winter Conference on Applications
  of Computer Vision}, pages 289--299, 2023.

\bibitem{mansimov2015generating}
Elman Mansimov, Emilio Parisotto, Jimmy~Lei Ba, and Ruslan Salakhutdinov.
\newblock Generating images from captions with attention.
\newblock {\em arXiv preprint arXiv:1511.02793}, 2015.

\bibitem{meng2021sdedit}
Chenlin Meng, Yang Song, Jiaming Song, Jiajun Wu, Jun-Yan Zhu, and Stefano
  Ermon.
\newblock Sdedit: Image synthesis and editing with stochastic differential
  equations.
\newblock {\em arXiv preprint arXiv:2108.01073}, 2021.

\bibitem{molad2023dreamix}
Eyal Molad, Eliahu Horwitz, Dani Valevski, Alex~Rav Acha, Yossi Matias, Yael
  Pritch, Yaniv Leviathan, and Yedid Hoshen.
\newblock Dreamix: Video diffusion models are general video editors.
\newblock {\em arXiv preprint arXiv:2302.01329}, 2023.

\bibitem{mou2023t2i}
Chong Mou, Xintao Wang, Liangbin Xie, Jian Zhang, Zhongang Qi, Ying Shan, and
  Xiaohu Qie.
\newblock T2i-adapter: Learning adapters to dig out more controllable ability
  for text-to-image diffusion models.
\newblock {\em arXiv preprint arXiv:2302.08453}, 2023.

\bibitem{nichol2021glide}
Alex Nichol, Prafulla Dhariwal, Aditya Ramesh, Pranav Shyam, Pamela Mishkin,
  Bob McGrew, Ilya Sutskever, and Mark Chen.
\newblock Glide: Towards photorealistic image generation and editing with
  text-guided diffusion models.
\newblock {\em arXiv preprint arXiv:2112.10741}, 2021.

\bibitem{qiao2019mirrorgan}
Tingting Qiao, Jing Zhang, Duanqing Xu, and Dacheng Tao.
\newblock Mirrorgan: Learning text-to-image generation by redescription.
\newblock In {\em Proceedings of the IEEE/CVF Conference on Computer Vision and
  Pattern Recognition}, pages 1505--1514, 2019.

\bibitem{CLIP_paper}
Alec Radford, Jong~Wook Kim, Chris Hallacy, Aditya Ramesh, Gabriel Goh,
  Sandhini Agarwal, Girish Sastry, Amanda Askell, Pamela Mishkin, Jack Clark,
  et~al.
\newblock Learning transferable visual models from natural language
  supervision.
\newblock In {\em International conference on machine learning}, pages
  8748--8763. PMLR, 2021.

\bibitem{Dalle2_paper}
Aditya Ramesh, Prafulla Dhariwal, Alex Nichol, Casey Chu, and Mark Chen.
\newblock Hierarchical text-conditional image generation with clip latents.
\newblock {\em arXiv preprint arXiv:2204.06125}, 2022.

\bibitem{Dalle_paper}
Aditya Ramesh, Mikhail Pavlov, Gabriel Goh, Scott Gray, Chelsea Voss, Alec
  Radford, Mark Chen, and Ilya Sutskever.
\newblock Zero-shot text-to-image generation.
\newblock In {\em International Conference on Machine Learning}, pages
  8821--8831. PMLR, 2021.

\bibitem{reed2016learning}
Scott~E Reed, Zeynep Akata, Santosh Mohan, Samuel Tenka, Bernt Schiele, and
  Honglak Lee.
\newblock Learning what and where to draw.
\newblock {\em Advances in neural information processing systems}, 29, 2016.

\bibitem{stable_diff}
Robin Rombach, Andreas Blattmann, Dominik Lorenz, Patrick Esser, and Bj{\"o}rn
  Ommer.
\newblock High-resolution image synthesis with latent diffusion models.
\newblock In {\em Proceedings of the IEEE/CVF Conference on Computer Vision and
  Pattern Recognition}, pages 10684--10695, 2022.

\bibitem{UNet_paper}
Olaf Ronneberger, Philipp Fischer, and Thomas Brox.
\newblock U-net: Convolutional networks for biomedical image segmentation.
\newblock In {\em Medical Image Computing and Computer-Assisted
  Intervention--MICCAI 2015: 18th International Conference, Munich, Germany,
  October 5-9, 2015, Proceedings, Part III 18}, pages 234--241. Springer, 2015.

\bibitem{ruiz2022dreambooth}
Nataniel Ruiz, Yuanzhen Li, Varun Jampani, Yael Pritch, Michael Rubinstein, and
  Kfir Aberman.
\newblock Dreambooth: Fine tuning text-to-image diffusion models for
  subject-driven generation.
\newblock {\em arXiv preprint arXiv:2208.12242}, 2022.

\bibitem{imagen}
Chitwan Saharia, William Chan, Saurabh Saxena, Lala Li, Jay Whang, Emily
  Denton, Seyed Kamyar~Seyed Ghasemipour, Burcu~Karagol Ayan, S~Sara Mahdavi,
  Rapha~Gontijo Lopes, et~al.
\newblock Photorealistic text-to-image diffusion models with deep language
  understanding.
\newblock {\em arXiv preprint arXiv:2205.11487}, 2022.

\bibitem{make-a-video}
Uriel Singer, Adam Polyak, Thomas Hayes, Xi Yin, Jie An, Songyang Zhang, Qiyuan
  Hu, Harry Yang, Oron Ashual, Oran Gafni, et~al.
\newblock Make-a-video: Text-to-video generation without text-video data.
\newblock {\em arXiv preprint arXiv:2209.14792}, 2022.

\bibitem{sohl2015deep}
Jascha Sohl-Dickstein, Eric Weiss, Niru Maheswaranathan, and Surya Ganguli.
\newblock Deep unsupervised learning using nonequilibrium thermodynamics.
\newblock In {\em International Conference on Machine Learning}, pages
  2256--2265. PMLR, 2015.

\bibitem{DDIM_paper}
Jiaming Song, Chenlin Meng, and Stefano Ermon.
\newblock Denoising diffusion implicit models.
\newblock {\em arXiv preprint arXiv:2010.02502}, 2020.

\bibitem{song2020score}
Yang Song, Jascha Sohl-Dickstein, Diederik~P Kingma, Abhishek Kumar, Stefano
  Ermon, and Ben Poole.
\newblock Score-based generative modeling through stochastic differential
  equations.
\newblock {\em arXiv preprint arXiv:2011.13456}, 2020.

\bibitem{VQ_VAE_paper}
Aaron Van Den~Oord, Oriol Vinyals, et~al.
\newblock Neural discrete representation learning.
\newblock {\em Advances in neural information processing systems}, 30, 2017.

\bibitem{vaswani2017attention}
Ashish Vaswani, Noam Shazeer, Niki Parmar, Jakob Uszkoreit, Llion Jones,
  Aidan~N Gomez, {\L}ukasz Kaiser, and Illia Polosukhin.
\newblock Attention is all you need.
\newblock {\em Advances in neural information processing systems}, 30, 2017.

\bibitem{villegas2022phenaki}
Ruben Villegas, Mohammad Babaeizadeh, Pieter-Jan Kindermans, Hernan Moraldo,
  Han Zhang, Mohammad~Taghi Saffar, Santiago Castro, Julius Kunze, and Dumitru
  Erhan.
\newblock Phenaki: Variable length video generation from open domain textual
  description.
\newblock {\em arXiv preprint arXiv:2210.02399}, 2022.

\bibitem{wang2021salient}
Wenguan Wang, Qiuxia Lai, Huazhu Fu, Jianbing Shen, Haibin Ling, and Ruigang
  Yang.
\newblock Salient object detection in the deep learning era: An in-depth
  survey.
\newblock {\em IEEE Transactions on Pattern Analysis and Machine Intelligence},
  44(6):3239--3259, 2021.

\bibitem{wu2022nuwa}
Chenfei Wu, Jian Liang, Lei Ji, Fan Yang, Yuejian Fang, Daxin Jiang, and Nan
  Duan.
\newblock N{\"u}wa: Visual synthesis pre-training for neural visual world
  creation.
\newblock In {\em Computer Vision--ECCV 2022: 17th European Conference, Tel
  Aviv, Israel, October 23--27, 2022, Proceedings, Part XVI}, pages 720--736.
  Springer, 2022.

\bibitem{tune-a-video}
Jay~Zhangjie Wu, Yixiao Ge, Xintao Wang, Weixian Lei, Yuchao Gu, Wynne Hsu,
  Ying Shan, Xiaohu Qie, and Mike~Zheng Shou.
\newblock Tune-a-video: One-shot tuning of image diffusion models for
  text-to-video generation.
\newblock {\em arXiv preprint arXiv:2212.11565}, 2022.

\bibitem{xu2018attngan}
Tao Xu, Pengchuan Zhang, Qiuyuan Huang, Han Zhang, Zhe Gan, Xiaolei Huang, and
  Xiaodong He.
\newblock Attngan: Fine-grained text to image generation with attentional
  generative adversarial networks.
\newblock In {\em Proceedings of the IEEE conference on computer vision and
  pattern recognition}, pages 1316--1324, 2018.

\bibitem{xu2022versatile}
Xingqian Xu, Zhangyang Wang, Eric Zhang, Kai Wang, and Humphrey Shi.
\newblock Versatile diffusion: Text, images and variations all in one diffusion
  model.
\newblock {\em arXiv preprint arXiv:2211.08332}, 2022.

\bibitem{parti_paper}
Jiahui Yu, Yuanzhong Xu, Jing~Yu Koh, Thang Luong, Gunjan Baid, Zirui Wang,
  Vijay Vasudevan, Alexander Ku, Yinfei Yang, Burcu~Karagol Ayan, et~al.
\newblock Scaling autoregressive models for content-rich text-to-image
  generation.
\newblock {\em arXiv preprint arXiv:2206.10789}, 2022.

\bibitem{zhang2017stackgan}
Han Zhang, Tao Xu, Hongsheng Li, Shaoting Zhang, Xiaogang Wang, Xiaolei Huang,
  and Dimitris~N Metaxas.
\newblock Stackgan: Text to photo-realistic image synthesis with stacked
  generative adversarial networks.
\newblock In {\em Proceedings of the IEEE international conference on computer
  vision}, pages 5907--5915, 2017.

\bibitem{controlnet}
Lvmin Zhang and Maneesh Agrawala.
\newblock Adding conditional control to text-to-image diffusion models.
\newblock {\em arXiv preprint arXiv:2302.05543}, 2023.

\end{thebibliography}
